\documentclass[twocolumn]{svjour3}
\usepackage{stfloats}
\usepackage[graphicx]{realboxes}
\usepackage{url}
\usepackage{longtable}
\usepackage{color, xcolor}
\usepackage{makecell}
\usepackage{array}
\usepackage{booktabs,lscape}
\usepackage[T1]{fontenc}
\usepackage[misc]{ifsym}
\usepackage[switch]{lineno}

\begin{document}

\title{A Comprehensive Survey with Quantitative Comparison of Image Analysis Methods for Microorganism Biovolume Measurements}

\author{Jiawei Zhang \textsuperscript{1} \and 
Chen Li \textsuperscript{1,\Letter} \and  
Md Mamunur Rahaman \textsuperscript{1,6} \and 
Yudong Yao \textsuperscript{2} \and 
Pingli Ma \textsuperscript{1} \and 
Jinghua Zhang \textsuperscript{1,5} \and 
Xin Zhao \textsuperscript{3} \and 
Tao Jiang  \textsuperscript{4} \and
Marcin Grzegorzek  \textsuperscript{5}}

\authorrunning{J. Zhang et al.}
\titlerunning{A Survey for Microorganism Biovolume Measurement}

\institute{
\Letter Chen Li\\
\email{lichen201096@hotmail.com}\\
\at
{1} Microscopic Image and Medical Image Analysis Group, College of Medicine and Biological Information Engineering, Northeastern University, Shenyang, 110169, China \\
\at
{2} Department of Electrical and Computer Engineering, Stevens Institute of Technology, Hoboken,
NJ 07030, USA\\
\at
{3} School of Resources and Civil Engineering, Northeastern University, Shenyang 110004, China\\
\at
{4} School of Control Engineering, Chengdu University of Information Technology, Chengdu 610225,
China\\
\at
{5} Institute of Medical Informatics, University of Luebeck, Luebeck 23538, Germany\\
\at
{6} School of Computer Science and Engineering, University of New South Wales, Sydney, NSW 2052,  Australia
}

\maketitle              

\begin{abstract}
With the acceleration of urbanization and living standards, microorganisms play an increasingly important role in industrial production, bio-technique, and food safety testing. 
Microorganism biovolume measurements are one of the essential parts of microbial analysis. 
However, traditional manual measurement methods are time-consuming and challenging to measure the characteristics precisely. 
With the development of digital image processing techniques, the characteristics of the microbial population can be detected and quantified. 
The applications of the microorganism biovolume measurement method have developed since the 1980s.
More than 62 articles are reviewed in this study, and the articles are grouped by digital image analysis methods with time. 
This study has high research significance and application value, which can be referred to as microbial researchers to comprehensively understand microorganism biovolume measurements using digital image analysis methods and potential applications.

\end{abstract}

\section{Introduction}

\subsection{Basic knowledge of microorganisms} 
Microorganisms are kinds of tiny organisms that are distributed all around the world. 
They exist in people's daily life but cannot be discovered by naked eyes. 
Various microscopes are designed to observe the microorganism precisely~\cite{Madigan-1997-BBOM}.
Microorganisms are classified into a large number of types based on different classification standards.
Generally, microorganisms consist of bacteria, viruses, fungi, and algae.

Bacteria are the most common unicellular organisms globally, which have a tiny size 
and simple biological structure. 
They lack nuclei, cytoskeletons, and membranous, and most of them are decomposers that can decompose dead organisms into simple inorganic substances and release energy, such as the works of \emph{Bacillus subtilis}. 
Viruses are kinds of microorganisms with a simple structure that can infect other organisms. 
It contains nucleic acid, such as ribonucleic acid (RNA) and deoxyribonucleic acid (DNA). 
The healthy cells are parasitized and produce the nutrition for virus propagation~\cite{Cui-2019-OAEP}. 
Viruses are harmful to human and agricultural production. 
For instance, the severe acute respiratory syndrome coronavirus 2 (SARS-CoV-2) has caused significant loss of life and property all over the world~\cite{Andersen-2020-TPOS,Li-2020-ASMI}.
Fungi are kinds of eukaryotic, sporogenic, chloroplast-free eukaryotes that contain molds, yeasts and other mushrooms known. 
The spores are produced by sexual and asexual propagation. 
Fungi play an essential role in agriculture and industrial production~\cite{Webster-2007-ITF}.
Algae are a group of eukaryotes of the Protista. 
Most of them are aquatic, without vascular bundles, and can carry out photosynthesis. 
Some algae are single-celled dinoflagellates, while others aggregate into colonies.
Algae play essential roles in food production, such as \emph{Undaria pinnatifida} and \emph{Laminaria japonica}, which can be used for medical purposes, like preventing and treating goitre~\cite{Coutteau-1996-MA}.

Microorganisms are in various forms. 
Some are corrupt, which can cause food decomposition and adverse changes to organizational structure, but some are beneficial. For example, healthy people have a large number of bacteria in the stomach. 
The normal flora contains hundreds of species of bacteria. 
These bacteria are mutually dependent and mutualistic in the gut environment, and the disorder of bacteria can lead to diarrhea.
However, some microorganisms are harmful to human health and society production.
Acquired immune deficiency syndrome (AIDS) is a dangerous infectious disease caused by the human immunodeficiency virus (HIV), which can attack the body's immune system. 
The CD4T lymphocytes system, which is an essential part of the human immune system, is attacked violently, the cells are destroyed in large numbers, and causes the loss of immune function in the human body;
SARS-CoV-2 broke out in 2019~\cite{Hui-2020-TCET,Rahaman-2020-ICSC}, and it continues to be widely spread. 
More than 120,000,000 people have been infected globally till March 17th, 2021, which has caused huge economic and productivity losses~\cite{JHU-2020-CCGC}.

Because of microorganisms' vital role globally, related microbial research is continuously developing. 
Microorganism quantification is one of the most important parts of microbial research. 
It is widely used in food and water safety test, biomedical test and environmental surveillance~\cite{Liu-2004-HTIB}. 
There are two main methods for microorganism biovolume measurement one is manual measurement, and the other is computer image analysis-based measurement~\cite{Rajapaksha-2019-ARMT}.
Manual measurement mainly includes the biovolume and dry cell weight (DCW) measurement methods.
The biovolume measurement method reflects the growth state of microorganisms by measuring the volume of mycelia contained in a certain volume of culture medium. 
It has the advantage of being fast and straightforward, but the deviation is relatively large.
The culture medium usually contains non-bacterial solid substances that can cause measurement errors. 
Moreover, it does not apply to the medium for mash fermentation. 
DCW method is applied by collecting a unit volume of microbial culture solution through centrifugation and repeatedly washing the microorganisms with water. 
After frequent pressure or vacuum drying, the total biomass of the culture can be calculated by accurate weighing. 
In an environment with high target microorganism content and less non-bacterial particle impurities, it is more accurate, but it is cumbersome and time-consuming, which is not convenient for online analysis.
The advantages of manual measurements are that when the sample is clear with fewer impurities, the measurement speed is rapid, resulting in higher accuracy.
Nevertheless, when the sample is turbid, the impurities cannot be excluded completely, and the measurement system cannot distinguish impurity debris from target microorganisms, so the target microorganisms cannot be measured accurately. 
Moreover, the measurement results are subjective, and the staff's workload is as heavy and frustrating, probably leading to deviation due to fatigue~\cite{Chien-2007-USEA}.

An automatic computer-aided image analysis system can significantly reduce the workload and enhance accuracy~\cite{Thiran-1994-ARCC,Zhang-2021-AANN}. 
The different microorganism measurement methods are applied because of the different microorganism species and microscopic imaging methods.

\begin{figure}[ht]
\centering
\includegraphics[trim={0cm 0cm 0cm 0cm},clip,width=0.48\textwidth]{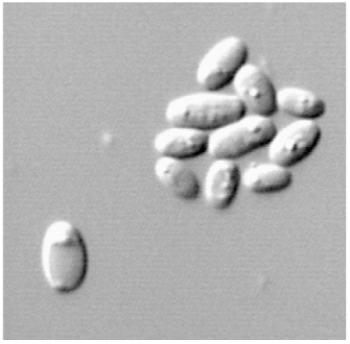}
\caption{The image of yeast cells, which shown the clear boundary between them (in \cite{Gray-2002-CIAS} fig.3(a))}
\label{fig:Gray-2002-CIAS-Example}
\end{figure}

Some images with larger magnification and more scattered colonies can be measured by image analysis.  
In Fig.~\ref{fig:Gray-2002-CIAS-Example}, an image of a \emph{Candida} yeast cell is shown. 
It has shown a clear boundary that can be separated. 
After that, the measurement of parameters such as area and perimeter can be obtained readily, and the number of objects can be counted precisely.
The corresponding survey of microorganism counting approaches can be found in our latest work~\cite{Zhang-2021-ACRI} for detail.

\begin{figure}[ht]
\centering
\includegraphics[trim={0cm 0cm 0cm 0cm},clip,width=0.48\textwidth]{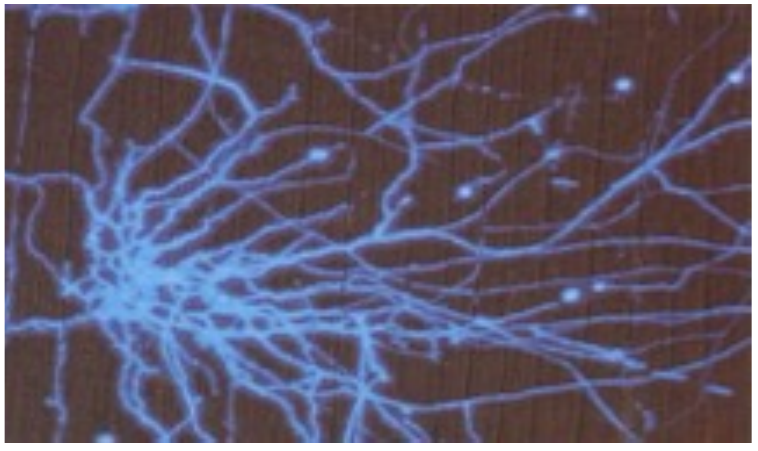}
\caption{The image of fungi hyphae in soil that are stained for observing(in~\cite{Morgan-1991-AIAM} fig.2)}
\label{fig:Morgan-1991-AIAM-Example}
\end{figure}

The approaches to microorganism biovolume measurement are different from those of microorganism counting. 
Biovolumes can be measured from each species or genus's geometric shape and size, including categorizing some genera by morphology. 
However, in some cases, the biovolume cannot be measured by using relationships between microorganism counts and the biovolume of each identified microorganism species or genus~\cite{Rousso-2022-CSDA}.
Some images are blurred with smaller magnification and not clear enough for separation.
For example, algae and fungi are entangled with each other that cannot be separated. 
Biovolume measurement can be used for quantification for this purpose. 
In Fig.~\ref{fig:Morgan-1991-AIAM-Example}, an image of fungal hyphae with complex distribution is illustrated.

\begin{figure}[ht]
\centering
\includegraphics[trim={0cm 0cm 0cm 0cm},clip,width=0.49\textwidth]{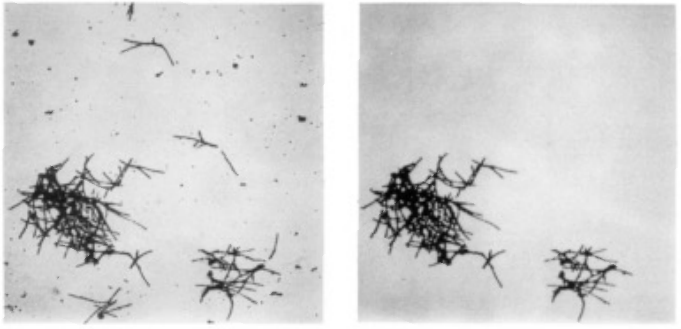}
\caption{The mycelia images obtained through the program (in~\cite{Tucker-1992-FAMM} fig.1)}
\label{fig:Tucker-1992-FAMM-Example}
\end{figure}

In Fig.~\ref{fig:Tucker-1992-FAMM-Example}, the mycelia image is shown. 
The classical measurement method of mycelial morphology relies on inaccurate and time-consuming manual measurements from images because of the random distribution of entangled mycelial. 
Moreover, the method cannot measure the hyphal branches individually.

\subsection{Motivation of this review} 

\begin{figure}[ht]
\centering
\includegraphics[trim={0cm 0cm 0cm 0cm},clip,width=0.48\textwidth]{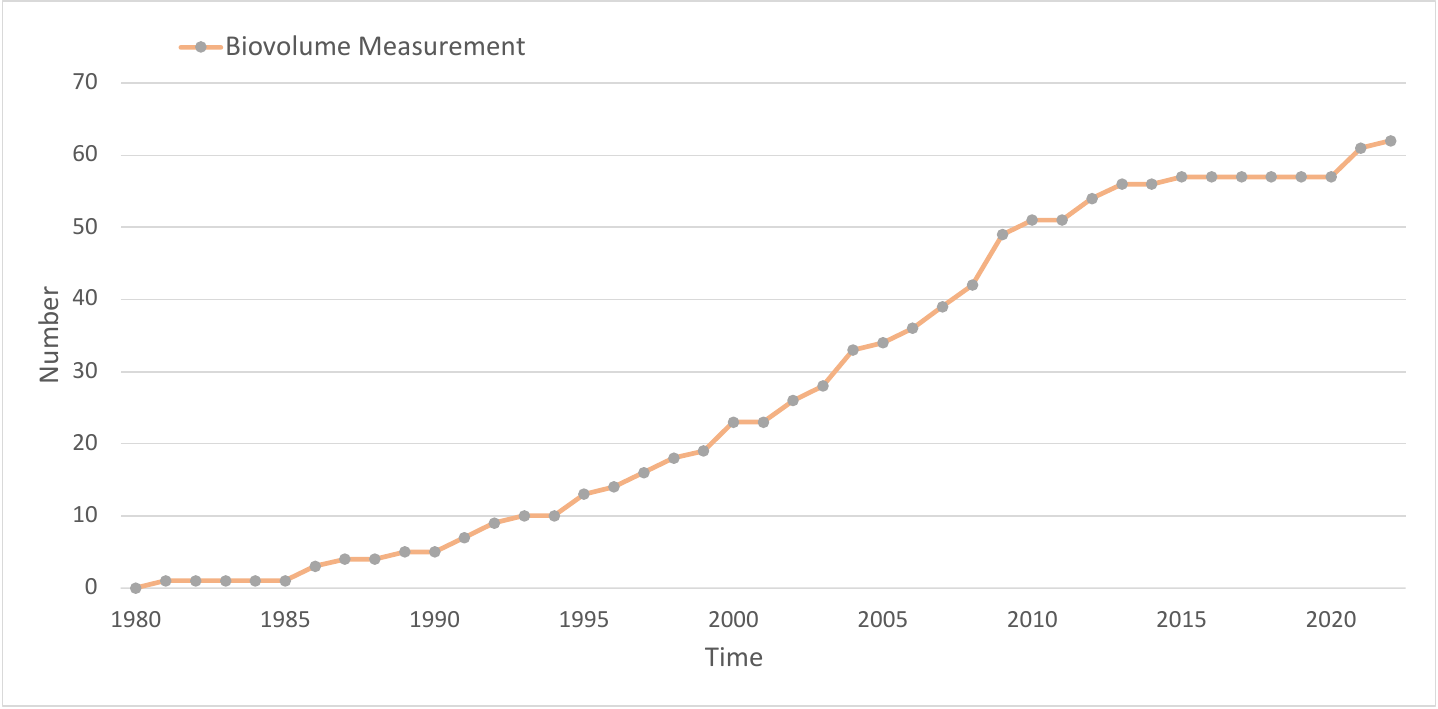}
\caption{The total number of related works on microorganism biovolume measurement approaches}
\label{fig:totalnum}
\end{figure}

Digital Image Processing (DIP) is applied for image denoising, image enhancement, image segmentation and feature extraction by using a computer.
DIP is firstly used in the 1950s, the computer is designed and can be applied to process graphics information to improve the image resolution~\cite{Gonzalez-2004-DIPU}. 
DIP has been widely used in many fields. 
Agricultural and forestry departments can detect pesticides in vegetables, identify the rice varieties,
design various piking robots and monitor the management of diseases and insect pests
~\cite{Amrita-2016-IPTA}. 
The water conservancy department can prevent the disaster of water in advance through remote sensing image analysis~\cite{Dodi-2012-APIP}. 
Image segmentation technology, image description method and pattern recognition technology are used to identify and detect fire disaster~\cite{Pesatori-2013-IISF}. 
The traffic controlling department can use DIP to monitor the road condition, recognize license plate number, and violate detection~\cite{Ebrahimi-2015-CLPR}. 
Medical departments can use amounts of DIP technology to automatically classify and segment various diseases~\cite{Chen-2020-ARCH,Li-2021-ACRC,Li-2021-ACRM,Zhou-2020-ACRB,Liu-2022-ITAT}. 
The professional equipment is necessary for microbial researches to achieve the better precision. 
However, the use of DIP can reduce the equipment costs~\cite{Ekstrom-2012-DIPT}. 
Meanwhile, DIP has been applied in the field of microorganism biovolume measurement widely. 
The trend of development is shown in Fig.~\ref{fig:totalnum}.

It can be seen that the use of DIP in the field of microorganism biovolume measurement maintains an upward trend. 
Since 1980s, DIP has been applied to microorganism biovolume measurement. 
From 1980 to 1995, the development are relatively slow. 
The developments of microorganism  biovolume measurement are relatively rapid from 1995 to 2010. 
However, after 2010, the biovolume measurement related research entered the platform period.  
According to the content of the papers, summed up two possible reasons for the development situation. 
At first, due to the development of the algorithm, especially the development of deep learning, 
more accurate image segmentation can be achieved. 
For example, the segmentation of adherent colonies can lead to more accurate microorganism counting. 
Secondly, it is very difficult to obtain the three-dimensional data only from the image.  
Image analysis can obtain the surface areas of microorganisms,  but cannot accurately calculate their volume.  
Accurate quantification will be a challenging task without a more accurate instrument.

Since microorganism counting and biovolume measurement suit different cases: 
In our previous work~\cite{Zhang-2021-ACRI}, there were more than 132 papers for microorganism counting; 
in this study, there are 58 papers for biovolume measurement. 
Among all these papers, only four papers are overlapping~\cite{Qiu-2004-AMTA,Gracias-2004-ARCD,Daims-2007-QUMF,Dazzo-2015-UCIA}, both for microorganism counting and biovolume measurement, indicating that they are two different research topics.

\subsection{Related reviews}
Due to the biovolume measurement is an essential topic in the research of microorganisms, plenty of literature reviews are published based on the abundant research, which are summed up as follows:

B{\"o}lter et \textit{al}.~\cite{Bolter-2002-EABD} describes the research progress of microbial direct microorganism counting method, and discusses the technology and application of each direction from two aspects of microorganism counting and biovolume measurement. 
This review discusses data analysis and error propagation, which are widely neglected in the quantitative study of microbial ecology, and compares different microbial staining techniques and their applications. 
This review discusses the common staining methods and compares the advantages of each method. 
In this review, there are 3 papers are about calculating the error propagation and more than 23 papers 
are about biovolume measurement.

Qiu et \textit{al}.~\cite{Qiu-2004-AMTA} reviews the application of both size measurement and counting for bacteria. The automatic flow analysis technology and classical counting methods are introduced.
At first, the plate counting method is the earliest developed method, then the immunofluorescence microscopic method and most probable number (MPN) are wildly used.
With the development of computer technologies, DIP has shown the extremely convenience for microbial counting analysis. 
In the field of cell measurement, the different microorganisms can be detected and classified by using the water flow image method. 
However, the microorganisms with small size are difficult to be detected. 
The application of resistance change method can precisely measure the number and size of the individual cell, but the classification is difficult.
Another measurement method with fast speed and high accuracy is flow cytometry. 
However, the concentrations of samples are limited to be high.
There are more than 25 papers are summarized to illustrate the application of the size measurement of cell.

Gracias et \textit{al}.~\cite{Gracias-2004-ARCD} summarized the application of chromogenic and impedance method for microorganism classification and counting. 
The classical microbial measurement methods for food safety testing are illustrated.  There are more than 28 papers are about microorganism measurement methods.

Daims et \textit{al}.~\cite{Daims-2007-QUMF} describes the microbial quantification methods based on DIP. 
The material difference of microbial counting and microbial biovolume measurement is the identification of a single cell. 
Thresholding and edge detection are most widely applied methods by using DIP. And the development of local thresholding can focus on the region of interest and obtain the better segmentation result comparing with global thresholding. 
There are 6 papers are about automatic cell counting, 14 papers are about quantification of cell size and 11 papers are about biomass quantification.

Dazzo et \textit{al}.~\cite{Dazzo-2013-SEMB} describes the application of CMEIAS 
(Center for Microbial Ecology Image Analysis System) computer-assisted microscopy 
for data extraction from images after accurately segmented, that can provide 63 different 
insights into the ecophysiology of microbial populations and communities within biofilms and other habitats. 
There are eight quantitative assessments topics are proposed, that are morphological diversity as an indicator of impacts that substratum physicochemistries have on biofilm community structure and dominance-rarity relationships among populations, morphotype specific distributions of biovolume body size that relate microbial allometric scaling, metabolic activity and growth physiology, fractal geometry of optimal cellular positioning for efficient utilization of allocated nutrient resources, morphotype-specific stress responses to starvation, environmental disturbance and bacteriovory predation, patterns of spatial distribution indicating positive and negative cell–cell interactions affecting their colonization behaviour, and significant improvements to increase the accuracy of color-discriminated ecophysiology. 
More than 50 papers are used to describe quantification of biofilms. 

Costa et \textit{al}.~\cite{Costa-2013-QIAT} describes the development of biological wastewater treatment (WWT) using image analysis.  This review introduces the fields of application of image analysis based WWT, 
such as aerobic wastewater treatment biological processes, calculation of sludge settling ability, measurement 
of sludge contents, detection of toxic compounds, biomass physiology analysis, full-scale WWT plants, 
aerobic granulation detection, quantification of the physical characteristics of the anaerobic granular sludge, 
analysis the behavior of anaerobic granules and biomass activity research.  More than 110 papers are used to 
list the application of image analysis in WWT field but there is no definite description for microorganism 
counting or biovolume measurement, and there is no mention for technique methods.

Dazzo et \textit{al}.~\cite{Dazzo-2015-UCIA} introduced the application of CMEIAS for microbial counting and biovolume measurement based on DIP. 
The shape-adaptable methods can be applied to measure the length and width of cell for the calculation of cell size. 
Then, a supervised tree is applied for individual cell classification.
After that, a k-Nearest Neighbour (kNN) classifier is used to cluster the complex particles, such as spherical coccus and branched filament.
Moreover, several parameters are calculated to location the spatial distribution of microorganisms in the biofilm. 
The extracted data can be analyzed by geospatial statistics, that is used to define the biogeography of microbial cells during the development of biofilm. And the colonization behavior models are generated by using the cell-to-cell interactions intensity. 
In the fourth part, the image segmentation algorithms are applied to show the color and 
spatial relationships of the foreground pixels. 
Finally, the applications of CMEIAS is summarized that contains the image analysis for morphological 
diversity, filamentous microbial morphotypes, ecophysiological attributes linked to accurate measures of 
biovolume body mass, spatial pattern analysis and its relationship to microbial biofilm ecology and color 
segmentation tool for cell-cell communication. More than 35 papers are about measurement of biovolume.

Li et \textit{al}.~\cite{Li-2019-ASTA} introduces the development of microbial analysis based on DIP and amounts of methods used for microorganism classification are summarized. In this survey, 
the microorganisms are grouped based on their application domains, including Agricultural Microorganisms (AMs), Food Microorganisms (FMs), Environmental Microorganisms (EMs), Industrial Microorganisms (IMs), 
Medical Microorganisms (MMs), Water-borne Microorganisms (WMs) and Scientific Microorganisms (SMs). 
They analyse the properties of the methods below and the practical conditions of the microorganism 
classification tasks jointly, the methods include image preprocessing methods such as image segmentation, 
feature extraction methods, post-processing methods,  feature fusion methods and classification methods. More than 60 related papers are summarized to illustrate different microbial classification methods. 
This review is a comprehensive microorganism classification paper and uses plenty of literatures for quoting, but there is no significant description for microorganism counting or biovolume measurement.

\begin{figure}[ht]
\centering
\includegraphics[trim={0cm 0cm 0cm 0cm},clip,width=0.48\textwidth]{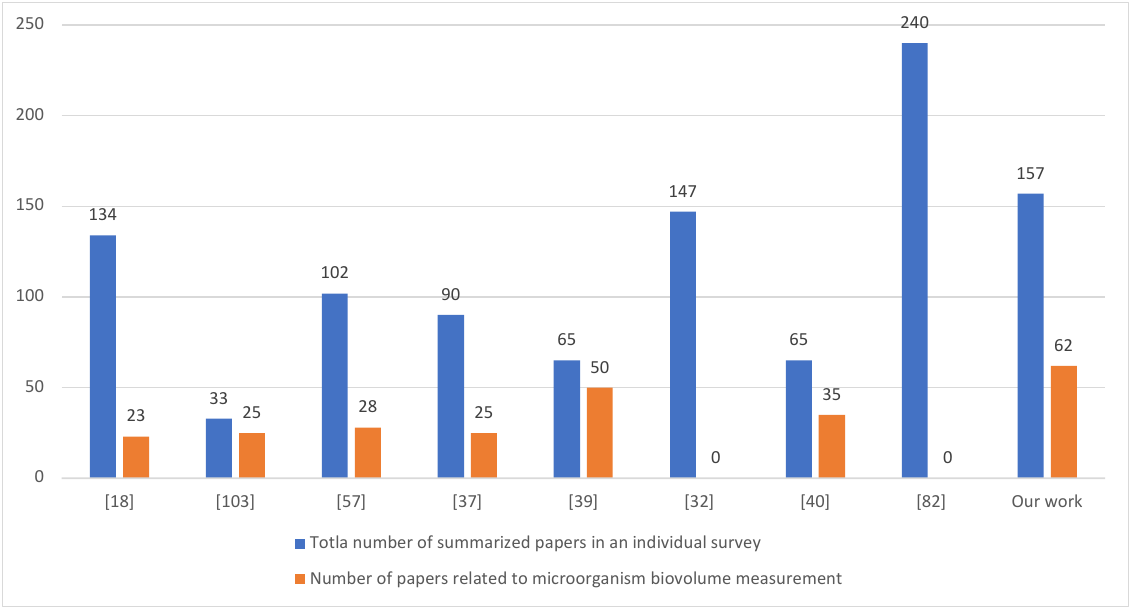}
\caption{The comparation of related survey papers. The related works are list with their contribution to image analysis based microorganism biovolume measurement}
\label{fig:reviewnumber}
\end{figure}

The comparation of several related works and their contribution to biovolume measurement are shown in Fig.~\ref{fig:reviewnumber}.
It is observed from the previous studies that the explanations about the current technologies of microorganism research are objective and detailed. 
However, the targeted research about biovolume measurement is limited. 
Therefore, additional research on this aspect needs to be completed. 
Since microorganism quantification plays a vital role in microbial research, this review focuses on applying microorganism biovolume measurement, and the prospects of the methods. 
This review has high research significance and application value for microbiological researchers and computer vision researchers.  There are more than 62 papers are summarized in this survey about biovolume measurement.

\subsection{Microorganism  biovolume measurement methods} 
\begin{figure*}[hb]
\centering
\includegraphics[trim={0cm 0cm 0cm 0cm},clip,width=1.0\textwidth]{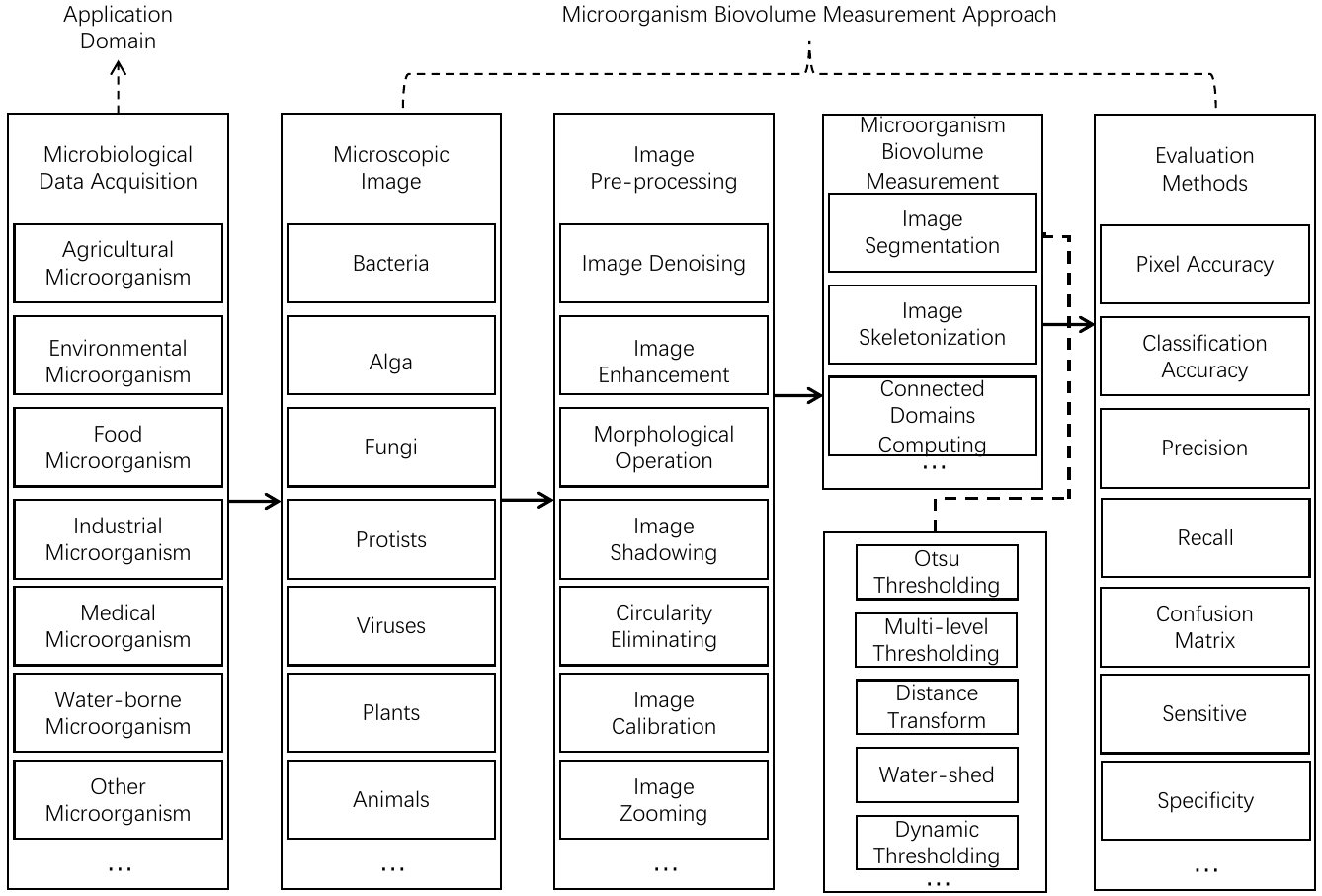}
\caption{The organisation chart of microorganism biovolume measurement approaches in this paper}
\label{fig:flowchart}
\end{figure*}

The flow chart of microbial biovolume measurement is shown in Fig.~\ref{fig:flowchart} to elaborate on the methods. 
The approach contains five steps: microbiological data acquisition, microscopic image, image preprocessing, microorganism biovolume measurement methods, and evaluation methods.

The microorganisms are composed of the following seven types according to the different application
domains, including agricultural microorganisms, environmental microorganisms, food microorganisms, industrial microorganisms, medical microorganisms, water-borne microorganisms and other microorganisms.
Then, the digital images are obtained after staining and slicing by using professional equipment~\cite{Gmur-2000-AIES}. 
The following part is pre-processing, including image denoising and contrast enhancement, which is helpful for image segmentation.
The next step is biovolume measurement methods~\cite{Li-2020-ARCM}.
Biovolume measurement aims to measure the morphological parameters and calculate the biovolume. 
The most significant part of microorganism biovolume measurement is image segmentation.
Finally, the evaluation methods are proposed.
However, the results obtained in these studies are area and volume, that are difficult to be compared with the 
results obtained by manual methods. 
So the proper evaluation methods can help to analyze the performance of the image analysis methods~\cite{Andreini-2016-AICT}.  

\subsection{Structure of this review} 

\begin{figure}[ht]
\centering
\includegraphics[trim={0cm 0cm 0cm 0cm},clip,width=0.5\textwidth]{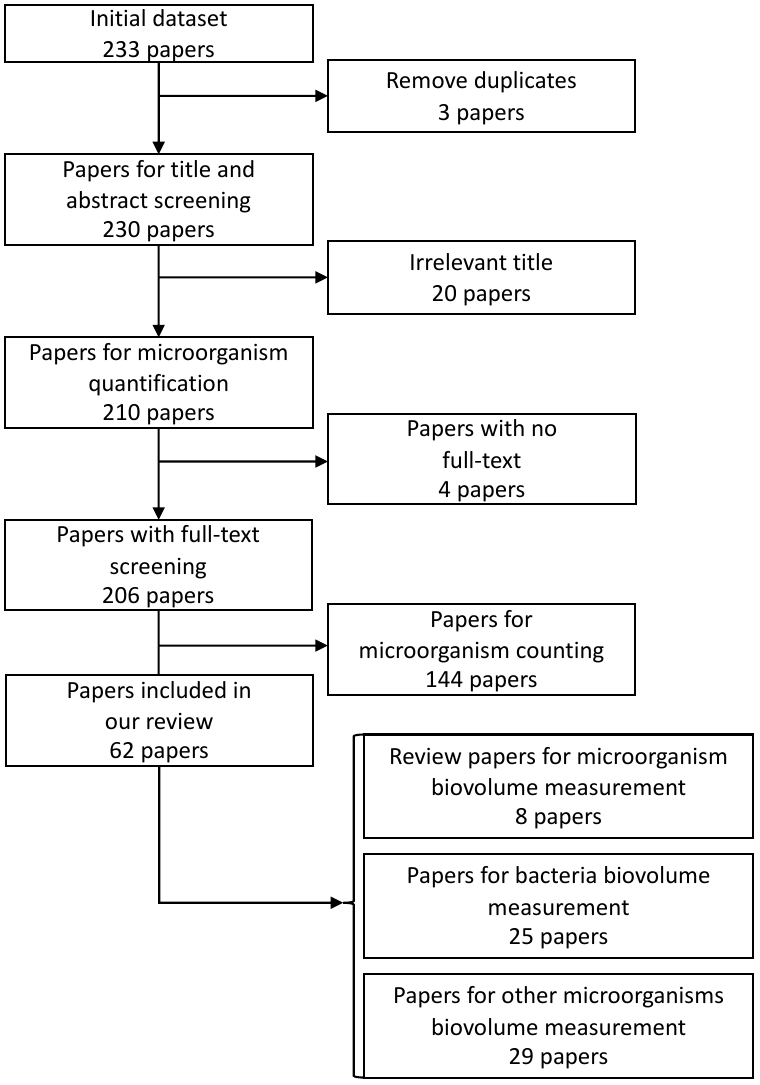}
\caption{The systematic flow chart of paper selection for our work}
\label{fig:flowchart2}
\end{figure}

This study proposes a comprehensive review of microorganism biovolume measurement using image analysis. 
The research of microbial application was first developed in 1980, and the articles, including
research articles and review articles, are summarized. 
Moreover, the application of microorganism biovolume measurement in various circumstances are discussed. 
More than 62 papers are selected from the initial paper dataset, and the structure of the systematic review is shown in Fig.~\ref{fig:flowchart2}.

The review is organized as follows: 
In Sec. 2, the DIP and evaluation approaches, which are widely applied for microorganism biovolume measurement, are summarized. 
In Sec. 3, the relevant research on DIP-based microorganism measurement are grouped and summarized.
In Sec.4, the quantitative analysis of different DIP approaches for microorganism biovolume measurement is summarized.
Then in Sec. 5, the commonly used approaches are summarized. 
Finally, in Sec. 6, this review is concluded by summarizing the whole paper. 
This review structure is clear enough to help researchers comprehend the DIP-based microorganism biovolume measurement's current situation.

\section{Image analysis based microorganism quantification methods}

The application of DIP technology to microbial biovolume measurement can promote stability and higher accuracy by comparing it with traditional measurement. 
This chapter introduces the commonly used DIP technologies applied in microorganism biovolume measurement.

\subsection{Image analysis technology} 
\subsubsection{Image pre-processing} 
In the process of image generation and transmission, the image quality is often degraded due 
to the interference and influence of various noises, which will have a negative impact on the 
subsequent image processing and image visual effect. 
Therefore, in order to suppress noise and improve image quality, denoising methods need to be applied.
The main methods to remove noise are spatial filter, transform-domain filter and so on.  
Spatial filtering directly performs data operation on the original image and processes the grey 
value of the pixel, such as the Median filter and low-pass filter~\cite{Li-2020-ARCM}. 
The transform-domain filter contains Fourier transform and wavelet transform~\cite{Pan-1999-TDMW}.

\subsubsection{Image segmentation} 
Image segmentation is an important method in digital image processing and computer vision, which 
contains threshold segmentation, edge detection and region extraction.

\paragraph{Threshold segmentation}
Thresholding is one of the classical methods for image segmentation, which has the advantages of simple calculation, high efficiency and fast speed~\cite{Zhu-2007-AISA}. 
The selection of the threshold is the core algorithm of threshold segmentation~\cite{Sekertekin-2021-ASGT}.  
There are two main methods, the first one is iterative thresholding method, which can obtain the good results
for the images with high global contrast~\cite{Perez-1987-AITA}.
Another one is the maximum inter class variance based on Otsu method, which can obtain the satisfactory 
result for most images~\cite{Otsu-1979-ATSM}. This method is considered to be the best method for automatic threshold selection because it is easy to calculate and is not affected by the change of image contrast and brightness~\cite{Xu-2011-CAOT}.

\paragraph{Edge detection based segmentation} 
In the DIP, edge detection can reduce the amount of data and maintain the essential architecture 
of the image~\cite{Haralick-1985-IST}. 
Edge detection includes gradient operator based method, second-order differential operator based 
method, LOG edge detection method and Canny edge detection method. The gradient operator based 
methods contain Roberts operator, Sobel operator, Prewitt operator, Kirsch operator and Robinson 
operator, which are the most widely applied for edge detection~\cite{Gonzales-2002-DIP}.
Sobel operator detection method has a good performance when image have a wide variety of noise, 
but the edge location is not accurate and the edge of the image is more than one pixel. 

\paragraph{Region based segmentation} 
Among the region extraction methods, watershed segmentation provides excellent results and widely 
used for segmentation. It belongs to a segmentation method based on region growth. 
The calculation process of the watershed is an iterative labeling operation, which is sensitive to feeble contours~\cite{Strahler-1957-QAWG}.

\subsubsection{Morphological operation} 
The commonly used morphological operations include dilation, erosion, close operation, open operation and so on, which are usually applied in pre- and post-processing in DIP~\cite{Mukhopadhyay-2003-MMSG}. The satisfactory results can be obtained after morphological operations.

\subsubsection{Morphological features} 
The commonly used features in biovolume measurement consist area, perimeter, length and width. 
The area can be obtained by calculating the inner pixels, the perimeter can be calculated by counting the contour pixels. 
The length and width can be obtained from the features of the central axis~\cite{Ziegler-2004-MFCD}.

\subsection{Evaluation} 
Evaluation is an essential part of DIP, which can help researchers find the reasons that affect the result.  
True positive (TP), false negative (FN), false positive (FP) and true negative (TN) are four basic metrics in image classification. 
Commonly used evaluation metrics are shown in Table~\ref{tab:metric}.

\renewcommand\arraystretch{2}
\begin{table*}[hb]\footnotesize
\centering
\caption{\label{tab:metric}The definitions of evaluation metrics for image classification and image segmentation.} 
\begin{tabular}{m{2.0cm}<{\centering}m{5.8cm}<{\centering}m{2.0cm}<{\centering}m{5.8cm}<{\centering}} 
\hline
\makecell[c]{Metric} & \makecell[c]{Definition} & \makecell[c]{Metric} & \makecell[c]{Definition}\\
\hline
\makecell[c]{Accuracy} & \makecell[c]{Accuracy = $\rm \frac{TP+TN}{TP+TN+FP+FN}$} & \makecell[c]{Precision} & \makecell[c]{Precision = $\rm \frac{TP}{TP+FP}$}\\ 
\makecell[c]{Recall} & \makecell[c]{Recall = $\rm \frac{TP}{TP+FN}$} & \makecell[c]{F1-score} & \makecell[c]{F1-score = $\rm \frac{1}{Precision} +  \frac{1}{Recall}$}\\ 
\makecell[c]{PA} & \makecell[c]{PA = $\rm \frac{\sum_{i=0}^{k}P_{ii}}{\sum_{i=0}^{k}\sum_{j=0}^{k}P_{ij}}$} & \makecell[c]{MPA} & \makecell[c]{MPA = $\rm \frac{1}{K+1}\sum_{i=0}^{k}\frac{P_{ii}}{\sum_{j=0}^{k}P_{ij}}$}\\
\makecell[c]{IoU} &\makecell[c]{IoU = $\rm \frac{TP}{FN+TP+FP}$} & \makecell[c]{DIce} & \makecell[c]{DIce = $\rm \frac{2TP}{FN+2TP+FP}$} \\
\hline
\end{tabular} 
\end{table*}

\subsubsection{Classification accuracy} 
Accuracy is the proportion of correctly classified samples from all samples~\cite{Mariey-2001-DCIM}. 
TP, FN, FP and TN can be composed as confusion matrix~\cite{Hay-1988-TDGE}. 
The columns of confusion matrix are predicted results and rows are real classification.
Precision is the probability of samples which are correct predicted from the predicted positive samples. 
Recall is the proportion of positive prediction from all positive samples~\cite{Buckland-1994-TRBR}. 
F1-score is the weighted harmonic average of precision and recall~\cite{Chicco-2020-TATM}.

\subsubsection{Segmentation accuracy} 

Pixel accuracy (PA) is a basic evaluation method in image segmentation, that is the proportion of pixels that are classified correctly. 
The mean pixel accuracy (MPA) is the improved methods of PA, which performs more objective~\cite{Zhang-2008-ISEA}.   
Dice is applied to measure the inner similarity between segmented image and GT~\cite{Anuar-2010-VCPU}.
Intersection over union (IoU), also named Jaccard,  is the proportion of intersection and union of the prediction and ground truth~\cite{Rahman-2016-OIOU}.

Evaluation is one of the most essential parts of DIP,  which is the performance measurement of the model.
The results of evaluation can affect the performance of microorganism biovolume measurement directly.

\section{Microorganism biovolume measurement methods}
Microorganisms come in various types, and different microorganism biovolume measurement methods show different performances. 
The microorganism biovolume measurement methods are suitable when colonies are adherent. 
In this chapter, the current applications are summarized and structured as follows: 
firstly, the semi-automatic methods, gray level histogram methods, thresholding methods, morphological methods, edge detection methods, and third-party tools for bacteria biovolume measurement are summarized. 
After that, the image enhancement-based methods, thresholding methods, machine learning-based methods, edge detection-based methods, region connection-based methods, morphological methods, and third-party tools methods of other microorganisms measurement methods are summarized, including animal, alga, plant, fungi, spore, virus, and protozoa.

\subsection{Bacteria measurement methods}
Bacteria is one of the essential parts of microorganisms, and the datasets of bacteria are relatively abundant, which are usually used in microbial research.

\subsubsection{Bacteria measurement methods based on semi-automatic operation}
In~\cite{Krambeck-1981-MABD}, an scanning electron micrograph based dialogue program is developed.  
The system needs users to mark the width and length of all bacteria and then calculates the biovolume increment of bacteria. 
The total time required could be reduced from 1 day to 2.5 h by using a M6520 microcomputer program.  

\subsubsection{Bacteria measurement methods based on gray level  histogram}

In~\cite{Camp-1992-CATD}, the mean value of the RGB histogram and the ratio of red to blue intensity components are used to determine the concentration of bacteria biomass. 
The sum of the mean values of the RGB distributions are linearly related to the total biomass protein concentration.

\begin{figure}[ht]
\centering
\includegraphics[trim={0cm 0cm 0cm 0cm},clip,width=0.49\textwidth]{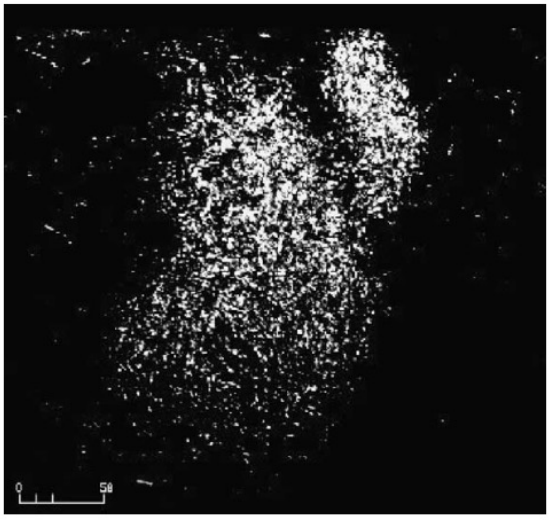}
\caption{The 3-D reconstruction image of the biovolume, which is obtained based on the five biofilm sections (In~\cite{Petrisor-2004-RACM} fig.6)}
\label{fig:Petrisor-2004-RACM}
\end{figure}

In~\cite{Lomander-2002-AMRA,Petrisor-2004-RACM}, biofilm is applied for reconstruction and quantification of bacteria biovolume.
First, a median filter~\cite{Lomander-2002-AMRA} and contrast stretching~\cite{Petrisor-2004-RACM} are applied for noise reduction, and then the image contrast is enhanced by using histogram stretching. 
After that, the method of histogram equalization and a high pass filter~\cite{Petrisor-2004-RACM} is applied to enhance images. 
In~\cite{Lomander-2002-AMRA}, the dead biofilm patches are eliminated by comparing the patch morphology with substrate morphology, and three metrics are then computed for the individual biofilm patches: patch area, patch perimeter, and circularity, which can be applied for bacteria colony biovolume measurement.
The result shows that the patch morphology is useful for investigating the relationships between surface morphology with biofilm growth and determining the surface morphology and biofilm shape.
In~\cite{Petrisor-2004-RACM}, the geographical information systems (GIS) are used to reconstruct the biofilm images and measure the biovolume of bacteria. 
The 3-D reconstruction of the bacteria volume is shown in Fig.~\ref{fig:Petrisor-2004-RACM}.

\subsubsection{Bacteria measurement methods based on thresholding}

\begin{figure}[ht]
\centering
\includegraphics[trim={0cm 0cm 0cm 0cm},clip,width=0.48\textwidth]{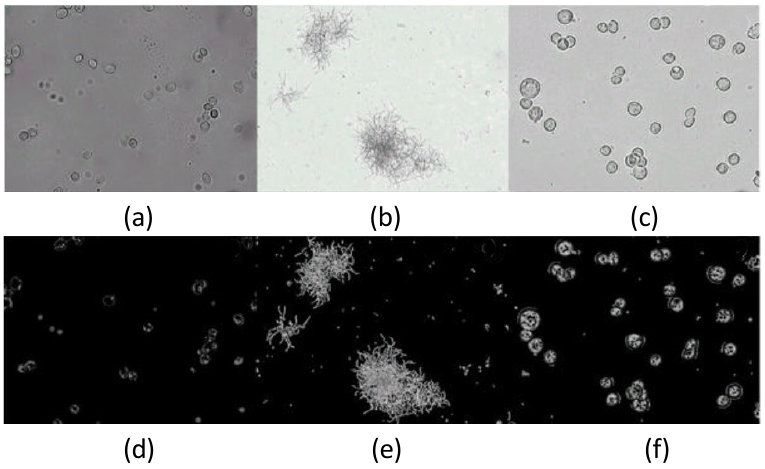}
\caption{The original and processed images. (a) yeast; (b) bacteria; (c) insect cell; (d) processed image of yeast; (e) processed image of bacteria; (f) processed image of insect cell (In~\cite{Garofano-2005-ATWI} fig.5)}
\label{fig:Garofano-2005-ATWI}
\end{figure}

In~\cite{Bjornsen-1986-ADBB,Heydorn-2000-ABST,Garofano-2005-ATWI}, thresholding is applied for bacteria biovolume measurement.
First, contour enhancement~\cite{Bjornsen-1986-ADBB} is applied for pre-processing.
Then the thresholding methods with fixed value are applied for determination of bacteria biomass.
The system can detect and measure the objects after segmentation and erosion operation, which provides estimates of cell number,  mean volume and biovolume with standard errors of about 5\%.
In~\cite{Heydorn-2000-ABST}, the connected-volume filtration can erase debris outside the region of interest after thresholding. 
Finally, the quantity of elements can be measured.
In~\cite{Garofano-2005-ATWI}, the useless particles are removed after thresholding segmentation. 
Then the wavelet transformation and densitometric techniques are applied to quantitatively estimate the growth of cells. 
The measurements of densitometric techniques are based on the measurements of the area occupied by the cells, the maximum and average cell grey-scale tones. 
It shows the predominance of larger wavelet coefficients for the images with a higher biomass content or number of cells. 
The processed images are shown in Fig.~\ref{fig:Garofano-2005-ATWI}.

\begin{figure}[ht]
\centering
\includegraphics[trim={0cm 0cm 0cm 0cm},clip,width=0.48\textwidth]{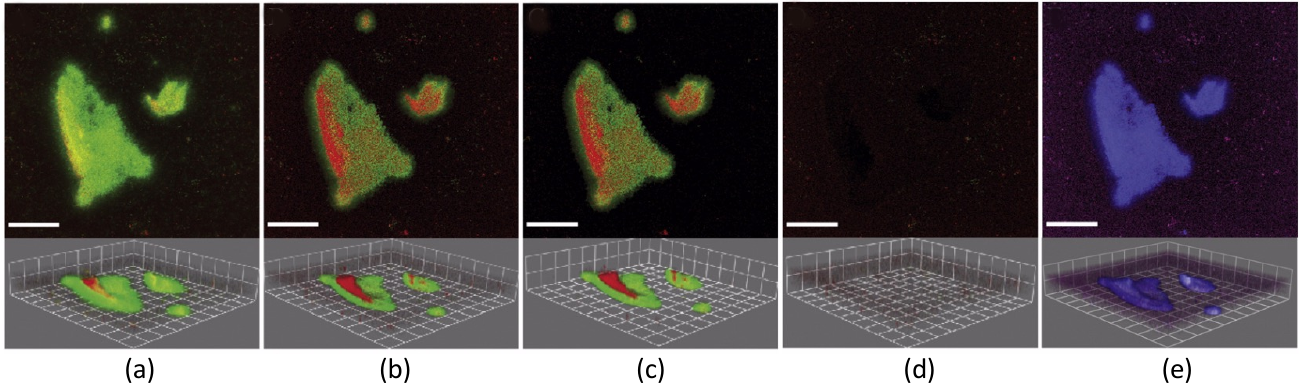}
\caption{The result images after processing (In~\cite{Ross-2012-EABS} fig.4)}
\label{fig:Ross-2012-EABS}
\end{figure}

In~\cite{Mueller-2006-AAMP,Ross-2012-EABS}, Otsu Thresholding is applied for bacteria biovolume measurement. 
Both methods use confocal laser scanning microscopy (CLSM) images of bacteria for quantification.
First, Otsu Thresholding is applied for CLSM images segmentation.
Then in~\cite{Mueller-2006-AAMP}, the parameters of biofilm are obtained by calculating the binary image stacks with connected volume filtration (CVF). 
The number of fore-ground pixels are multiplied by the volume of voxel in a stack, that represents the biovolume. 
The volume of voxel is the square of the pixel size multiplied by the scanned step size.
In~\cite{Ross-2012-EABS}, the color channels are connected with the connected volume filtration matrix (1 = connected-biofilm bacteria and 0 = either unconnected bacteria or background noise) after Otsu Thresholding.
The same pixel location of dead and live bacteria is compared. 
If both live and dead channels are 1, the dead channel remains 1 and the live channel converts to 0. 
Moreover, if the CVF matrix has the same value as the color channel, this pixel is labeled as connected bacteria. 
On the contrary, if the value of color channel is one, but the CVF matrix has a value of 0, the particular pixel is labeled as unconnected bacteria. 
Furthermore, if the CVF matrix has the same value of 0 as the color channel, this pixel is labeled as noise. 
Finally, the green and red channel points are summed up to calculate the percentage of dead and live bacteria populations. 
The result is shown in Fig.~\ref{fig:Ross-2012-EABS}.

In~\cite{Boman-2008-SDBS}, the adaptive thresholding is used to determine the structure of bacterial biofilm.
The images of bacteria biofilm are obtained using CLSM, then an adaptive thresholding is 
used for image segmentation. 
Finally,  the image structure analyzer is used to extract the features of biofilm and the thickness of biofilm is calculated for biovolume quantification.

\subsubsection{Bacteria measurement methods based on morphological operation}

In~\cite{Mesquita-2010-DAME}, an image analysis system is used to determine the biovolume of 
filamentous bacteria. 
The image processing program consists of three stages: image pretreatment, segmentation and debris elimination. 
Finally, the morphological parameters are measured to calculate the total recognized aggregates area per volume, the number of recognized aggregates per image, percentage number of recognized aggregates and recognized aggregates area percentage.

In~\cite{Bolter-1993-DBBE}, the images are separated into three types that contains cocci, rods and 
others, then different formulas are used to calculate the biovolume of each type.  
The integration of volume performs better than simple addition.

\begin{figure}[ht]
\centering
\includegraphics[trim={0cm 0cm 0cm 0cm},clip,width=0.48\textwidth]{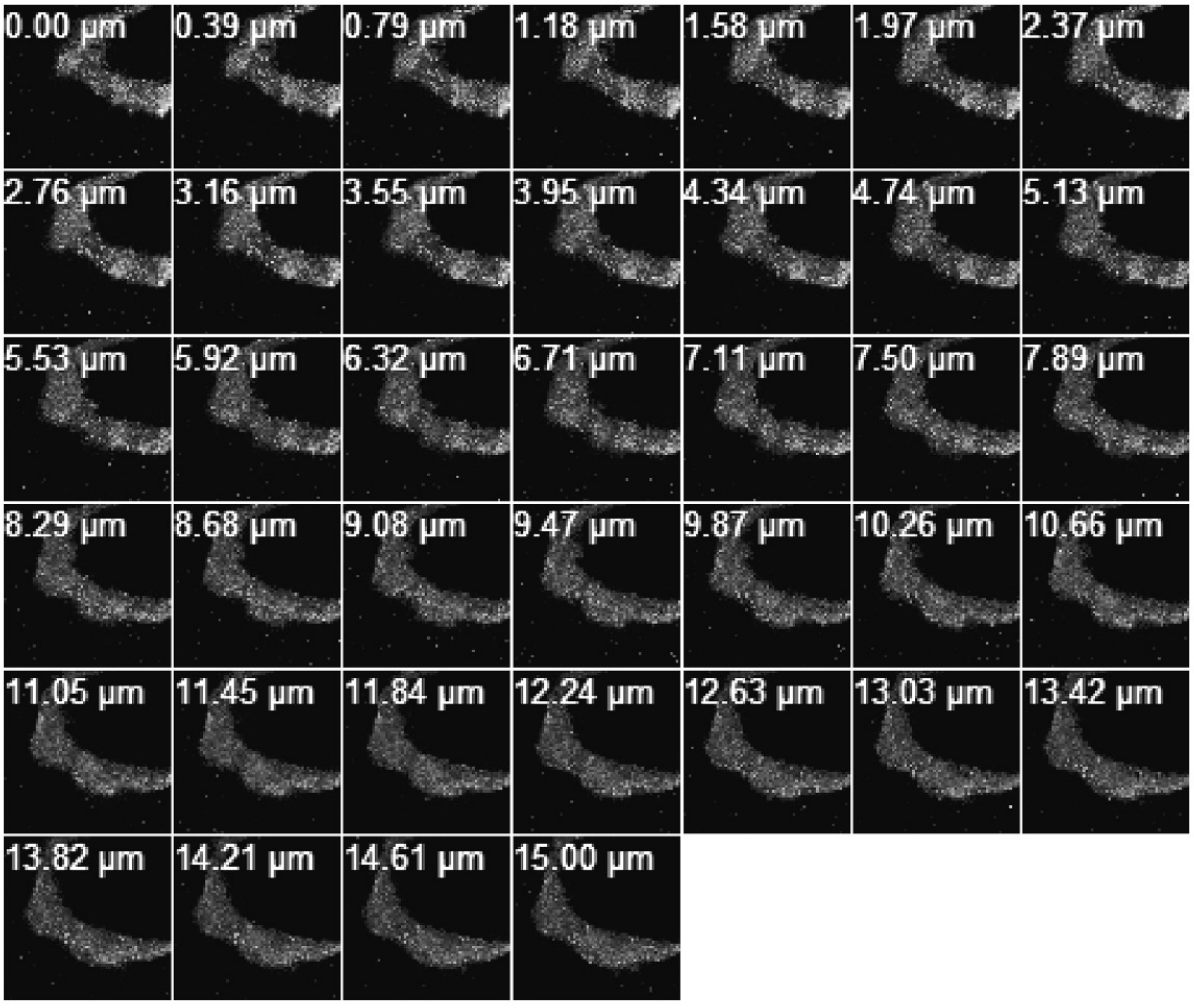}
\caption{The biofilm images captured by CLSM (In~\cite{Rodriguez-2007-TDQS} fig.3)}
\label{fig:Rodriguez-2007-TDQS}
\end{figure}

In~\cite{Rodriguez-2007-TDQS}, more than 30 biofilm samples are captured for measurements 
of several parameters, such as the area of biosurface, biovolume, and biothickness. 
After that, the biofilm area on each slice is obtained for measurement of biovolume.
The stack of 39 biofilm images in a 15-$\mu$m biofilm is shown in Fig.~\ref{fig:Rodriguez-2007-TDQS}.
Finally, the measurement of the area can be obtained by calculating the heterogeneous growth. 

\subsubsection{Bacteria measurement methods based on edge detection}

In~\cite{Schonholzer-1999-OAFF}, the contrast of images are enhanced and the channel is separated 
into blue and green. 
Then, the Marr-Hildreth filter is used for the detection of objects in green channel. 
After that, the useless debris are removed. 
Finally, the area and fiber length of fungi and bacteria in the gut of earthworm are measured.

\subsubsection{Third-party tools}
In~\cite{Estep-1986-MAUF}, the marine microorganisms are measured by using the `Artek 810' 
image analyzer (Artek Systems Corp.,Farmingdale,N.Y.), and a variable gray-level is set to 
obtain the binary images. 
The area and perimeter are measured to estimate the abundance of marine organisms.

In~\cite{Lawrence-1989-CEDM,Korber-1989-ELFV,An-1995-RQSA,Tackx-1995-MSFE},  the `IBAS' (Kontron Inc., Eching, West Germany) series image processors are applied for biovolume measurement of bacteria.
In~\cite{Lawrence-1989-CEDM,Korber-1989-ELFV}, `IBAS 2000'  image processor is applied to measure the area of darkfield bacteria images~\cite{Lawrence-1989-CEDM},microcolony count, area and size distribution~\cite{Korber-1989-ELFV} by enhancing the cell contrast. 
The moving objects can be highlighted using the motility of bacteria difference images of exercised images with initial images. 
The result shows that computer enhanced microscopy (CEM) is better than direct microbial observation in terms of speed, precision, and quantitative data collection. 
In~\cite{An-1995-RQSA}, the intensity changing is corrected by using a low-pass filter. 
Then,  `IBAS 2000' is used to count the \emph{Staphylococci} and calculate the surface area.  
In~\cite{Tackx-1995-MSFE}, discriminate the area using different threshold after contrast and contour enhancement, and then the `IBAS II' is applied to calculate the total particulate matter area of copepods.

In~\cite{Kuehn-1998-ACLS}, the morphological processor is used to enhance the bacteria image 
and the `Quanitmet 570' computer system (Leica, Cambridge, United Kingdom) is used for 
image processing. Then images are binarized using threshold and the calculation of 
biovolumes is guided by a numerical integration method by following the trapezoidal rule.

In~\cite{Gayen-2008-QCSD}, `Image Pro Plus version 4.5' is used for quantification of bacteria. 
The cluster of bright pixels can be recognized as cells, whose area can be measured  automatically 
by using `Image Pro Plus'. The data is stored as normalized accumulated distribution of cell. 
Finally the cumulative cell distribution is measured. 

\begin{figure}[ht]
\centering
\includegraphics[trim={0cm 0cm 0cm 0cm},clip,width=0.48\textwidth]{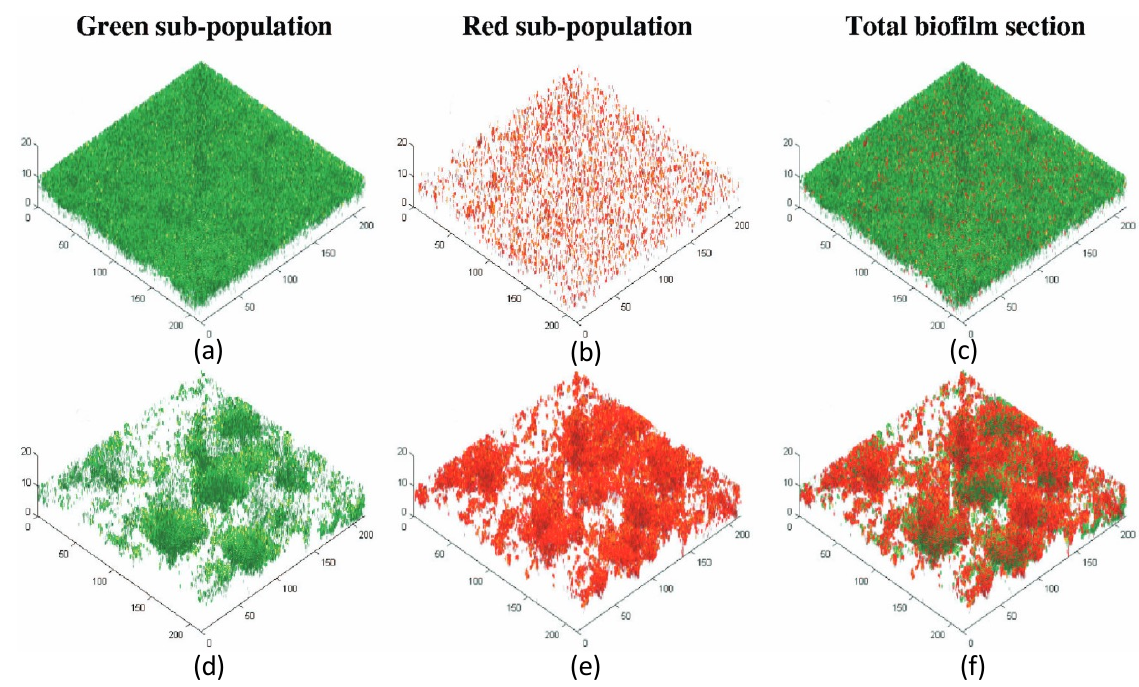}
\caption{Reconstructions of dental plaque biofilms, image (a) to (c) are cultivated on a polystyrene surface, 
and image; (d) to (f) are cultivated on a saliva surface (In~\cite{De-2009-IASB} fig.2)}
\label{fig:De-2009-IASB}
\end{figure}

In~\cite{De-2009-IASB}, `bioImage\_L’ is used to determine the baseline physiology of dental 
plaque grown in a mini-flow cell system. First, the CTC (red) and Syto24 (green) are used to stain 
the biofilm image. Then, a Gaussian low-pass filter and Otsu Thresholding are applied for image 
segmentation. After that, the green channel and red channel are reconstructed in three dimensions 
that is shown in Fig.~\ref{fig:De-2009-IASB}. Finally, the biovolume of dental plaque is measured 
and the results on the viability of dental plaque bacteria indicates that the biovolume of the 
subpopulation of microbes with undamaged cell membranes accounts for 96\% $\pm$ 2\% of 
the total biofilm biovolume.

\begin{figure}[ht]
\centering
\includegraphics[trim={0cm 0cm 0cm 0cm},clip,width=0.48\textwidth]{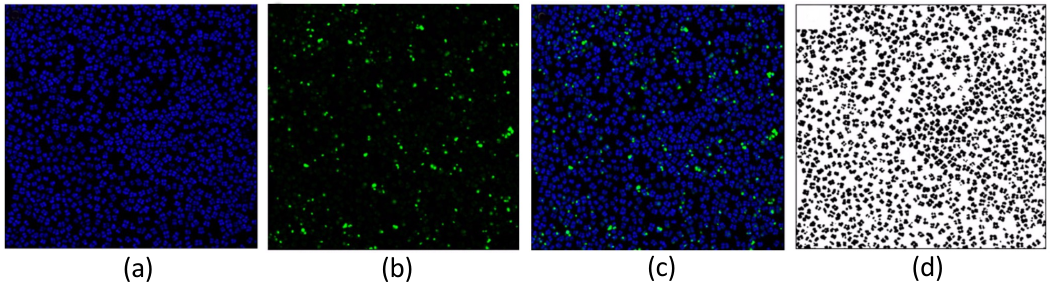}
\caption{The process of image analysis for CLSM images (In~\cite{Puyen-2012-VABM} fig.1)}
\label{fig:Puyen-2012-VABM}
\end{figure}

In~\cite{Puyen-2012-VABM}, `ImageJ' v1.41 (US National Institutes of Health) is used to determine the total biomass and the individual biomass.
The calculation of total biomass is the sum of all living cells biovolume, and the individual biomass is the biovolume of single cell. 
Once the CLSM images are obtained, the ImageJ v1.41 software is applied for green and blue pixel enumeration of each image. Then the images are transformed to binary images using an automatic threshold. 
The final fluorescence image of blue channel is determined by subtracting the green channel 
from initial image, on the contrast, the fluorescence image of green channel is determined by 
subtracting the blue channel from the initial image. Then the image is smoothed by median filtering. 
The ratio of the blue or green threshold voxels to the total threshold voxels can represent the ratio 
of live and dead cells.
Finally, the biovolume of bacteria is measured and multiplied by a conversion factor to convert 
it to biomass. The process of image analysis is shown in Fig.~\ref{fig:Puyen-2012-VABM}.

In~\cite{Ghitua-2013-OTHE}, the `ImageJ' and `CellC' are applied for bacteria counting in 
ballast water. The size and shape of individual cells or the intensity of fluorescently labelled 
microorganisms are measured with quantum dots or fluorochromes. The user can adjust the 
parameters of the modules in software to adapt various experimental environments, then 
the measurement results can be obtained from hundreds of images. 

In~\cite{Folland-2014-ABFC}, the CMEIAS, which is proposed in~\cite{Dazzo-2013-CMEI}, is used for 
biovolume measurement. The shape and morphological features are extracted by using CMEIAS 
to calculate cellular biovolume. The accuracy and error rates are calculated, which contains 
seventeen distinct methods for calculation biovolume by comparing with the ground truth using DIP.
It is proved that this improved technology of biovolume measurement has high accuracy in the 
determination of the selection of algorithm, which is applied for biovolume computation of 
morphologically diverse communities captured by classical microscope at single cell resolution. 

\subsubsection{Summary of image analysis based biovolume measurement for bacteria}

By reviewing the works of image analysis based on bacteria biovolume measurement and referring to Table~\ref{tab:bacteriacounting}, we find that:
\paragraph - Development trend: The development of image analysis methods based on bacteria biovolume measurement started in the 1980s and developed rapidly in the 2010s, showing the increasing importance of bacteria research in human society. 
With the continuous improvement of DIP and pattern recognition, the bacteria biovolume measurement approaches will have appreciable development in the future.
\paragraph - Measurement techniques: The medial filter and contrast enhancement are the most frequently used pre-processing method. 
Image segmentation methods are thresholding and Otsu thresholding.

\begin{table*}[hb]
\scriptsize
\caption{\label{tab:bacteriacounting}Summary of image analysis based biovolume measurement for bacteria} 
\begin{tabular}{p{2cm}p{2cm}p{5cm}p{6.7cm}}
\hline
Related work        & Microorganism & Pre-processing methods& Segmentation methods \\ \hline
\cite{Krambeck-1981-MABD} & Bacteria &  & Width and length marking \\
\cite{Bjornsen-1986-ADBB} & Bacteria & Contour enhancement & Thresholding\\
\cite{Bolter-1993-DBBE} & Bacteria &  & Thresholding\\
\cite{Heydorn-2000-ABST} & Bacteria & Connected-volume filtration & Thresholding\\
\cite{Lomander-2002-AMRA} & Bacteria & Median filter and histogram stretching & Thresholding\\
\cite{Mesquita-2010-DAME} & Bacteria & Morphological parameters measurement & Thresholding\\
\cite{Camp-1992-CATD} & Bacteria &  & RGB histogram segmentation\\
\cite{Schonholzer-1999-OAFF} & Bacteria & Contrast enhancement, morphological operations and thresholding & Marr-Hildreth filter\\
\cite{Petrisor-2004-RACM} & Bacteria & Contrast stretching, histogram equalization and high pass filter & Geographical information systems\\
\cite{Garofano-2005-ATWI} & Bacteria & Thresholding & Wavelet transform and densitometric techniques\\
\cite{Mueller-2006-AAMP} & Bacteria &  & Otsu Thresholding\\
\cite{Rodriguez-2007-TDQS} & Bacteria &  & Stack superficial area measurement\\
\cite{Boman-2008-SDBS} & Bacteria &  & Adaptive thresholding \\
\cite{Ross-2012-EABS} & Bacteria & Connected volume filtration & Otsu Thresholding\\
\cite{Estep-1986-MAUF} & Bacteria &  & Artek 810 image analyzer (Artek Systems Corp.,Farmingdale,N.Y.) and gray-level converting\\
\cite{Lawrence-1989-CEDM} & Bacteria &  & IBAS 2000 (Kontron Inc., Eching, West Germany) \\
\cite{Korber-1989-ELFV} & Bacteria & Contrast enhancement & IBAS 2000 (Kontron Inc., Eching, West Germany) \\
\cite{An-1995-RQSA} & Bacteria & Low pass filter & IBAS 2000 (Kontron Inc., Eching, West Germany)\\
\cite{Tackx-1995-MSFE} & Bacteria & Contour enhancement & IBAS II\\
\cite{Kuehn-1998-ACLS} & Bacteria & Morphological processor & Quanitmet 570 computer system (Leica, Cambridge, United Kingdom) and thresholding \\
\cite{Gayen-2008-QCSD} & Bacteria &  & Image Pro Plus version 4.5 \\
\cite{De-2009-IASB} & Bacteria & Gaussian low-pass filter & BioImage\_L and Otsu Thresholding \\
\cite{Puyen-2012-VABM} & Bacteria & Median filter & ImageJ v1.41 (US National Institutes of Health) \\
\cite{Ghitua-2013-OTHE} & Bacteria &  & ImageJ and CellC \\ 
\cite{Folland-2014-ABFC} & Bacteria &  & CMEIAS \\

\hline
\end{tabular}

\end{table*}

\subsection{Other microorganism measurement methods}
\subsubsection{Measurement methods based on image enhancement}
In~\cite{Congestri-2000-EBBF}, the `Equalize/Best fit' method is used to enhance the contrast of phytoplankton images for obtaining good detection of filaments. 
The measurement of biovolume is determined by the volume calculation of a cylinder, which contains measurement for trichome and section of trichome. 
The diameter represents the minimum diameter and the length represents the maximum diameter.

\subsubsection{Measurement methods based on thresholding}
In~\cite{Morgan-1991-AIAM,Billones-1999-IAAA,Shuxin-2004-MAMS,Ernst-2006-DTFC,Almesjo-2007-AMFC,Cordoba-2010-EIGC}, thresholding approaches are applied for microorganism biovolume measurement.
First, a median filter~\cite{Ernst-2006-DTFC}, contour enhancement and contrast enhancement~\cite{Shuxin-2004-MAMS} are applied for denoising and image enhancement.
Then, the local adaptive background correction~\cite{Ernst-2006-DTFC} is used for spatial threshold changing, which caused by the inhomogeneity of the illumination and fluorescence while imaging. 
After that, a high-pass filter `Black tophat' ~\cite{Almesjo-2007-AMFC} is used to extract black detail from a background with variable intensity.
In~\cite{Morgan-1991-AIAM}, a threshold is used to remove the background and the lines of fungi images are enhanced, then thinning operation is used to keep the hyphae into one-pixel width and calculate the total length. 
The manual method and automated method have obtained very close results.
In~\cite{Billones-1999-IAAA}, a threshold is used to convert grey images into binary images.
The average particle thickness is used to convert the area (two-dimensional) into biovolume (three-dimensional). 
The resulting linear relationships are found to be significant in homogeneous samples like the spheres and diatoms.
In~\cite{Shuxin-2004-MAMS}, thresholding is used to separate the colony from the background and then the features are extracted. 
Then the area and biovolume of colony are calculated. 
The result shows that image processing method can be applied in interface cultivation to measure the growth conditions of colony. 
In~\cite{Ernst-2006-DTFC}, thresholding is used to quantify the plankton and improve the accuracy of measurement. 
The images are automatically segmented into filaments and background by binarization with a single intensity threshold. 
After that, the segmented images are thinned and the total filament length is measured.
The result shows that the filamentous bacteria can be quantified automatically, that has high accuracy and robustness.
In~\cite{Almesjo-2007-AMFC}, a thresholding based method is used for automated measurement of 
filamentous cyanobacteria. 
The holes within the object are filled after thresholding.
Finally, the parameters such as area and perimeter of all particles are measured, which are applied for the calculation of curve length and curve width. By comparing with the manual measurement procedure, the error of filament length calculations is less than 1\%.

In~\cite{Cordoba-2010-EIGC}, digital image processing is used for quantification of alga. 
First, the RGB images of alga are obtained form a CCD camera. 
The color in RGB channel of region of interest (ROI) are segmented and filtered by using two algorithms. 
The colored images are converted to gray-level images in first algorithm and the weights of RGB channels are obtained separately. 
Then the second algorithm is applied for color feature extraction.
Finally, the densities of cell are calculated and the correlation coefficient between the method and manual measurement method is 0.997.

\subsubsection{Measurement methods based on machine learning}
In~\cite{Jung-2003-IALD}, an artificial neural network is applied for algal biovolume measurement.
First, a neighborhood-averaging filtering method is used to remove the noises of algal images. 
The filtered light distribution image is converted to a contour image by grouping the original 8-bit gray scale into a desired value. 
Then, an artificial neural network model is used to relate the cell density in the photobioreactor with the digitized images. 
The present study shows that image analysis techniques can be used to measure light intensity in a photobioreactor, and then the cell concentration in a photobioreactor can be predicted with high accuracy by analyzing the image data using a neural network model.

In~\cite{Alvarez-2012-IPBE}, SVM is used for plankton biovolume estimation. 
First, SVM is applied for plankton classification. Then, three methods are designed for biovolume measurement. 
In the first method, for the spherical shape particles, the projected area of each object is applied for biovolume calculation. 
In the second method, the measurement of biovolume is considered as the calculation of a evolution volume, which is designed manually, and then the width and length are measured for calculation.
In the third method, the particles are classified based on shape features and DIP is applied for biovolume measurement.
The accuracy of proposed method for plankton classification can be improved up to 86\%.

\begin{figure}[ht]
\centering
\includegraphics[trim={0cm 0cm 0cm 0cm},clip,width=0.48\textwidth]{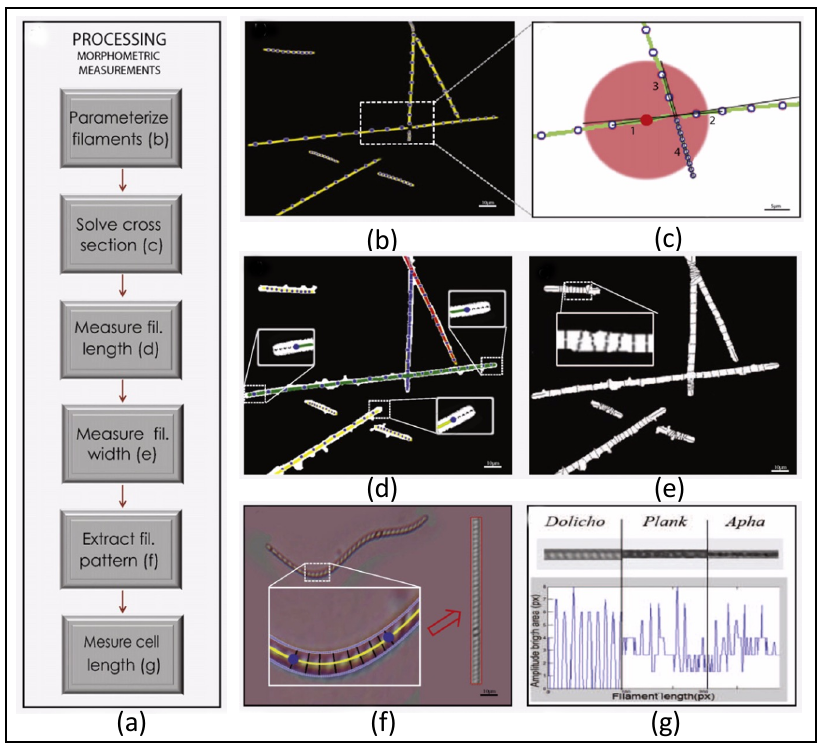}
\caption{Processing algorithm—morphometric measurements (In~\cite{Gandola-2016-ACQU} fig.2)}
\label{fig:Gandola-2016-ACQU}
\end{figure}

In~\cite{Gandola-2016-ACQU}, an automatic cyanobacterial quantification system is developed using pattern recognition and machine learning. 
First, the initial images are denoised and the regions of interest are extracted, which have seven main steps.
First, adaptive brightness and contrast adjustment are applied for exposure minimization. 
Then a Sobel based bidimensional convolution is used for contrast enhancement. 
After that, thresholding is applied to convert images to binary images for enhancement of connected elements in the image. 
Moreover, contours closing and holes filling are applied to extract explicit parts from all closed shapes. 
The noises and spurious objects are erased to reduce computational complexity, and the objects with low roundness value are eliminated based on roundness filter. 
Finally the extracted objects are thinned and pruned for debris elimination.
After image processing, the filamentous objects in the image are processed as several points. 
And the crossing and interrupted filaments are recombined for accurate measurement. 
Afterwards, the parameters of filament are measured and 17 parameters are selected for machine learning based on a Random Forest. The proposed method is shown in Fig.~\ref{fig:Gandola-2016-ACQU}. 

\subsubsection{Measurement methods based on edge detection}
In~\cite{Akiba-2000-DATA}, the higher order local autocorrelational masks are used to extract plankton with their contours. 
The canonical correlation analysis is used to determine the area of plankton after plankton classification.
The result shows that this method can measure the size of all plankton in the image with not contour labelling.

\subsubsection{Measurement methods based on region connection}
\begin{figure}[ht]
\centering
\includegraphics[trim={0cm 0cm 0cm 0cm},clip,width=0.48\textwidth]{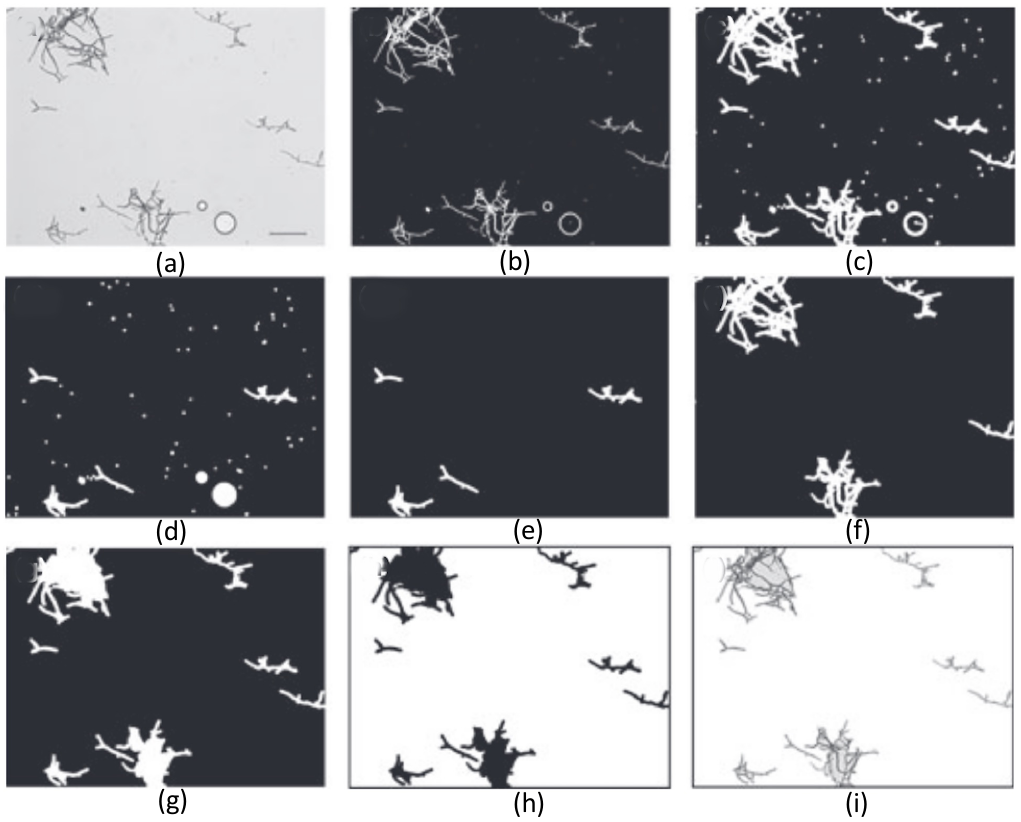}
\caption{The processing steps of image analysis system for area determination of fungi (In~\cite{Lecault-2009-AIAT} fig.2)}
\label{fig:Lecault-2009-AIAT}
\end{figure}

In~\cite{Lecault-2009-AIAT}, image analysis is used to estimate the cell density and biomass 
concentration of fungi.
First, the total difference between background and foreground are minimized to obtain the threshold, that allows satisfactory result while adjusting for variations in background intensity. 
After that, an octagonal filter is applied for region connection, which can help to remain the tiny objects after subsequent denoising operations. 
Then the height and width filter are applied to remove fungi clumps. 
Finally, the 50 measured values of mycelium width in the stochastic selected images are averaged to estimate the cell diameter.
The process of image analysis method is shown in Fig.~\ref{fig:Lecault-2009-AIAT}.

\subsubsection{Measurement methods based on morphological operation}
In~\cite{Albertano-2002-IAQA}, DIP is used for determine the biovolume of cyanobacteria. 
The biovolume of aggregates is calculated by dividing the whole aggregate area with average area in the same image. 
As for curved and coiled filaments, the images are segmented to obtain the straight parts. 
Then the biovolume of images are calculated and added with other measurement results, that is used for determination of total biovolume of cyanobacteria.

In~\cite{Leal-2016-MBEF}, quantitative image analysis is applied for quantification of microorganisms in waste water.
Firstly, the binary images of the biomass of aggregate and the filamentous bacteria are obtained by using DIP, which contains image preprocessing, image segmentation and denoising.
Then, the size of biomass of aggregate and filamentous bacteria are calculated by using this DIP system from the binary image obtained. 
The biomass of aggregate can be calculated by using the projected area, and the filamentous of biomass can be measured by using the total filament length.

\subsubsection{Third-party method}
In~\cite{Alcaraz-2003-EZBI}, `NIH-Image 1.62' (National Institute of Health, Bethesda, Md., USA) 
image analysis system is applied for biomass and biovolume determination of zooplankton. 
The major and minor axis of individual cell are measured and the corresponding  volumes 
of the revolution ellipsoids are calculated. The proposed method can accurately estimate the 
biovolume and number of zooplankton based on DIP.

In~\cite{Miszkiewicz-2004-PPAE}, a digital image analysis software called `MicroImage 4.0' 
(Media Cybernetics for OLYMPUS) is used for fungi measurement. The median and 
high-pass Gauss filter are used to remove noises. Then the images are segmented in HSI 
channel and the projected areas are estimated. Moreover, the weighed mean value of segmented 
region to total area are calculated as the zone fraction.
The result shows that the digital image analysis of \emph{R. oligosporus} hyphae can obtain the 
fast and accurate segmentation results.

In~\cite{Couri-2006-DIPA}, `KS400 software' (Kontron Electronic GMB, Carl Zeiss 
MicroImaging, Inc., Thornwood, NY) is used to measure the biomass growth of fungi. 
A high pass filter is used to obtain the binary images, then the binary filters are used to 
obtain the features of images for biovolume 
measurement. In~\cite{Dutra-2007-LPSS}, `KS400' is used for monitoring the biomass 
growth of fungi. 
First, the binary filters are applied to obtain morphological data. 
After that, a high-pass filter is applied to enhance the contrast of boundaries, and the images is 
binarized by using a fixed threshold. Finally, the debris are pruned and the biovolume are measured. 

In~\cite{Sole-2007-ANMI}, `ImageJ' v1.33f (US National Institutes of Health) is used to determine the biomass of cyanobacteria based on image analysis. 
The threshold is calculated by `ImageJ' that can separate the cyanobacteria clearly and then measure 
the biomass of cyanobacteria.  
It can be observed the significant differences between the manual measurement method and the proposed method ($p$<0.001), with a 95\% confidence interval (2.31, 4.14). 
The results show that the proposed method has higher accuracy when comparing with traditional manual method, particularly when the filamentous of samples have high density. 
In~\cite{Barry-2009-MQFF}, `ImageJ' is used for quantification of filamentous fungal. 
Firstly, the uneven illumination is adjusted by filtering. 
Then, the image is converted to a 2-dimentional signal and the high-pass filtering is applied to correct the background greyscale changes. 
After that, the gray-level threshold is calculated by `ImageJ' and applied to obtain the binary image. 
Then the watershed algorithm is applied to separate the clustered objects and a single binary close operation is applied to erase any tiny crevices of images. 
Finally, the image is skeletonized and measured for determination of biovolume.  
In~\cite{Sole-2009-CLSM}, `ImageJ' 1.37v is applied to determine the biomass of cyanobacteria. 
First, a total of 4103 confocal images corresponding to 156 stacks were obtained. 
Then the threshold is used to transform to binary images and the ratio of the biovolume of object to sediment are calculated. 
To determine the individual biomass, a type of cyanobacteria in the stack is eliminated separately. 
Finally, each stack of biomass is measured.

In~\cite{Posch-2009-NIAT}, the image analysis software `LUCIA G/F' (Laboratory Imaging) is 
used for the measurement of the size and morphological features of microorganisms. There are two 
channels are used to correct possible pixel shift that caused by different excitation wavelengths. 
The first one is designed to detect DAPI stained particles by using UV-excitation, and another one 
is designed to detect hybrid bacteria by using blue excitation. 
Then, the `AND’ operation is applied for combination of binary images obtained by two channels.  
After that, a `Mexican Hat' type edge detection filter is applied and the morphological 
operations like `opening' and `eroding' are applied to decrease the size of particles of processed image. 
Finally, the morphological parameters are measured for determination of biovolume. 

In~\cite{Alum-2009-IABN}, `Scion Image' 4.0.2 is used for quantification of alga. 
Thresholding is used to obtain binary image. After that, the area that covered by algae 
is dividing by the area of total image, which is calculated to represent the percentage area of alga. 
The number of zero pixels can be counted to quantify the area of the specific color channel in the 
binary image. Finally, the total number of chlorophyll is compared with the obtained result to verify 
the accuracy of this system.

In~\cite{Song-2013-RSPA}, `ZooScan Integrated System' is used for quantification of 
zooplankton and the regression between image parameters with biovolume is obtained. 
The segmented image obtained by `ZooScan' system is used for measurement of 46 
different parameters including size, shape, gray level and the fractal dimension. The area 
is calculated with the size of the projection on a horizontal plane while the volume is 
calculated by assuming the shape of zooplankton is spheroid.  

\begin{figure}[ht]
\centering
\includegraphics[trim={0cm 0cm 0cm 0cm},clip,width=0.48\textwidth]{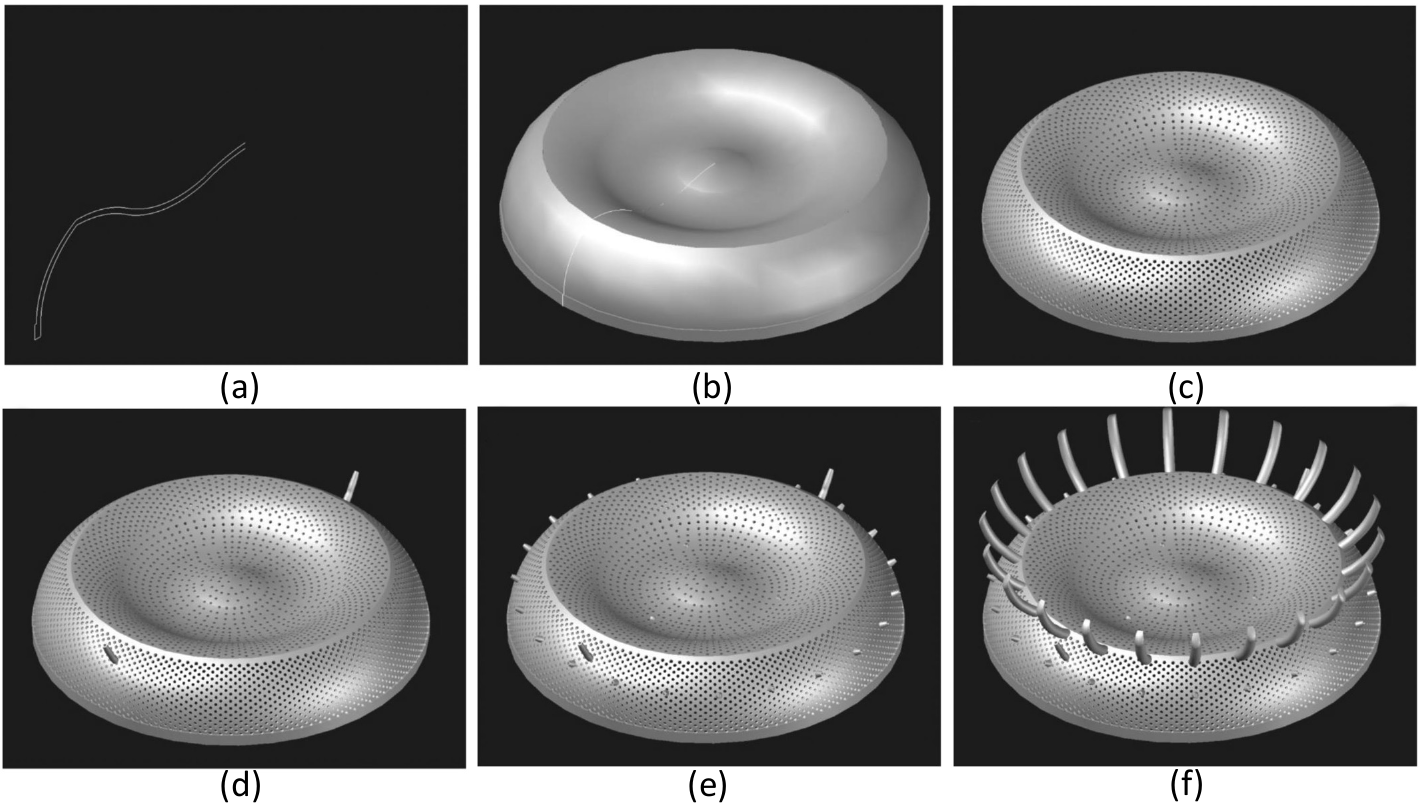}
\caption{The creating process of 3D-model for phytoplankton biovolume measurement (In~\cite{Mohan-2021-OTMP} fig.2)}
\label{fig:Mohan-2021-OTMP}
\end{figure}

In~\cite{Mohan-2021-OTMP}, AutoCAD is applied to create the digital models of phytoplankton, then the parameters including volume and mass of object can be measured.
Firstly, the cross-sectional outline of the phytoplankton images is detected.
Then, the outline is revolved around the central axis to create the diatom disc. 
After that, holes are removed from the disc to meet the real distribution of areolae. 
Finally, other structural components of the phytoplankton are added. 
The creating process of 3D-model is shown in Fig.~\ref{fig:Mohan-2021-OTMP}. 
Moreover, nine representative 3D-models are created in different size.
To measure the biovolume of the phytoplankton, the proposed models are fit with a regression model to get the mathematical parameters of biovolume measurement.

In~\cite{Mcnair-2021-PCAN}, Opera Phenix software is applied for imaging and measuring the biovolume of phytoplankton.
After data acquiring, the images are processed in two methods.
First, a 3D projection is rendered from the discrete confocal images, and the software can measure the biovolume of detected region of cells.
Second, a 2D projection can be converted from the confocal images, which has the same view of a light microscope.
Finally, the biovolume of phytoplankton can be measured.

In~\cite{Borics-2021-BASA}, Blender 2.79 is applied to create the 3-D model of algae and measure the  biovolume. 
After modelling, a software tool, NeuroMorph, is applied to measure the area of the surface by calculating all area of each quadrangulated polygon of the mesh.
Finally, the biovolume can be calculated.

\begin{figure}[ht]
\centering
\includegraphics[trim={0cm 0cm 0cm 0cm},clip,width=0.48\textwidth]{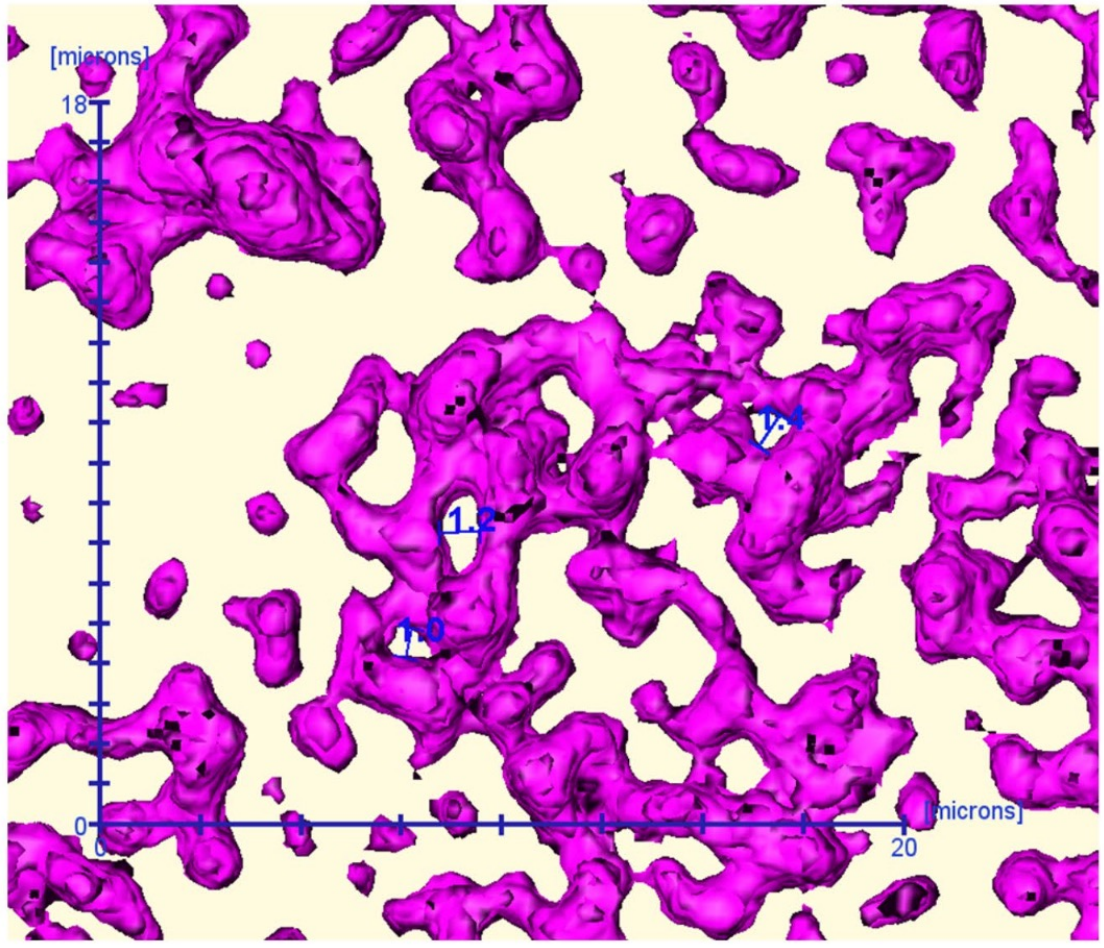}
\caption{The result of determining channel diameter sizes (In~\cite{Awadh-2021-VABQ} fig.5)}
\label{fig:Awadh-2021-VABQ}
\end{figure}

In~\cite{Awadh-2021-VABQ}, Minitab version 16 (Minitab Inc., USA) is applied to determine the biovolume of \emph{Mycoplasma fermentans}.
First, a median filter is used for noise removing in each slice of images. 
Then, the filtered images are segmented based on thresholding to separate the microcolonies and realize the visualisations of 3D iso-surface.
After that, Amira, version 5.4 software (Visualisation Sciences Group, USA), is applied to measure the biovolume of attached biofilm cells. 
Finally, MATLAB (Version R2013b, The MathWorks Inc., USA) is applied to measure the holes between cells.
The measurement of channel diameter sizes is shown in Fig.~\ref{fig:Awadh-2021-VABQ}.

In~\cite{Eniko-2022-UCNE}, the biovolume estimation of cyanobacterial colonies by using NeuroMorf software is compared with the other two methods. 
The first method is the traditional microorganism counting determined by experts.
The second one is based on morphological approach, measuring the sphere packing, cell size and cell distance.
Then, an approach based on regression is applied to calculate the relationship between cell number and the result of biovolume measurement.
The ration between the measured biovolume using estimating and 3D modelling can be applied as the bias of estimation.
Finally, ordinary least square analysis is applied to study the relationship between the colony volumes and overall cell biovolume in the colonies.

\subsubsection{Summary of image analysis based other microorganisms measurement}

By summarizing the relevant research on DIP based on other microorganism's biovolume measurements and referring to Table~\ref{tab:othercounting}, it can be seen that:
\paragraph - Development trend: The DIP-based measurement for other microorganisms methods began in the 1980s and developed in the 2000s. 
It can be seen that the researches on other microorganism biovolume measurement have a similar development trend to related research on bacteria. 
However, bacteria research is relatively abundant and comprehensive, which may be caused by the limited dataset of other microorganisms such as fungi, alga, and viruses. 
On the other hand, compared with the hyphae of alga, bacteria have a relatively simple structure, which makes it challenging to segment completely.
\paragraph - Measurement techniques: The most frequently used pre-processing method are the medial filter and contrast enhancement. Image segmentation methods are thresholding and Otsu Thresholding.

\begin{table*}[hb]
\scriptsize
\caption{\label{tab:othercounting}Summary of image analysis based biovolume measurement for other microorganisms} 
\begin{tabular}{p{2cm}p{2cm}p{5.5cm}p{6.2cm}}

\hline
Related work        & Microorganism  & Pre-processing methods& Segmentation  methods\\ \hline
\cite{Morgan-1991-AIAM} & Fungi & Image enhancement and thinning operation & Thresholding\\
\cite{Billones-1999-IAAA} & Alga &  & Thresholding\\
\cite{Jung-2003-IALD} & Alga & Neighborhood-averaging filter & \\
\cite{Shuxin-2004-MAMS} & Fungi & Contour enhancement & Thresholding\\
\cite{Ernst-2006-DTFC} & Plankton & Median filter and local adaptive background correction & Thresholding\\
\cite{Akiba-2000-DATA} & Plankton &  & Higher order local autocorrelational masks \\
\cite{Albertano-2002-IAQA} & Cyanobacteria &  & Thresholding\\
\cite{Congestri-2000-EBBF} & Phytoplankton & Equalize/Best fit & Thresholding\\
\cite{Almesjo-2007-AMFC} & Cyanobacteria & High-pass filter and holes filling & Thresholding\\
\cite{Lecault-2009-AIAT} & Fungi & Roundness filter, width filter and height filter & Thresholding\\
\cite{Cordoba-2010-EIGC} & ALga &  & Weighted RGB channels\\
\cite{Alvarez-2012-IPBE} & Plankton &  & SVM\\
\cite{Leal-2016-MBEF} & Waste water microorganisms & Debris elimination & Thresholding\\
\cite{Gandola-2016-ACQU} & Cyanobacteria & Adaptive brightness and contrast adjustment, Sobel operation and roundness filter & Random forest \\
\cite{Alcaraz-2003-EZBI} & Zooplankton &  & NIH-Image 1.62  \\
\cite{Miszkiewicz-2004-PPAE} & Fungi & Median and 
high-pass Gauss filter & MicroImage 4.0  \\
\cite{Couri-2006-DIPA} & Fungi & High pass filter & KS400 software  \\
\cite{Dutra-2007-LPSS} & Fungi & High-pass filter and a fixed threshold & KS400 software \\
\cite{Sole-2007-ANMI} & Cyanobacteria &  & ImageJ v1.33f  \\
\cite{Barry-2009-MQFF} & Fungi & High-pass filter & ImageJ v1.33f \\
\cite{Sole-2009-CLSM} & Cyanobacteria &  & ImageJ v1.37v \\
\cite{Posch-2009-NIAT} & Microorganisms & Morphological operations & LUCIA G/F \\
\cite{Alum-2009-IABN} & Alga & Thresholding & Scion Image 4.0.2 \\
\cite{Song-2013-RSPA} & Zooplankton &  & ZooScan Integrated System \\
\cite{Mohan-2021-OTMP} & Phytoplankton &  & AutoCAD \\
\cite{Mcnair-2021-PCAN} & Phytoplankton &  &Opera Phenix \\
\cite{Borics-2021-BASA} & Algae &  &  Blender 2.79 and NeuroMorph software\\
\cite{Awadh-2021-VABQ}&  Mycoplasma & Median filter  & Thresholding, Minitab and Amira \\
\cite{Eniko-2022-UCNE} & Cyanobacterial &  & NeuroMorf software \\

\hline
\end{tabular}
\end{table*}

\section{Quantitative analysis of microorganism biovolume measurement methods with a benchmark dataset}
\begin{figure}[ht]
\centering
\includegraphics[trim={0cm 0cm 0cm 0cm},clip,width=0.48\textwidth]{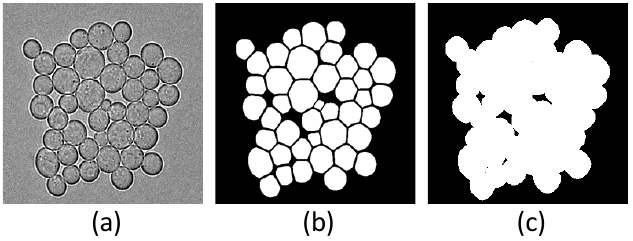}
\caption{(a) The original image; (b) The original mask; (c) Modified mask }
\label{fig:modifieddata}
\end{figure}

In order to compare the performance of the DIP approaches mentioned above for microorganism biovolume measurement, a benchmark database is applied for the experiment of quantitative analysis.
There are few open-source databases of the microorganism with adherent mycelia~\cite{Zhao-2022-EEMI}. 
Therefore, we downloaded and adjusted the yeast database proposed in~\cite{Dietler-2020-ACNN} to meet the requirements of biovolume measurement.
An image with and without a mask and an adjusted mask of the yeast database is shown in Fig.~\ref{fig:modifieddata}.

By reviewing the biovolume measurement approaches proposed in Sec. 3, DIP techniques are commonly applied in image pre-processing and segmentation tasks. 
The outcomes of different approaches can be evaluated systematically, which can be further used to analyze the measurement performance quantitatively.
Moreover, with the development of deep learning, we found that there is no existing work on the deep learning-based biovolume measurement method. 
Hence several deep learning image segmentation methods are also applied for biovolume measurement and compared with classical segmentation methods.

This section is organized as follows: In Section 4.1, several image pre-processing methods, such as filtering and morphological operations,  are compared.
In Section 4.2, several segmentation approaches, which contain classical segmentation and deep learning based segmentation, are applied and evaluated to show their performance in the task of  biovolume measurement. 

\subsection{Quantitative analysis of pre-processing methods}
Digital Image pre-processing methods are applied to deal with the noise and enhance the contrast. 
Gaussian filter is one of the linear filters, which can effectively suppress noise and smooth the image~\cite{De-2009-IASB,Miszkiewicz-2004-PPAE}. 
The process of the Gaussian filter is similar to the mean filter. 
It takes the mean value of pixels in the filter window as the output. 
The mean filter window value is 1, but the value of the Gaussian filter decreases with the increase of distance from the center of the window. 
Therefore, the Gaussian filter performs less fuzzy to the image than a mean filter. 
Another widely applied filter is the median filter, which replaces the value of the center pixel as the median value of the sorted pixel in the filter window~\cite{Lomander-2002-AMRA,Puyen-2012-VABM,Ernst-2006-DTFC}.

\begin{figure}[ht]
\centering
\includegraphics[trim={0cm 0cm 0cm 0cm},clip,width=0.48\textwidth]{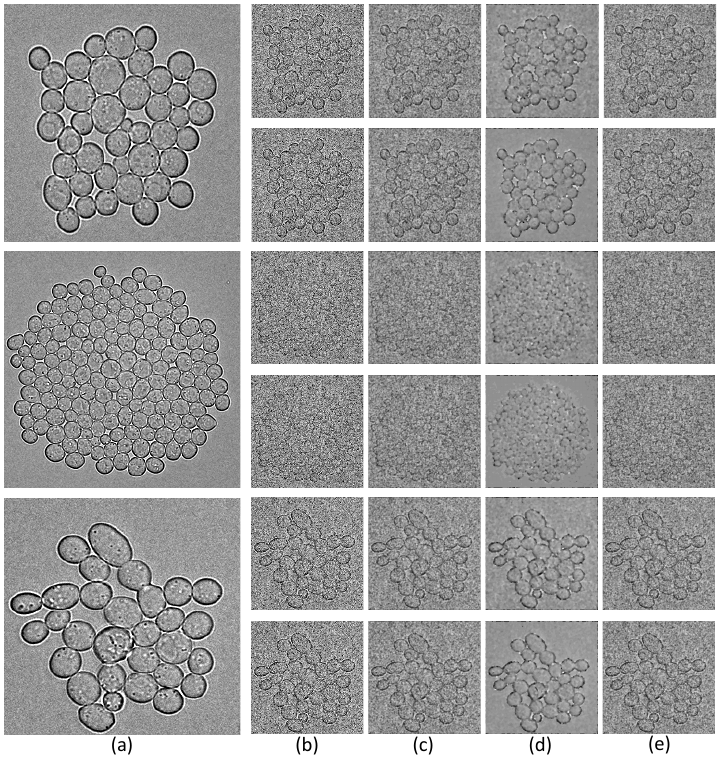}
\caption{(a) Original images; (b) Images after adding noises (upper for salt and pepper noise and below for Gaussian noise); (c) Images after mean filtering (upper for salt and pepper noise and below for Gaussian noise); (d) Images after median filtering (upper for salt and pepper noise and below for Gaussian noise); (e) Images after Gaussian filtering (upper for salt and pepper noise and below for Gaussian noise)}
\label{fig:preprocessing1}
\end{figure}

To show the denoising abilities of three filters above,  the salt and pepper noise with a ratio of 0.3 and Gaussian noise with a mean of 0 and a variance of 0.01 are added to the images, respectively.
Then, the images are denoised with three filters above. 
The images after filtering with 3$\times$3 mean filter, 3$\times$3 median filter and Gaussian filter with the cut-off frequency of 50 are shown in Fig.~\ref{fig:preprocessing1}.
Moreover, mean square error (MSE) and signal noise ratio (SNR)~\cite{Fan-2019-BRID} are applied to evaluate the performance of denoising, which is shown in Table~\ref{table:filterresult}.

\begin{table}[]
\centering
\caption{The MSE and SNR (in dB) of denoised images based on three filters}
\label{table:filterresult}
\begin{tabular}{m{2.5cm}m{2.1cm}m{1cm}m{1cm}}
\hline
Noise   & Filters     & MSE & SNR      \\ \hline
                                     & Mean filter             & 110.56    &    27.71     \\
Salt and pepper           & Median filter             & \textbf{83.98}     & \textbf{28.90}       \\ 
                                      & Gauss filter             & 106.72     &   27.86    \\ 
                                     & Mean filter                  & 110.42    &     27.71      \\ 
Gaussian                     & Median filter             & \textbf{73.11}       &     \textbf{29.52}  \\ 
                                     & Gauss filter             & 106.06     &  27.89      \\   \hline
\end{tabular}
\end{table}

By comparing the denoising results, the images after Median filtering have lowest MSE and highest SNR in both cases, indicating the process of Median filtering can maintain more information and suppress more noises. 
However, the results after Mean filtering and Gauss filtering seem unsatisfactory because of noises and blur contours, and showed low contrast between the objects and background.

\begin{figure}[ht]
\centering
\includegraphics[trim={0cm 0cm 0cm 0cm},clip,width=0.48\textwidth]{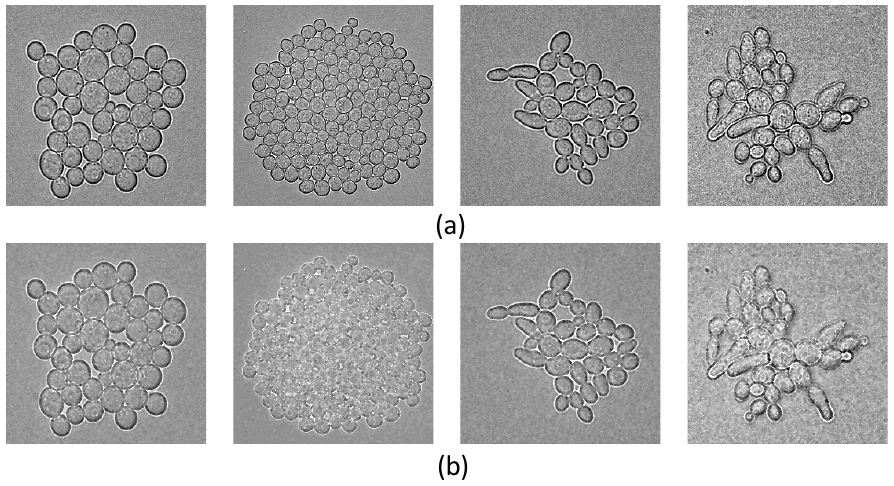}
\caption{(a) Original images; (b) Images after close operation}
\label{fig:preprocessing2}
\end{figure}

Morphological close operation can also be applied for denoising and debris erasing~\cite{Sole-2007-ANMI,Gandola-2016-ACQU}.
The particles smaller than the size of kernel can be eliminated after close operation.
The yeast images after close operation are shown in Fig.~\ref{fig:preprocessing2}.
The noise of images are apparently reduced, but the contours of yeast cells are relatively fuzzy.

\begin{figure}[ht]
\centering
\includegraphics[trim={0cm 0cm 0cm 0cm},clip,width=0.48\textwidth]{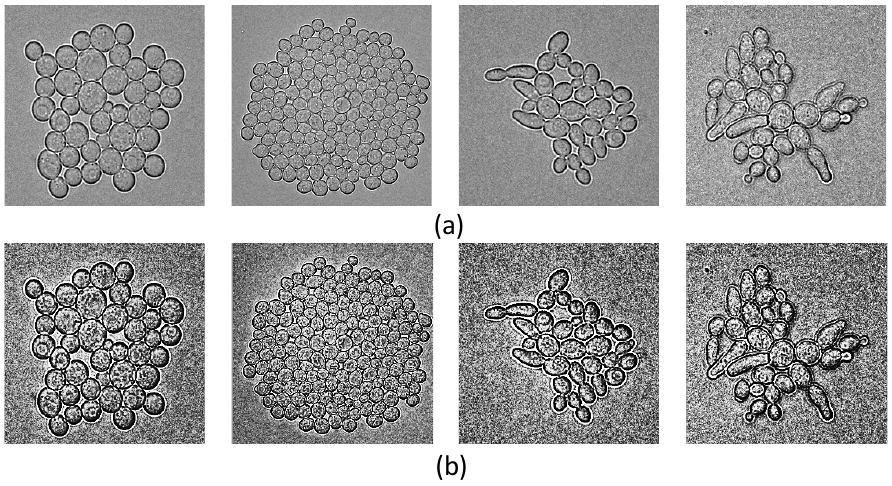}
\caption{(a) Original images; (b) Images after histogram equalization}
\label{fig:preprocessing3}
\end{figure}

Histogram equalization is one of the most used approaches for image enhancement~\cite{Petrisor-2004-RACM,Lomander-2002-AMRA}.
It stretches the histogram of the images as an approximately uniform distribution to enhance the contrast of the images.
Histogram equalization can make the brightness distributes better on the histogram. 
It can enhance the local contrast without affecting the overall contrast, and histogram equalization achieve this function by effectively extending the commonly used brightness.
The yeast images after global histogram equalization are shown in Fig.~\ref{fig:preprocessing3}.
The result is unsatisfactory because the noises spread all around the image, which makes the computer challenging to focus on the foreground and the background.

\subsection{Quantitative analysis of segmentation methods}
\begin{figure}[ht]
\centering
\includegraphics[trim={0cm 0cm 0cm 0cm},clip,width=0.48\textwidth]{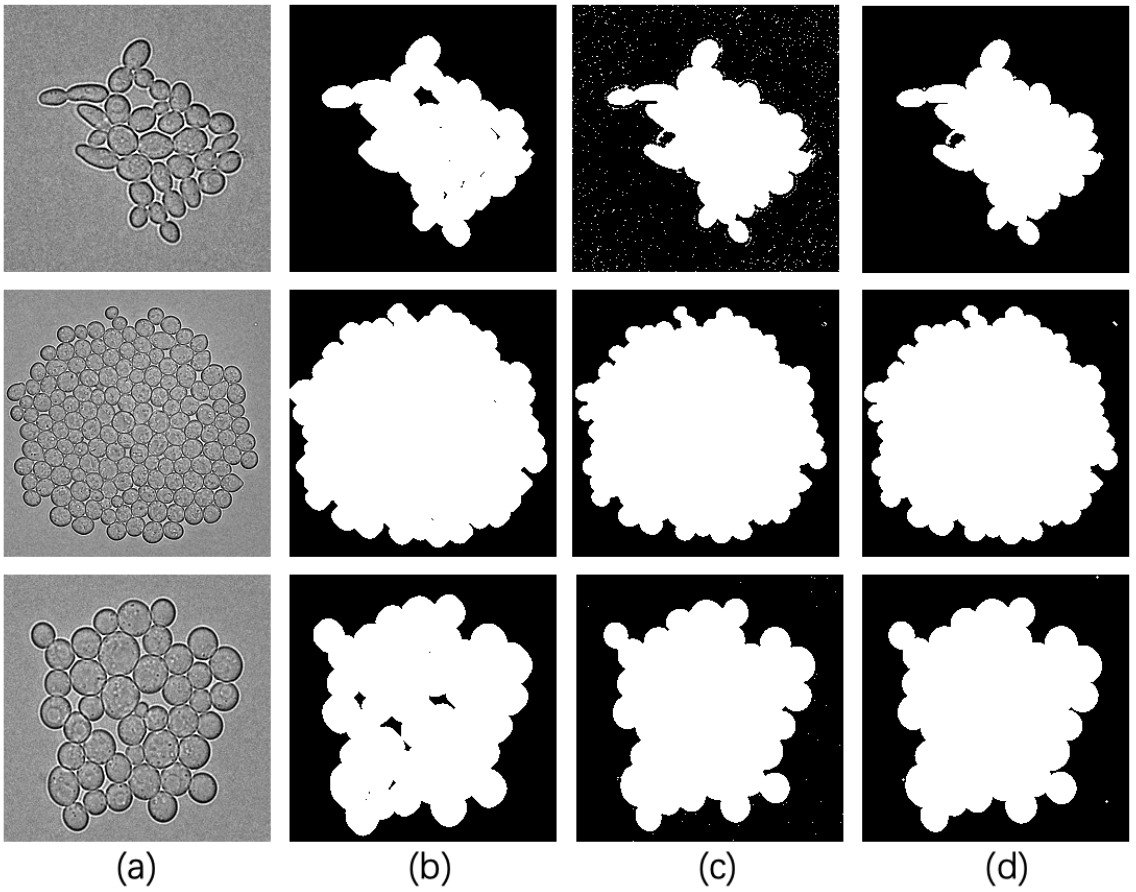}
\caption{(a) Original images; (b) Ground truth; (c) Thresholding based segmentation (value = 115); (d) Otsu Thresholding based segmentation}
\label{fig:segmentation1}
\end{figure}

In the task of image analysis based microorganism biovolume measurement, the segmentation performance determines whether the measurement result is good or not.
The approaches based on thresholding can perform well when the foreground and background are in different range of gray scale, and they can easily be applied due to the small computation cost.
The comparative results of thresholding (value=115) and Otsu Thresholding are shown in Fig.~\ref{fig:segmentation1}.
It can be seen that both results have satisfactory segmentation performance, but the results after thresholding have more noise than the Otsu Thresholding results obviously, which need to be post-processed.

\begin{figure}[ht]
\centering
\includegraphics[trim={0cm 0cm 0cm 0cm},clip,width=0.48\textwidth]{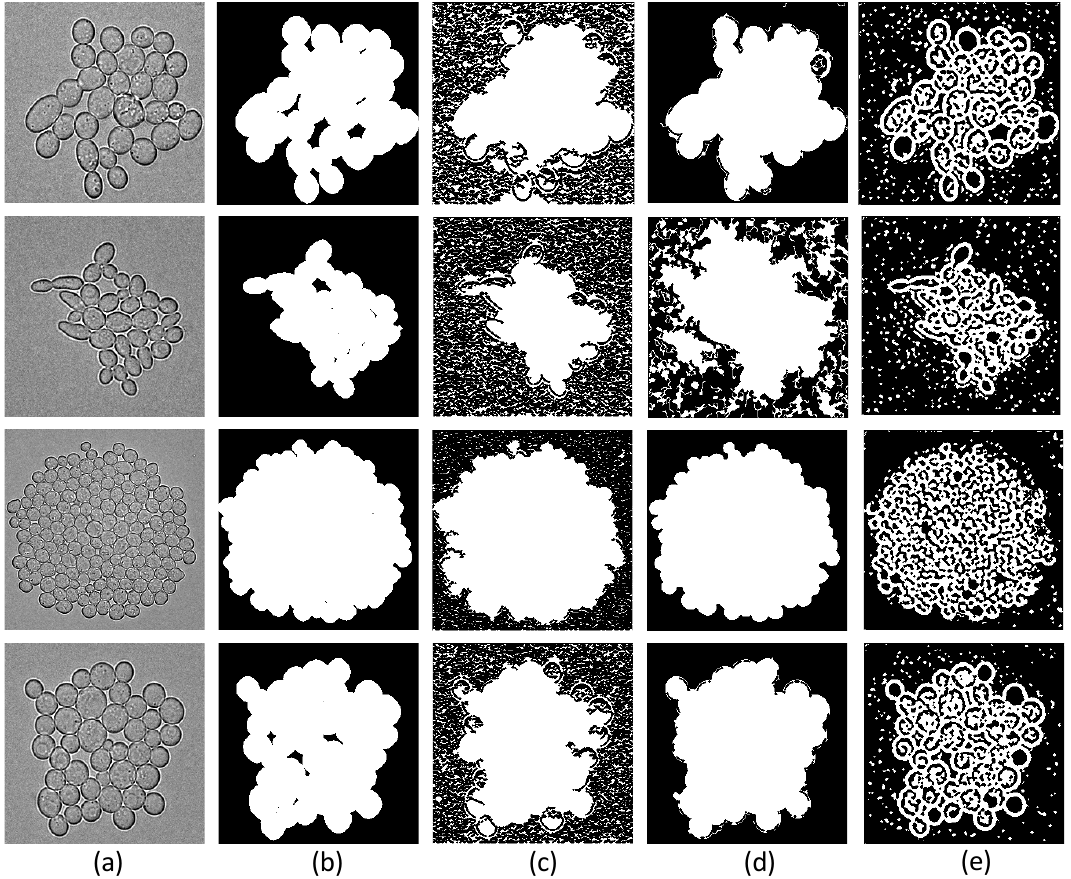}
\caption{(a) Original images; (b) Ground truth; (c) Sobel operator based segmentation; (d) Canny operator based segmentation; (e) Marr-Hildreth operator based segmentation}
\label{fig:segmentation2}
\end{figure}

By reviewing the works of biovolume measurement, edge detection approaches are also widely applied~\cite{Gandola-2016-ACQU,Schonholzer-1999-OAFF,Posch-2009-NIAT}. 
The core algorithm of edge detection is to determine the edge pixels in the image first rather than connect these pixels to form the region boundary. 
Sobel operator is a discrete differentiation operator, which can be applied to calculate the approximate gradient of the grayscale image~\cite{Gandola-2016-ACQU}. 
The pixel with a large gradient is more likely to be the edge. 
The Sobel operator can calculate the derivative in the horizontal and vertical directions, but it is relatively sensitive to the noise. 
Hence, the pre-processing of denoising is necessary. 
The canny operator is a multi-level detection algorithm that contains denoising, gradient calculation, non-maximal suppression, and the method of double thresholding~\cite{Barbedo-2013-AACM}. 
Non-maximum suppression is an edge thinning method to retain the local maximum gradients and suppress the other gradients. 
The canny operator performs well for most of the images. 
In the Marr-Hildreth operator, a Gaussian filter is applied first for smoothing, and then the Laplacian is applied for image enhancement~\cite{Schonholzer-1999-OAFF}. 
Finally, the point of intersection between zero axis and the second derivative line can be determined as the edge.
The comparative results of Sobel, Canny and Marr-Hildreth are shown in Fig.~\ref{fig:segmentation2}.
The segmentation results of Sobel operator have more noise obviously, and the Marr-Hildreth results have the problem of over-detection, which means many inner pixels are classified as edge.
Canny operator performs best by comparing with the other approaches.

\begin{figure}[ht]
\centering
\includegraphics[trim={0cm 0cm 0cm 0cm},clip,width=0.48\textwidth]{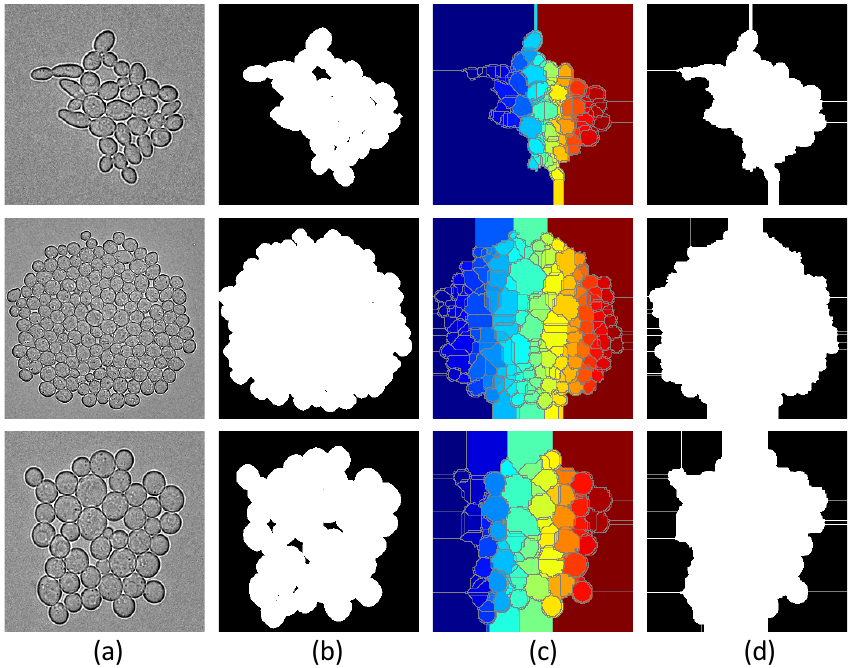}
\caption{(a) Original image; (b) Ground truth; (c) Watershed based segmentation of each region; (d) Watershed based segmentation for foreground and background}
\label{fig:segmentation3}
\end{figure}

Watershed is a segmentation method based on region growing~\cite{Barry-2009-MQFF}.
The gray level of all pixels are sorted and then the growing region is labelled based on local minimum.
The results of watershed is shown in Fig.~\ref{fig:segmentation3}.
It can be seen that the watershed approach has the problem of over-segmentation,hence it may meet the requirements of adherent objects separation, but should not be applied for global segmentation.

\begin{figure}[ht]
\centering
\includegraphics[trim={0cm 0cm 0cm 0cm},clip,width=0.48\textwidth]{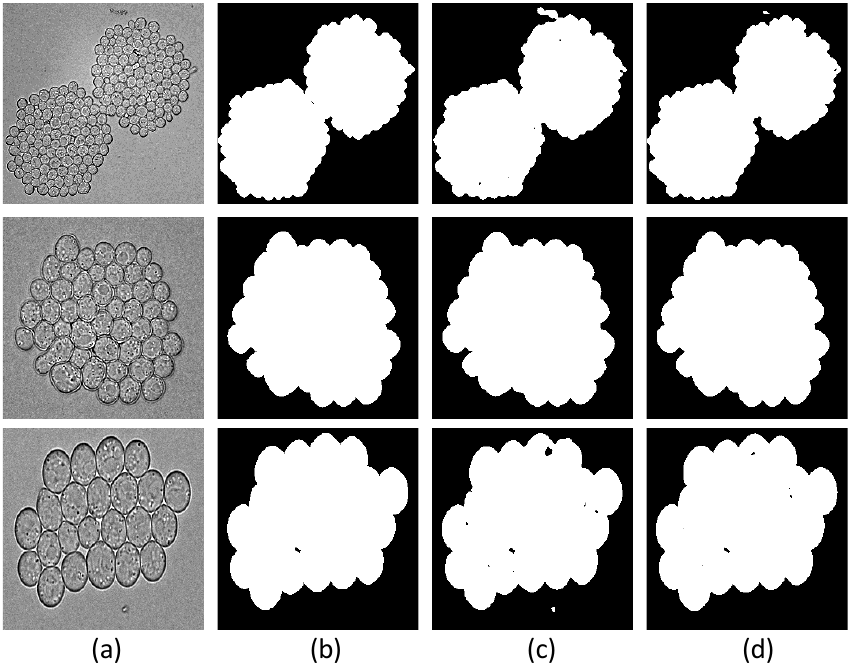}
\caption{(a) Oiginal images; (b) Ground truth; (c) SegNet segmentation; (d) U-Net segmentation}
\label{fig:segmentation4}
\end{figure}

Since the development of deep learning, several excellent backbones (such as SegNet and U-Net) have been widely applied for image segmentation and can obtain satisfactory results~\cite{Zhang-2021-LANL,Rani-2021-MLAD}. 
The first 13-layer convolutional networks of VGG16 are applied in SegNet as an encoder, and SoftMax is applied for classification after the decoder. 
U-Net is an end-to-end net of image segmentation with an encoder and a decoder. 
It was designed for medical image segmentation, which meets the similar requirements of biovolume measurement as both of them are microscopic images, and the datasets are limited. 
SegNet and U-Net train the yeast dataset for 60 epochs with a batch size of 8. 
The CNN-based models can be trained independently without relying on feature engineering, and the models can obtain the expected results, showing strong generalization abilities.
The most-used pre-processing approach in deep learning is data augmentation to avoid overfitting and to increase generalization capabilities, but a limited pre-processing technique is used for denoising~\cite{Garcia-2017-ARDL}. 
Here, SegNet and U-Net are end-to-end models trained without any pre-processing or pre-training. 
The segmentation results of SegNet and U-Net are shown in Fig.~\ref{fig:segmentation4}. 
The results show that U-Net has better performance for biovolume measurement, which means that concatenated operation of encoder and decoder can improve the precision of segmentation.

To quantitative analyze the performance of segmentation approaches above, several classical indices are applied for evaluation, including Accuracy, Dice, Jaccard, Precision and Hausdorff distance. 
The definition of Accuracy, Dice, Jaccard (IoU) and Precision are summarized in Table~\ref{tab:metric}.
Hausdorff distance is applied to measure the Euclidean distance between the predicted and GT images~\cite{Huttenlocher-1993-CIUT}.
Dice is more sensitive to the inner filling of the mask, while Hausdorff Distance is more sensitive to the separated boundary.
The comparative results of different approaches is shown in Table.~\ref{table:finalresult}.

\begin{table}[]
\centering
\caption{The average segmentation evaluation indices of predicted images. A, D, J, P, C, H are abbreviations of Accuracy, Dice, Jaccard, Precision and Hausdorff Distance (in pixels/per image), respectively. }
\label{table:finalresult}
\begin{tabular}{m{2cm}m{0.7cm}m{0.8cm}m{0.8cm}m{0.8cm}m{0.6cm}}
\hline
Methods   & A     & D     & J     & P         & H      \\ \hline
Thresholding   & 92.31             & 87.67              & 80.83             & 93.05                 & 6.14 \\
Otsu        & 90.24             & 89.35              & 83.73             & 90.47                & 5.91 \\ 
Sobel         & 76.34             & 76.55              & 64.43             & 70.21                 & 10.31 \\ 
Canny        & 91.22             & 90.03              & 84.69             & 88.95                & 5.90 \\ 
Marr-Hildreth & 72.17             & 63.71              & 47.49             & 78.07                 & 10.11 \\ 
Watershed & 88.09             & 85.86              & 76.68              & 84.12                 & 8.21 \\ 
SegNet      & 98.64              & \textbf{98.51} & 97.10              & 99.31              & 3.77 \\ 
U-Net        & \textbf{98.97} & 98.43              & \textbf{97.11} & \textbf{99.61} & \textbf{3.39} \\  \hline
\end{tabular}
\end{table}

It is observed that the segmentation performance of deep learning models are much more better than the classical segmentation approaches, obviously.
Canny operator achieves the best result in classical segmentation methods, which is much more better than Sobel and Marr-Hildreth operators. 
U-Net is the best model in the task of microorganism biovolume measurement, indicating that the architecture of encoder-to-decoder and concatenation operation can indeed improve the capability of image segmentation.

\section{Measurement methods analysis and discussion}
The microorganism biovolume measurement methods based on DIP are summarized in Sect. 3. 
By reviewing all the approaches, image pre-processing and segmentation are the most import 
tasks in biovolume measurement.
The performances of these methods are summarized and analyzed, including their advantages 
and potential applications.

\subsection{Image pre-processing methods}

Most microorganisms are mirrored as colorless and have the same color as the environment. 
So, the color feature is not worthy of image segmentation. 
Moreover, due to the inhomogeneity of the illumination and imaging noise, pre-processing approaches can be used for denoising and contrast enhancement.

At first, the images are converted to gray-level images to eliminate the worthless color feature. 
After that, the Intensity of images with Hue-Saturation-Intensity (HSI) can be changed individually, 
that has no influence to color of images, such as the works in~\cite{Cordoba-2010-EIGC},~\cite{Miszkiewicz-2004-PPAE}.

Secondly, the Gaussian low-pass filter and linear transformation can be applied to reduce the influence of uneven illumination, which is caused by uneven shading, such as the work in~\cite{Sole-2007-ANMI}.

\begin{figure}[ht]
\centering
\includegraphics[trim={0cm 0cm 0cm 0cm},clip,width=0.48\textwidth]{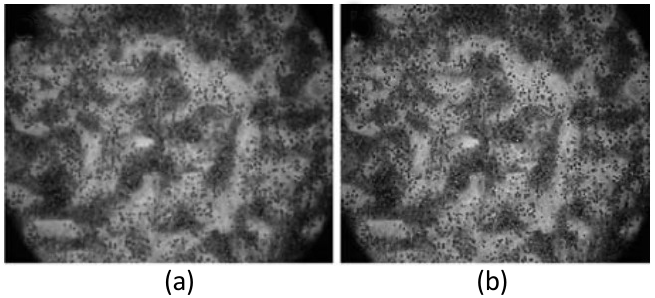}
\caption{(a) The original image; (b) The image after high pass filtering (In~\cite{Dutra-2007-LPSS} fig.1)}
\label{fig:Dutra-2007-LPSS1}
\end{figure}

Thirdly, denoising is necessary for preparation of subsequent image segmentation. 
The median filter and Gaussian filter perform well for noise removal, which are widely applied. 
Gaussian filter is a linear smoothing filter to eliminate Gaussian noise, which is widely applied in image denoising. 
Gaussian filtering denoising is the weighted average of the gray value of the whole image. 
The value of each pixel is calculated by the weighted average of its own value and the value of the neighbor pixels.
Median filter is a non-linear smoothing filter, which replaces the gray value of the pixel as the median gray value of neighbor pixels.
The comparison of Gaussian filter and Median filter is illustrated in Fig.~\ref{fig:preprocessing1}, which shows the Gaussian filter with proper parameters performs best in the task of image denoising.
The application of morphological close operation can remove the useless debris. 
The performance of close operation is shown in Fig.~\ref{fig:preprocessing2}, the noise of image is erased and the contour lines are smoothed, but the edges are relatively fuzzy, which makes it difficult for segmentation.
High-pass filter can also be used for denoising, such as the work in~\cite{Dutra-2007-LPSS}, 
the performance is shown in Fig.~\ref{fig:Dutra-2007-LPSS1}.

\begin{figure}[ht]
\centering
\includegraphics[trim={0cm 0cm 0cm 0cm},clip,width=0.48\textwidth]{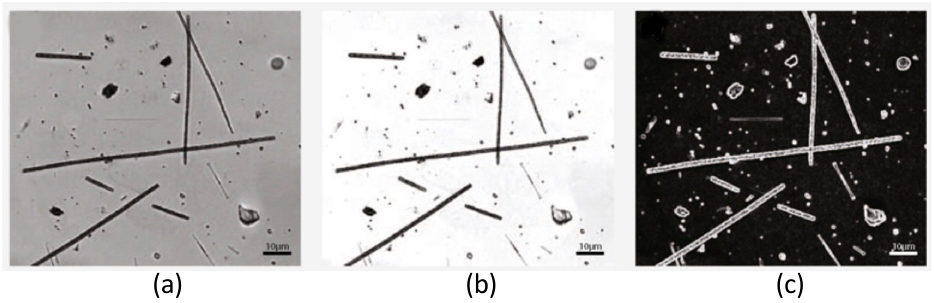}
\caption{(a) The original image; (b) The image after contrast enhancement; (c) The image after using Sobel filter (In~\cite{Gandola-2016-ACQU} fig.1)}
\label{fig:Gandola-2016-ACQU1}
\end{figure}

Finally, the colorless microorganisms have low contrast resolution, which are inconspicuous in environment. 
Contours with high contrast can perform better in image segmentation. 
One of the most frequently-used global methods is gray level histogram equalization, which has satisfactory result in most images, such as the work in~\cite{Petrisor-2004-RACM}.  
Histogram equalization can indeed improve the contrast of the foreground and the background, which is shown in Fig.~\ref{fig:preprocessing2}.
It shows that the contour lines of yeast cells are enhanced and easy to identify, which is helpful for image segmentation.
However, the image noises in the background are enhanced meanwhile, hence the denoising operation is necessary to reduce the influence of noises.
In~\cite{Gandola-2016-ACQU}, contours closing, holes filling and Sobel operation are applied to make solid and unambiguous areas for contour enhancement, the performance is shown in Fig.~\ref{fig:Gandola-2016-ACQU1}.

\subsection{Image segmentation methods}
Image segmentation is the most vital task in microorganism biovolume measurement. 
The colony region needs to be segmented from the whole image for subsequent processing.

At first, thresholding based image segmentation methods are widely used for microorganism biovolume measurement. 
Global thresholding can easily segment the regions of colony after proper pre-processing, such as the works in~\cite{Garofano-2005-ATWI}.
The results of global thresholding on a benchmark dataset are shown in Fig.~\ref{fig:segmentation1}.
By comparing with the ground truth, the contour lines of yeast cells are segmented precisely, but some inner holes are wrongly classified as the foreground.
Hence, the global thresholding can obtain the satisfactory segmentation results in general, but there are still some problems in detail.

\begin{figure}[ht]
\centering
\includegraphics[trim={0cm 0cm 0cm 0cm},clip,width=0.48\textwidth]{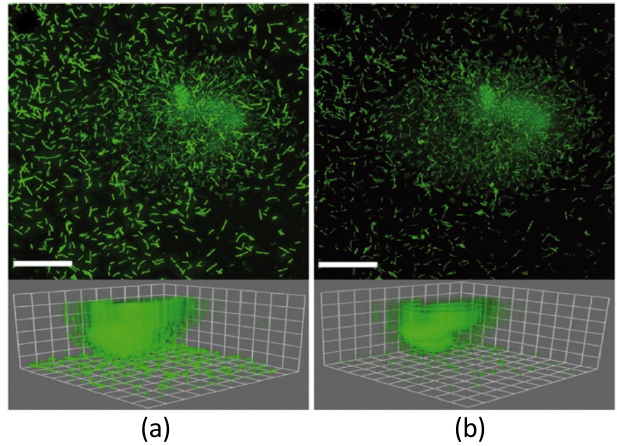}
\caption{(a) The original image; (b) The image after Otsu Thresholding (In~\cite{Ross-2012-EABS} fig. 2)}
\label{fig:Ross-2012-EABS2}
\end{figure}

Secondly, the performance of Otsu Thresholding, such as the works in~\cite{Mueller-2006-AAMP}, ~\cite{Ross-2012-EABS} and \cite{De-2009-IASB}, seems more satisfactory with fast computing speed. 
However, the process of noise removal is necessary before Otsu Thresholding calculation because of the high sensitivity. 
The performance of Otsu Thresholding based image segmentation is shown in Fig.~\ref{fig:Ross-2012-EABS2}.
The results of Otsu Thresholding on a benchmark dataset are shown in Fig.~\ref{fig:segmentation1}.
By comparing with the segmentation results of global thresholding, Otsu Thresholding based segmentation have less noise and the result is more similar to the ground truth, which is proved in Table.~\ref{table:finalresult}, that is, he Hausdorff distance of Otsu Thresholding based segmentation result is less than the result obtained by global thresholding.

Thirdly, the approaches of edge detection are widely spread in microorganism biovolume measurement. 
Sobel edge detection method performs well in microorganism segmentation, such as the work in~\cite{Gandola-2016-ACQU}. 
The combination method of Gaussian filter and Laplacian filter, that is, Marr-Hildreth operator, performs well for image denoising and contour enhancement, such as the work in~\cite{Schonholzer-1999-OAFF}.
In~\cite{Posch-2009-NIAT}, a Mexican hat filter is applied for edge detection, which is the second derivative of the Gauss function. 
The advantage of the Mexican Hat wavelet transform is that it has good localization in the time domain and frequency domain. 
It is symmetric and can be used for the continuous wavelet transform. 
And it is close to the spatial response characteristics of human eyes.
The comparison results based on the proposed edge detection approaches on a benchmark dataset are shown in Fig.~\ref{fig:segmentation2}.
It shows that Sobel operator based edge detection is sensitive to noises, which means lots of noises of the background are recognized as the foreground.
Hence the denoising operation is necessary before Sobel filtering.
The segmentation results of Marr-Hildreth operator exist the problem of over-segmentation.
Many inner pixels are identified as the edge and the results remain a lot of noises.
Canny operator based segmentation performs best, most of the noises are erased and the contour lines are relatively clear and accurate. 
The quantitative analysis in Table.~\ref{table:finalresult} also proves that Canny operator based segmentation has the highest accuracy and the lowest Hausdorff distance, which shows it has the highest similarity with ground truth.

Finally, the adherent colonies need to be separated, which is usually achieved by using watershed algorithm. 
Separation of attached colonies is necessary when the biovolume need to be calculated individually, 
such as the work in~\cite{Barry-2009-MQFF}.
The segmentation results based on watershed is shown in Fig.~\ref{fig:segmentation3}.
Watershed can segment the images into several individual parts and the contour lines can be detected accurately, indicating it can perform well in the task of microorganism counting (meet the requirement of separate the clustered colony into individual cells).
However, the problems of over-segmentation still need to be solved.

\subsection{Image post-processing methods}
\begin{figure}[ht]
\centering
\includegraphics[trim={0cm 0cm 0cm 0cm},clip,width=0.48\textwidth]{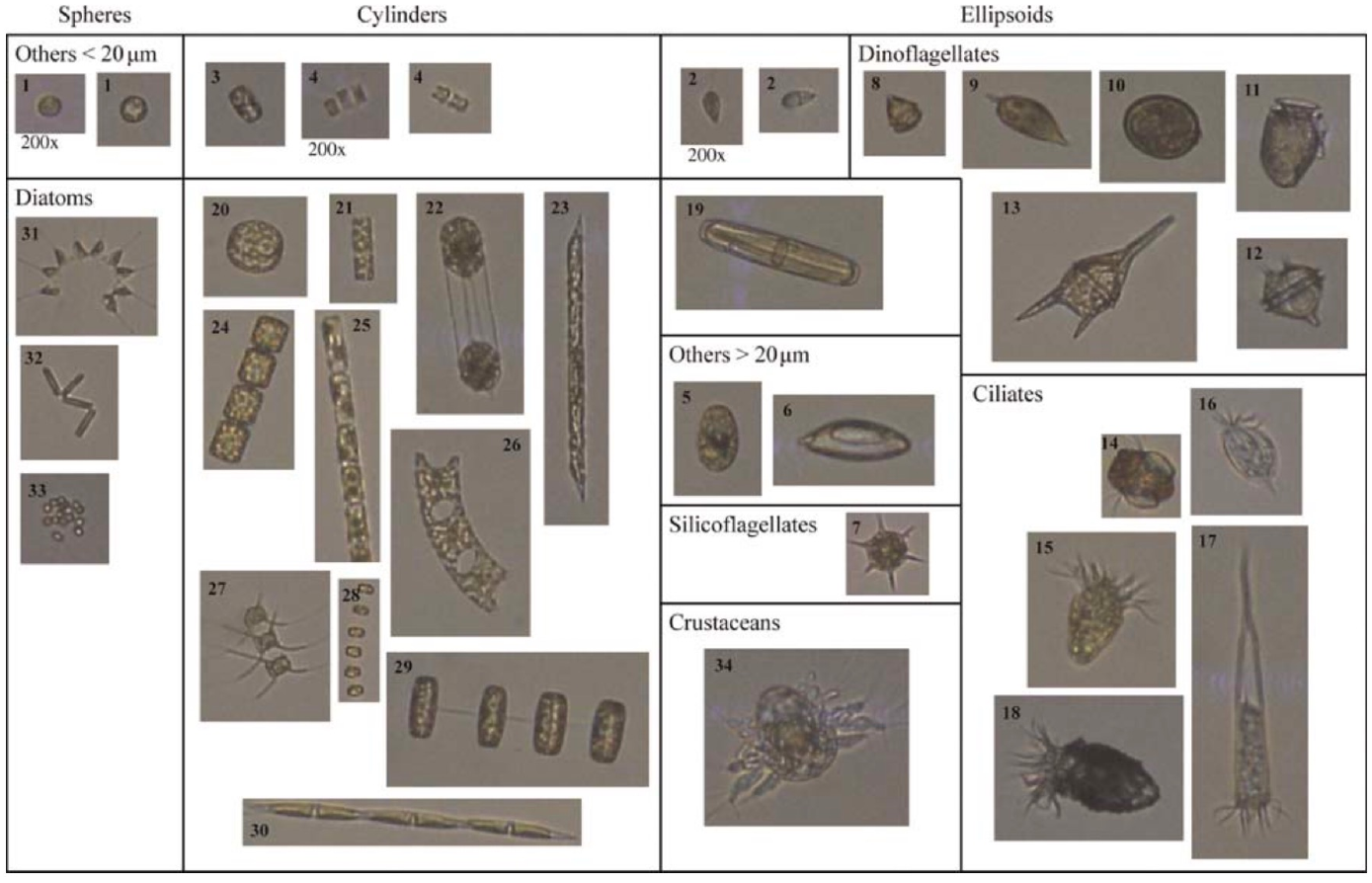}
\caption{The classification groups with their morphological features (In~\cite{Alvarez-2012-IPBE} fig. 2)}
\label{fig:Alvarez-2012-IPBE2}
\end{figure}

Classification is necessary when measuring the biovolume of microorganisms. 
According to supervised learning, the support vector machine (SVM) is one of the classical linear classifiers, such as the work in~\cite{Alvarez-2012-IPBE}.  
The morphological features are obtained for training, and the classification groups are shown in Fig.~\ref{fig:Alvarez-2012-IPBE2}. 
SVM performs well with small samples and can be applied for high-dimensional classification using different kernels.

Secondly, an artificial neural network (ANN) is a self-learning non-linear network with strong fault tolerance.
In~\cite{Jung-2003-IALD}, an ANN is used to relate the cell density in the photobioreactor with the 
digitized images. 
ANN can learn the classification mechanism automatically, which can be applied to obtain the satisfactory results, but the learning process is unobservable and the output is hard to interpret.

\begin{figure}[ht]
\centering
\includegraphics[trim={0cm 0cm 0cm 0cm},clip,width=0.48\textwidth]{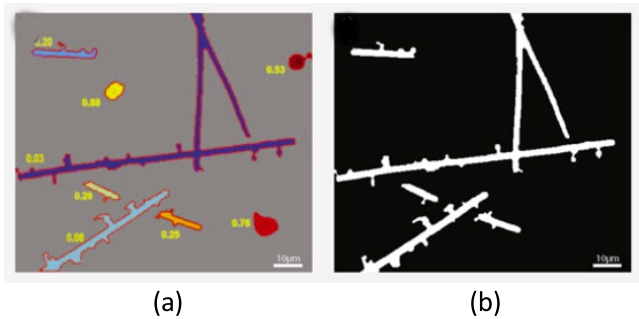}
\caption{(a) Roundness calculation; (b) The round particles are removed using thrsholding (In~\cite{Gandola-2016-ACQU} fig. 2)}
\label{fig:Gandola-2016-ACQU12}
\end{figure}

Thirdly, roundness filter is applied to remove useless debris after segmentation, such as the work 
in~\cite{Lecault-2009-AIAT}. In~\cite{Gandola-2016-ACQU}, the roundness value of remaining elements is estimated to erase all objects with a roundness value > 0.4. When the target particles are circular, the particles with low roundness value should be erased. However, when the target particles are non-circular, such as fungi and alga, the particles with high roundness value should be erased. The performance is shown
in Fig.~\ref{fig:Gandola-2016-ACQU12}. 

\subsection{Analysis of potential methods} 
The microorganism biovolume measurement approaches based on DIP can not only be referred by microbial measurement researches, but also referred by other DIP fields. 
For example, such as the environmental microorganism classification~\cite{Kosov-2018-EMCC,Zhao-2022-ACAD,Kulwa-2021-ANPD,Li-2016-EMAC}, blood cell classification~\cite{Su-2014-ANNA}, classification for different types of microorganisms~\cite{Li-2019-ASTA,Kosov-2018-EMCC,Li-2015-ACIA}, microorganism image generation~\cite{Xu-2020-AEFG}, microorganism object detection~\cite{Li-2021-ASSO}, medical image classification~\cite{Rahaman-2021-DADL,Rahaman-2020-ASCC}, medical object tracking~\cite{Zou-2022-TAEC,Chen-2022-SDAN} and human health monitoring~\cite{Li-2018-CFLM,Shirahama-2016-TLMR,Huang-2020-SSCC}.
After that, the segmentation methods for microorganisms can be referred to by DIP researchers, such as stem cell segmentation~\cite{Huang-2016-SCMI}, near-duplicate detection~\cite{Thyagharajan-2020-ARND}, image enhancement~\cite{Qi-2021-ACOI}, cancer cell segmentation~\cite{Chen-2006-ASCA}, and environmental microorganism segmentation~\cite{Zhang-2021-LANL,Kulwa-2019-ASSM,Zhang-2022-AAPI}. 
Moreover, microscopic image processing performs an essential role in industrial analysis, such as the monitoring for wastewater~\cite{Amaral-2005-ASMA}, beef carcass evaluation~\cite{Cross-1983-BCEU}, monitoring of bacteria in milk~\cite{Pettipher-1982-SACB}, monitoring flames in an industrial boiler~\cite{Yu-2004-MFAI}, softwood lumber grading~\cite{Bharati-2003-SLGO} and so on. 

Microorganism quantification consists of microorganism biovolume measurement and microorganism 
colony counting, and the colony counting can be regarded as a measurement approach with higher accuracy.
Object counting is one of the most essential parts of computer vision, which can be modified and applied 
in microorganism quantification. 

\begin{figure}[ht]
\centering
\includegraphics[trim={0cm 0cm 0cm 0cm},clip,width=0.48\textwidth]{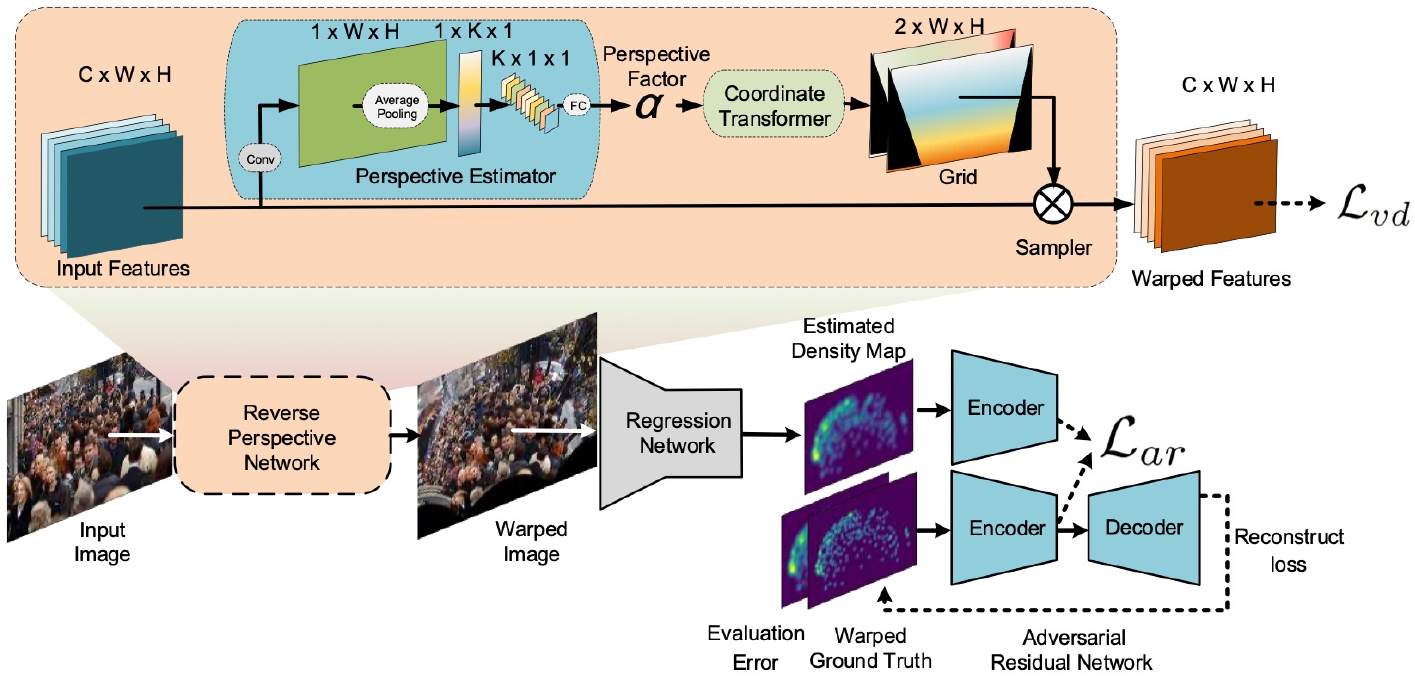}
\caption{The framework of reverse perspective network(In~\cite{Yang-2020-RPNP} fig. 4)}
\label{fig:Yang-2020-RPNP}
\end{figure}

In~\cite{Yang-2020-RPNP}, the reverse perspective network is applied for object counting to solve the problem of input image scale variation. The perspective estimator can calculate the perspective parameter and the coordinate transformer can convert the images to similar size. The weight of ground truth is promoted for training by using an adversarial network. The proposed method can obtain the satisfactory result for the dataset with large variation of scale. The frame work of proposed method is shown in Fig.~\ref{fig:Yang-2020-RPNP}.

\begin{figure}[ht]
\centering
\includegraphics[trim={0cm 0cm 0cm 0cm},clip,width=0.48\textwidth]{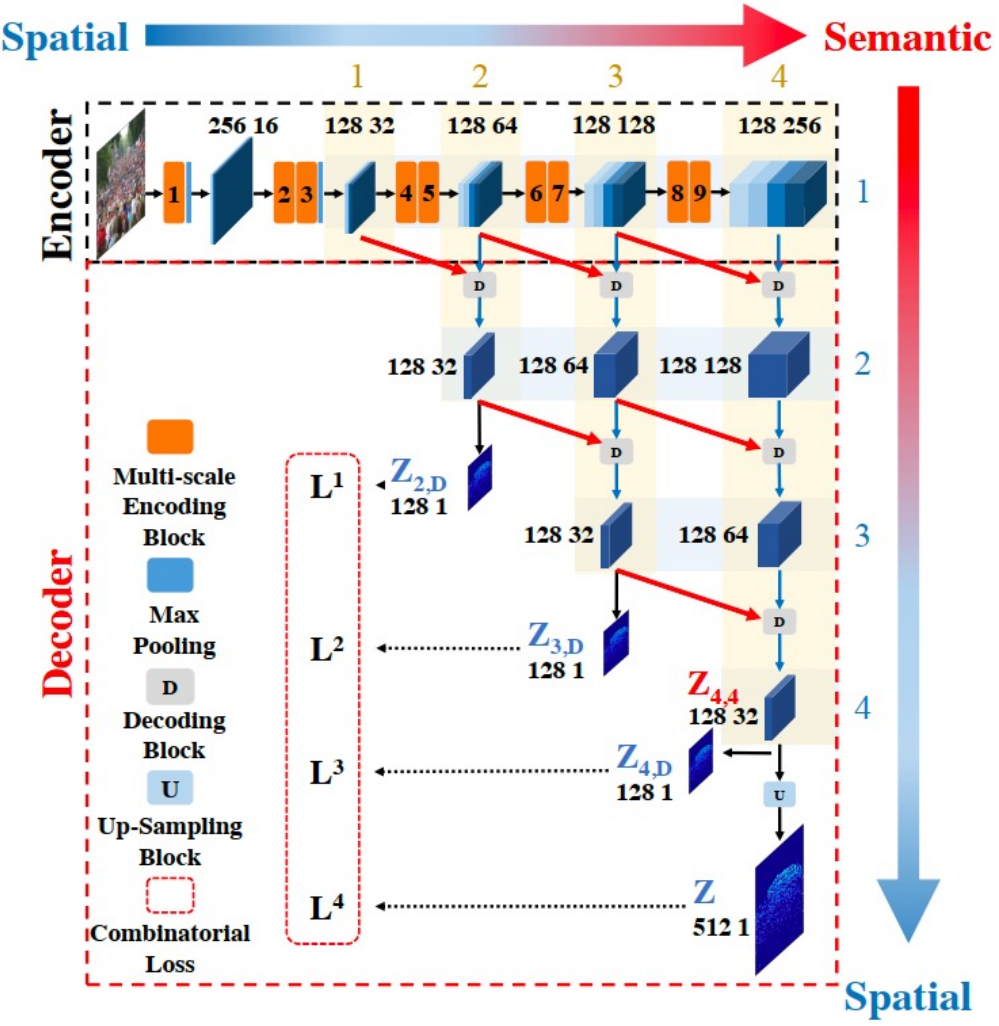}
\caption{The architecture of  Trellis Encoder-Decoder network(In~\cite{Jiang-2019-CCAD} fig. 1)}
\label{fig:Jiang-2019-CCAD}
\end{figure}

In~\cite{Jiang-2019-CCAD}, the traditional detection based methods are replaced by the density estimation based crowd counting methods, which can detect the density of crowd with high performance 
and high precision.  However, the encoder and decoder of traditional CNN based 
density estimation methods have low correlation, which is difficult to remain the spatial information 
of the image. The Trellis Encoder-Decoder networks including multi-path decoder and encoder, is 
proposed to improve the accuracy of counting. The structure of network is shown in Fig.~\ref{fig:Jiang-2019-CCAD}. Only two pooling layers are applied to reduce the loss of spatial pixel accuracy. In addition, a multi-scale encoder is designed to improve the adaptability of network to the large variation of the object size.

\begin{figure}[ht]
\centering
\includegraphics[trim={0cm 0cm 0cm 0cm},clip,width=0.48\textwidth]{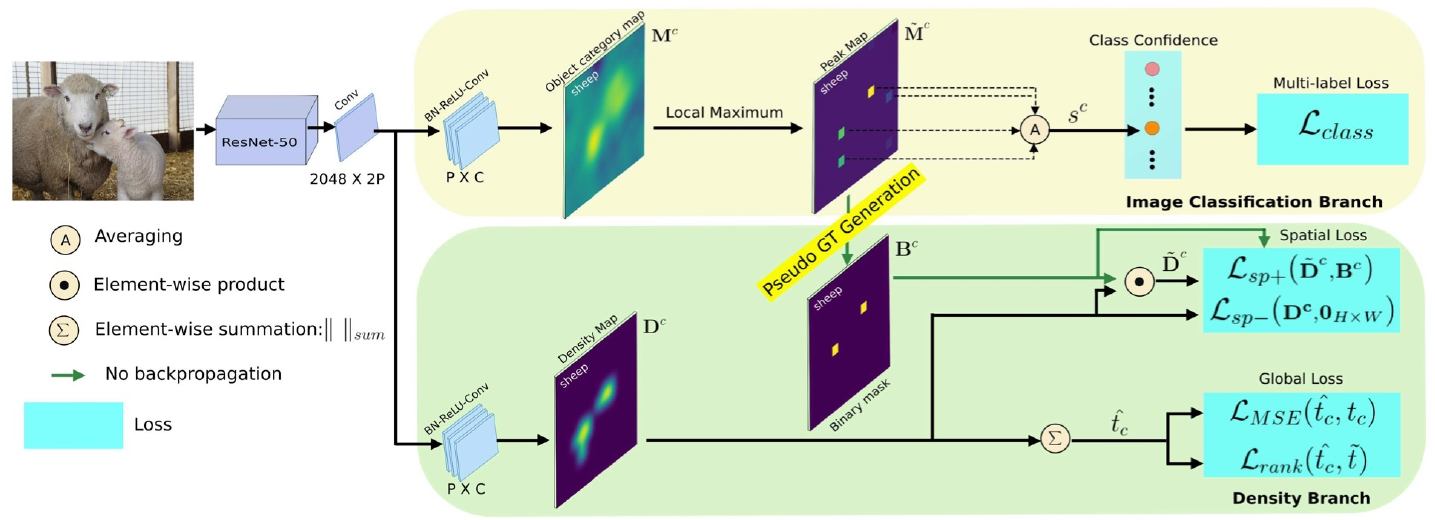}
\caption{The architecture of object counting method(In~\cite{Cholakkal-2019-OCAI} fig. 3)}
\label{fig:Cholakkal-2019-OCAI}
\end{figure}

In~\cite{Cholakkal-2019-OCAI}, an image-leve supervision based image segmentation method is proposed for object counting. The traditional object counting approaches based on regression can obtain the counting results with high accuracy. However, the results can only reflect the global counting but cannot locate the objects that are counted. The Image-level lower-count (ILC) supervised density map estimation is proposed in this article for crowd counting and segmentation. The architecture of the network is shown in Fig.~\ref{fig:Cholakkal-2019-OCAI}. 

\begin{figure}[ht]
\centering
\includegraphics[trim={0cm 0cm 0cm 0cm},clip,width=0.48\textwidth]{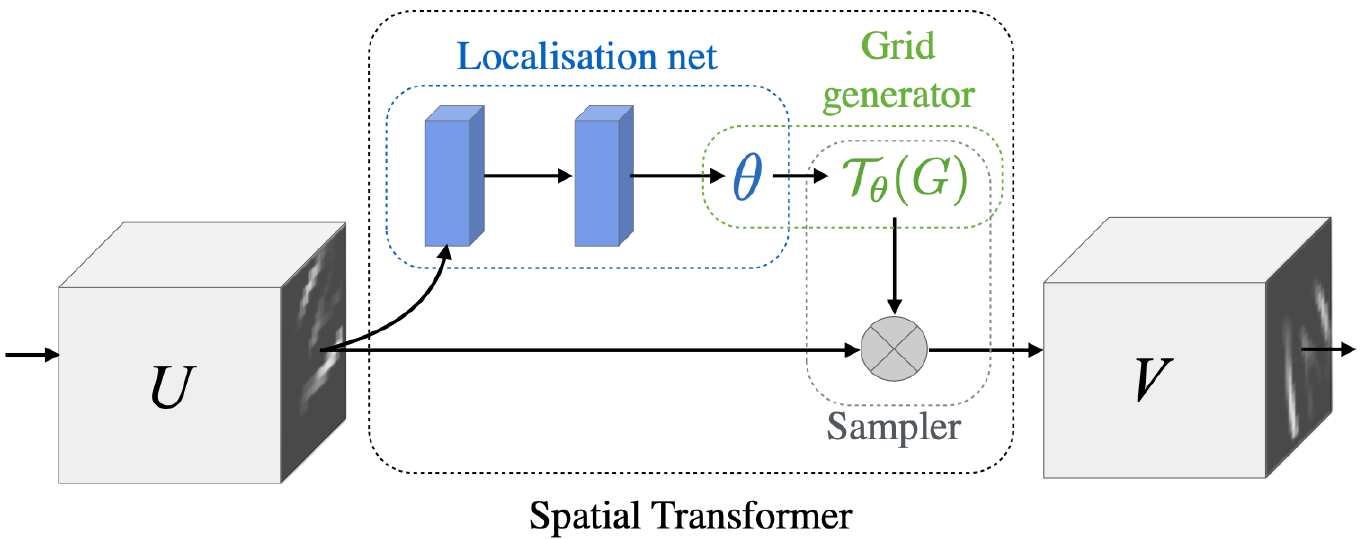}
\caption{The structure of a spatial transformer module (In~\cite{Jaderberg-2015-STNW} fig. 2)}
\label{fig:Jaderberg-2015-STNW}
\end{figure}

In addition, the development of an attention mechanism can be applied in image processing to focus on 
the region of interest (ROI). Spatial transformer networks (STN) can locate the region of interest and transform it into an ideal image by affine transformation, and then put it into the neural network for training~\cite{Jaderberg-2015-STNW}.
STN consists of Localisation net, Grid generator and Sampler. The Localisation net can be applied for calculation of affine transformation parameters. The Grid generator can find the mapping between output and input features,
and the corresponding relation of pixel coordinates is obtained. The Sampler can select input features
based on the position mapping and transformation parameters, and output the image with bilinear interpolation.
The architecture of STN is shown in Fig.~\ref{fig:Jaderberg-2015-STNW}.

\begin{figure}[ht]
\centering
\includegraphics[trim={0cm 0cm 0cm 0cm},clip,width=0.48\textwidth]{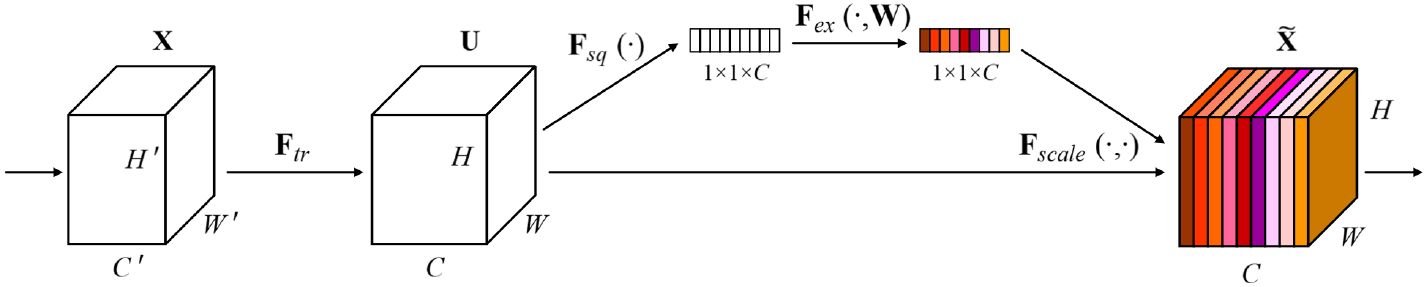}
\caption{The structure of one Squeeze-and-Excitation block in SENET (In~\cite{Hu-2018-SAEN} fig. 1)}
\label{fig:Hu-2018-SAEN}
\end{figure}

Another network based on attention mechanism is Squeeze-and-Excitation Networks (SENET). 
The application of attention mechanism in STN is based on the spatial domain, while SENET is based on 
the channel domain. SENET can learn the correlation of each feature channel, and the weight value of 
each channel is calculated, so that the following network can improve the selected significant 
feature channels and the unworthy feature channels are restrained.
The structure of SENET is shown in Fig.~\ref{fig:Hu-2018-SAEN}. 
The satisfactory ROI extraction method provided by the attention mechanism can support a possibility 
to extract the target microorganism region in an image with a complex environment.

\begin{figure}[ht]
\centering
\includegraphics[trim={0cm 0cm 0cm 0cm},clip,width=0.48\textwidth]{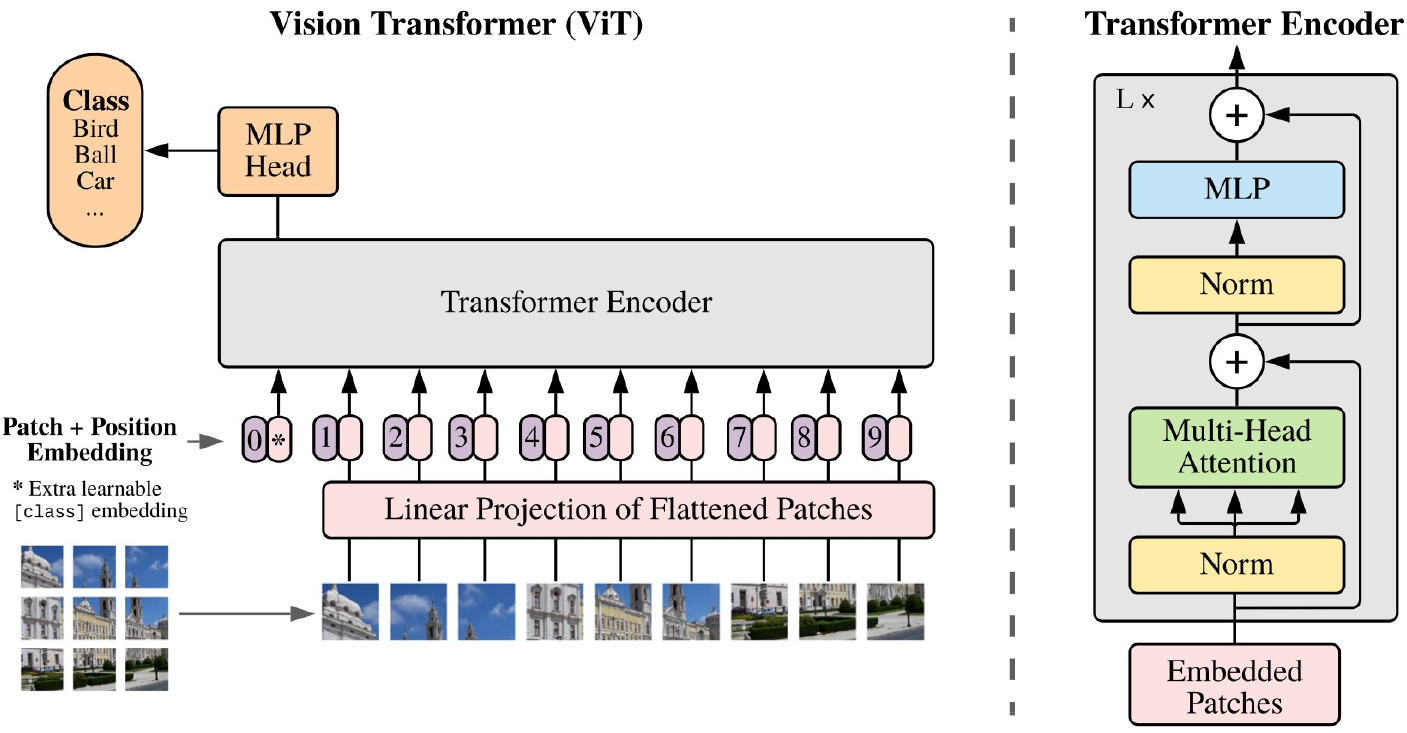}
\caption{The structure of the ViT(In~\cite{Dosovitskiy-2020-AIWW} fig. 1)}
\label{fig:Dosovitskiy-2020-AIWW}
\end{figure}

Furthermore, Transformer is a type of self-attention mechanism-based deep neural network that is 
wildly used in natural language processing (NLP). A transformer has been developed to satisfy 
computer vision because of the strong representation ability recently. The use of Transformer performs 
better by comparing with other networks, such as CNN, which has shown strong competitiveness~\cite{Han-2020-ASVT}. A transformer has a more robust ability of global information representation than CNNs, making it possible to describe the microorganism structure in a complete image. Especially, Vision 
Transform (ViT) is one of the most remarkable visual transformer method till now, which directly applies 
sequences of image patches (with position information) as input first. The ViT projects the patches 
to the original transformer encoder and classifies the images with a multi-head attention 
mechanism as it does in NLP tasks.
The framework of ViT is shown in Fig.~\ref{fig:Dosovitskiy-2020-AIWW}.

Finally, the development of microorganism biovolume measurement based on DIP has high correlation with 
the development of relevant researches,
such as the technique of immunofluorescence staining for cells~\cite{Lee-2020-MISA}, 
confocal laser scanning microscope (CLSM) imaging~\cite{Bloem-1995-FADS}, epifluorescence 
microscope equipment~\cite{Pernthaler-2003-AEGM}, and so on.

\section{Conclusion}
In this paper, a comprehensive review of DIP based microorganism biovolume measurement is proposed. 
The current methods are generalized and summarized, and grouped as bacteria and other microorganisms, such as fungi, alga and so on.
In each type of microorganism summarization, the methods are grouped based on image segmentation algorithm, including classical thresholding methods, self-designed methods, and third-party tools. 
In Sect. 3.1.1 and Sect. 3.2.1, it can be seen that the classical approaches are developed first, from 1980s to 2000s. By reviewing the classical methods, the performances are limited and unsatisfied.
In Sect. 3.1.1 and Sect. 3.2.1, the development of self-designed methods for biovolume measurement is rapid. Self-designed methods can process the microorganism data by using specific approaches to match their different features, which show satisfactory result by comparing with the classical methods.
In Sect. 3.1.3 and Sect. 3.2.3, a mass of professional DIP systems are applied for microorganism biovolume measurement, such as `ImageJ' and `CellC'. 
The integration software are more professional and performs well in specific field of biovolume measurement, which shows the microorganism biovolume measurement is increasing important.
In conclusion, the prosperous development of DIP based microorganism biovolume measurement methods show the extensive research potential in microbial research. 
After that, in Sect.4, the most used image pre-processing and segmentation approaches are quantitatively analyzed to show their performance in the task of microorganism biovolume measurement.
Finally, the widely applied microorganism biovolume measurement approaches such as image pre-processing, image segmentation, and connected domain detection are analyzed in Sect. 5.

\section*{Acknowledgements}
This work is supported by the ``Natural Science Foundation
of China'' (No. 61806047), ``Sichuan Science and Technology Program'' (No. 2021YFH0069, 2021YFQ0057, 2022YFS0565). 
We thank Miss Zixian Li and Mr. Guoxian Li for their important discussion.

{\small
\bibliographystyle{splncs04}
\bibliography{Jiawei}

\begin{thebibliography}{100}
\providecommand{\url}[1]{\texttt{#1}}
\providecommand{\urlprefix}{URL }
\providecommand{\doi}[1]{https://doi.org/#1}

\bibitem{Akiba-2000-DATA}
Akiba, T., Kakui, Y.: {Design and testing of an underwater microscope and image
  processing system for the study of zooplankton distribution}. IEEE journal of
  oceanic engineering  \textbf{25}(1),  97--104 (2000)

\bibitem{Albertano-2002-IAQA}
Albertano, P.: {Image analysis for qualitative and quantitative evaluation of
  planktic cyanobacteria}. RAPPORTI ISTISAN  \textbf{2}(9),  64--69 (2002)

\bibitem{Alcaraz-2003-EZBI}
Alcaraz, M., Saiz, E., Calbet, A., Trepat, I., Broglio, E.: {Estimating
  zooplankton biomass through image analysis}. Marine Biology  \textbf{143}(2),
   307--315 (2003)

\bibitem{Almesjo-2007-AMFC}
Almesj{\"o}, L., Rolff, C.: {Automated measurements of filamentous
  cyanobacteria by digital image analysis}. Limnology and Oceanography: Methods
   \textbf{5}(7),  217--224 (2007)

\bibitem{Alum-2009-IABN}
Alum, A., Mobasher, B., Rashid, A., Abbaszadegan, M.: {Image analyses-based
  nondisruptive method to quantify algal growth on concrete surfaces}. Journal
  of Environmental Engineering  \textbf{135}(3),  185--190 (2009)

\bibitem{Alvarez-2012-IPBE}
{\'A}lvarez, E., Lopez-Urrutia, A., Nogueira, E.: {Improvement of plankton
  biovolume estimates derived from image-based automatic sampling devices:
  application to FlowCAM}. Journal of Plankton Research  \textbf{34}(6),
  454--469 (2012)

\bibitem{Amaral-2005-ASMA}
Amaral, A., Ferreira, E.: {Activated sludge monitoring of a wastewater
  treatment plant using image analysis and partial least squares regression}.
  Analytica Chimica Acta  \textbf{544}(1-2),  246--253 (2005)

\bibitem{Amrita-2016-IPTA}
Amrita, Kaur, L.: {Image Processing Techniques on Agriculture-A Review}.
  Research Cell: An International Journal of Engineering Sciences
  \textbf{22}(0),  515--526 (2016)

\bibitem{An-1995-RQSA}
An, Y.H., Friedman, R.J., Draughn, R.A., Smith, E.A., Nicholson, J.H., John,
  J.F.: {Rapid quantification of staphylococci adhered to titanium surfaces
  using image analyzed epifluorescence microscopy}. Journal of Microbiological
  Methods  \textbf{24}(1),  29--40 (1995)

\bibitem{Andersen-2020-TPOS}
Andersen, K.G., Rambaut, A., Lipkin, W.I., Holmes, E.C., Garry, R.F.: {The
  proximal origin of SARS-CoV-2}. Nature medicine  \textbf{26}(4),  450--452
  (2020)

\bibitem{Andreini-2016-AICT}
Andreini, P., Bonechi, S., Bianchini, M., Garzelli, A., Mecocci, A.: {Automatic
  image classification for the urinoculture screening}. Computers in biology
  and medicine  \textbf{70},  12--22 (2016)

\bibitem{Anuar-2010-VCPU}
Anuar, N., Sultan, A.B.M.: {Validate conference paper using dice coefficient}.
  Computer and Information Science  \textbf{3}(3), ~139 (2010)

\bibitem{Awadh-2021-VABQ}
Awadh, A.A., Kelly, A.F., Forster-Wilkins, G., Wertheim, D., Giddens, R.,
  Gould, S.W., Fielder, M.D.: {Visualisation and biovolume quantification in
  the characterisation of biofilm formation in Mycoplasma fermentans}.
  Scientific reports  \textbf{11}(1), ~1--9 (2021)

\bibitem{Barbedo-2013-AACM}
Barbedo, J.G.A.: {An algorithm for counting microorganisms in digital images}.
  IEEE Latin America Transactions  \textbf{11}(6),  1353--1358 (2013)

\bibitem{Barry-2009-MQFF}
Barry, D.J., Chan, C., Williams, G.A.: {Morphological quantification of
  filamentous fungal development using membrane immobilization and automatic
  image analysis}. Journal of Industrial Microbiology and Biotechnology
  \textbf{36}(6), ~787 (2009)

\bibitem{Bharati-2003-SLGO}
Bharati, M., MacGregor, J., Tropper, W.: {Softwood lumber grading through
  on-line multivariate image analysis techniques}. Industrial \& Engineering
  Chemistry Research  \textbf{42}(21),  5345--5353 (2003)

\bibitem{Billones-1999-IAAA}
Billones, R., Tackx, M., Flachier, A., Zhu, L., Daro, M.: {Image analysis as a
  tool for measuring particulate matter concentrations and gut content, body
  size, and clearance rates of estuarine copepods: validation and application}.
  Journal of Marine Systems  \textbf{22}(2-3),  179--194 (1999)

\bibitem{Bjornsen-1986-ADBB}
Bj{\o}rnsen, P.K.: {Automatic determination of bacterioplankton biomass by
  image analysis}. Applied and Environmental Microbiology  \textbf{51}(6),
  1199--1204 (1986)

\bibitem{Bloem-1995-FADS}
Bloem, J., Veninga, M., Shepherd, J.: {Fully automatic determination of soil
  bacterium numbers, cell volumes, and frequencies of dividing cells by
  confocal laser scanning microscopy and image analysis}. Applied and
  Environmental Microbiology  \textbf{61}(3),  926--936 (1995)

\bibitem{Bolter-2002-EABD}
B{\"o}lter, M., Bloem, J., Meiners, K., M{\"o}ller, R.: {Enumeration and
  biovolume determination of microbial cells--a methodological review and
  recommendations for applications in ecological research}. Biology and
  Fertility of Soils  \textbf{36}(4),  249--259 (2002)

\bibitem{Bolter-1993-DBBE}
B{\"o}lter, M., M{\"o}ller, R., Dzomla, W.: {Determination of bacterial
  biovolume with epifluorescence microscopy: comparison of size distributions
  from image analysis and size classifications}. Micron  \textbf{24}(1),
  31--40 (1993)

\bibitem{Boman-2008-SDBS}
Boman, C., Jialin, Y., Guanxin, L., Linyan, H., Fang, L., Hua, Y.: {Sequential
  development of biofilm spatial structure of Pseudomonas Aeruginosa tagged
  with SYTO9/PI}. Journal of third military medical university  \textbf{30}(5),
   390--392 (2008)

\bibitem{Borics-2021-BASA}
Borics, G., Lerf, V., Enik{\H{o}}, T., Stankovi{\'c}, I., Pick{\'o}, L.,
  B{\'e}res, V., V{\'a}rb{\'\i}r{\'o}, G., et~al.: {Biovolume and surface area
  calculations for microalgae, using realistic 3D models}. Science of The Total
  Environment  \textbf{773},  145538 (2021)

\bibitem{Buckland-1994-TRBR}
Buckland, M., Gey, F.: {The relationship between recall and precision}. Journal
  of the American society for information science  \textbf{45}(1),  12--19
  (1994)

\bibitem{Camp-1992-CATD}
Camp, C.E., Sublette, K.L.: {Control of a Thiobacillus denitrificans bioreactor
  using machine vision}. Biotechnology and bioengineering  \textbf{39}(5),
  529--538 (1992)

\bibitem{Chen-2022-SDAN}
Chen, A., Li, C., Zou, S., Rahaman, M.M., Yao, Y., Chen, H., Yang, H., Zhao,
  P., Hu, W., Liu, W., et~al.: {SVIA dataset: A new dataset of microscopic
  videos and images for computer-aided sperm analysis}. Biocybernetics and
  Biomedical Engineering  (2022)

\bibitem{Chen-2006-ASCA}
Chen, X., Zhou, X., Wong, S.T.: {Automated segmentation, classification, and
  tracking of cancer cell nuclei in time-lapse microscopy}. IEEE Transactions
  on Biomedical Engineering  \textbf{53}(4),  762--766 (2006)

\bibitem{Chicco-2020-TATM}
Chicco, D., Jurman, G.: {The advantages of the Matthews correlation coefficient
  (MCC) over F1 score and accuracy in binary classification evaluation}. BMC
  genomics  \textbf{21}(1), ~6 (2020)

\bibitem{Chien-2007-USEA}
Chien, T.I., Kao, J.T., Liu, H.L., Lin, P.C., Hong, J.S., Hsieh, H.P., Chien,
  M.J.: {Urine sediment examination: a comparison of automated urinalysis
  systems and manual microscopy}. Clinica Chimica Acta  \textbf{384}(1-2),
  28--34 (2007)

\bibitem{Cholakkal-2019-OCAI}
Cholakkal, H., Sun, G., Khan, F.S., Shao, L.: {Object counting and instance
  segmentation with image-level supervision}. In: Proceedings of the IEEE/CVF
  Conference on Computer Vision and Pattern Recognition. pp. 12397--12405
  (2019)

\bibitem{Congestri-2000-EBBF}
Congestri, R., Federici, R., Albertano, P.: {Evaluating biomass of Baltic
  filamentous cyanobacteria by image analysis}. Aquatic microbial ecology
  \textbf{22}(3),  283--290 (2000)

\bibitem{Cordoba-2010-EIGC}
C{\'o}rdoba-Matson, M.V., Guti{\'e}rrez, J., Porta-G{\'a}ndara, M.{\'A}.:
  {Evaluation of Isochrysis galbana (clone T-ISO) cell numbers by digital image
  analysis of color intensity}. Journal of applied phycology  \textbf{22}(4),
  427--434 (2010)

\bibitem{Costa-2013-QIAT}
Costa, J., Mesquita, D., Amaral, A., Alves, M., Ferreira, E.: {Quantitative
  image analysis for the characterization of microbial aggregates in biological
  wastewater treatment: a review}. Environmental Science and Pollution Research
   \textbf{20}(9),  5887--5912 (2013)

\bibitem{Couri-2006-DIPA}
Couri, S., Merces, E., Neves, B., Senna, L.: {Digital image processing as a
  tool to monitor biomass growth in Aspergillus niger 3T5B8 solid-state
  fermentation: preliminary results}. Journal of Microscopy  \textbf{224}(3),
  290--297 (2006)

\bibitem{Coutteau-1996-MA}
Coutteau, P.: {Micro-algae}. Manual on the production and use of live food for
  aquaculture  \textbf{361},  7--48 (1996)

\bibitem{Cross-1983-BCEU}
Cross, H., Gilliland, D., Durland, P., Seideman, S.: {Beef carcass evaluation
  by use of a video image analysis system}. Journal of Animal Science
  \textbf{57}(4),  908--917 (1983)

\bibitem{Cui-2019-OAEP}
Cui, J., Li, F., Shi, Z.L.: {Origin and evolution of pathogenic coronaviruses}.
  Nature Reviews Microbiology  \textbf{17}(3),  181--192 (2019)

\bibitem{Daims-2007-QUMF}
Daims, H., Wagner, M.: {Quantification of uncultured microorganisms by
  fluorescence microscopy and digital image analysis}. Applied microbiology and
  biotechnology  \textbf{75}(2),  237--248 (2007)

\bibitem{Dazzo-2013-CMEI}
Dazzo, F., Gross, C.: {CMEIAS Quadrat Maker: a digital software tool to
  optimize grid dimensions and produce quadrat images for landscape ecology
  spatial analysis}. J Ecosystem and Ecography  \textbf{3}(4), ~136 (2013)

\bibitem{Dazzo-2013-SEMB}
Dazzo, F.B., Klemmer, K.J., Chandler, R., Yanni, Y.G.: {In situ ecophysiology
  of microbial biofilm communities analyzed by CMEIAS computer-assisted
  microscopy at single-cell resolution}. Diversity  \textbf{5}(3),  426--460
  (2013)

\bibitem{Dazzo-2015-UCIA}
Dazzo, F.B., Niccum, B.C.: {Use of CMEIAS image analysis software to accurately
  compute attributes of cell size, morphology, spatial aggregation and color
  segmentation that signify in situ ecophysiological adaptations in microbial
  biofilm communities}. Computation  \textbf{3}(1),  72--98 (2015)

\bibitem{Dietler-2020-ACNN}
Dietler, N., Minder, M., Gligorovski, V., Economou, A.M., Joly, D.A.H.L.,
  Sadeghi, A., Chan, C.H.M., Kozi{\'n}ski, M., Weigert, M., Bitbol, A.F.,
  et~al.: {A convolutional neural network segments yeast microscopy images with
  high accuracy}. Nature communications  \textbf{11}(1), ~1--8 (2020)

\bibitem{Dosovitskiy-2020-AIWW}
Dosovitskiy, A., Beyer, L., Kolesnikov, A., Weissenborn, D., Zhai, X.,
  Unterthiner, T., Dehghani, M., Minderer, M., Heigold, G., Gelly, S., et~al.:
  {An image is worth 16x16 words: Transformers for image recognition at scale}.
  arXiv preprint arXiv:2010.11929  (2020)

\bibitem{Dutra-2007-LPSS}
Dutra, J.C., da~Terzi, S.C., Bevilaqua, J.V., Damaso, M.C., Couri, S., Langone,
  M.A., Senna, L.F.: {Lipase production in solid-state fermentation monitoring
  biomass growth of Aspergillus niger using digital image processing}. In:
  Biotechnology for Fuels and Chemicals, pp. 431--443. Springer (2007)

\bibitem{Ebrahimi-2015-CLPR}
Ebrahimi, A., Amirkhani, A., A~Raie, A., Mosavi, M.R.: {Car license plate
  recognition using color features of Persian license plates}. Journal of
  Advances in Computer Research  \textbf{6}(4),  27--38 (2015)

\bibitem{Ekstrom-2012-DIPT}
Ekstrom, M.P.: {Digital image processing techniques}, vol.~2. Academic Press
  (2012)

\bibitem{Eniko-2022-UCNE}
Eniko, T., Lerf, V., Toth, I., Kisantal, T., Varbiro, G., Vasas, G., Viktoria,
  B., Gorgenyi, J., Lukacs, A., Kokai, Z., et~al.: {Uncertainties of cell
  number estimation in cyanobacterial colonies and the potential use of sphere
  packing}. bioRxiv  (2022)

\bibitem{Ernst-2006-DTFC}
Ernst, B., Neser, S., O’Brien, E., Hoeger, S.J., Dietrich, D.R.:
  {Determination of the filamentous cyanobacteria Planktothrix rubescens in
  environmental water samples using an image processing system}. Harmful algae
  \textbf{5}(3),  281--289 (2006)

\bibitem{Estep-1986-MAUF}
Estep, K.W., MacIntyre, F., Hjorleifsson, E., Sieburth, J.: {MacImage: a
  user-friendly image-analysis system for the accurate mensuration of marine
  organisms}. Mar. Ecol. Prog. Ser  \textbf{33},  243--253 (1986)

\bibitem{Fan-2019-BRID}
Fan, L., Zhang, F., Fan, H., Zhang, C.: {Brief review of image denoising
  techniques}. Visual Computing for Industry, Biomedicine, and Art
  \textbf{2}(1),  1--12 (2019)

\bibitem{Folland-2014-ABFC}
Folland, I., Trione, D., Dazzo, F.: Accuracy of biovolume formulas for cmeias
  computer-assisted microscopy and body size analysis of morphologically
  diverse microbial populations and communities. Microbial ecology
  \textbf{68}(3),  596--610 (2014)

\bibitem{Gandola-2016-ACQU}
Gandola, E., Antonioli, M., Traficante, A., Franceschini, S., Scardi, M.,
  Congestri, R.: {ACQUA: Automated Cyanobacterial Quantification Algorithm for
  toxic filamentous genera using spline curves, pattern recognition and machine
  learning}. Journal of microbiological methods  \textbf{124},  48--56 (2016)

\bibitem{Garcia-2017-ARDL}
Garcia-Garcia, A., Orts-Escolano, S., Oprea, S., Villena-Martinez, V.,
  Garcia-Rodriguez, J.: {A review on deep learning techniques applied to
  semantic segmentation}. arXiv preprint arXiv:1704.06857  (2017)

\bibitem{Garofano-2005-ATWI}
Gar{\'o}fano, G., Venancio, C., Suazo, C., Almeida, P.: {Application of the
  wavelet image analysis technique to monitor cell concentration in
  bioprocesses}. Brazilian Journal of Chemical Engineering  \textbf{22}(4),
  573--583 (2005)

\bibitem{Gayen-2008-QCSD}
Gayen, K., Venkatesh, K.: {Quantification of cell size distribution as applied
  to the growth of Corynebacterium glutamicum}. Microbiological research
  \textbf{163}(5),  586--593 (2008)

\bibitem{Ghitua-2013-OTHE}
Ghiț{\u{a}}, S., Acomi, N.: {Optimizing the heat exchange in ballast water
  treatment using software monitoring methods}. Global Journal on Technology
  \textbf{4}(2) (2013)

\bibitem{Gmur-2000-AIES}
Gm{\"u}r, R., Guggenheim, B., Giertsen, E., Thurnheer, T.: {Automated
  immunofluorescence for enumeration of selected taxa in supragingival dental
  plaque}. European journal of oral sciences  \textbf{108}(5),  393--402 (2000)

\bibitem{Gonzales-2002-DIP}
Gonzales, R.C., Woods, R.E.: {Digital image processing} (2002)

\bibitem{Gonzalez-2004-DIPU}
Gonzalez, R.C., Woods, R.E., Eddins, S.L.: {Digital image processing using
  MATLAB}. Pearson Education India (2004)

\bibitem{Gracias-2004-ARCD}
Gracias, K.S., McKillip, J.L.: {A review of conventional detection and
  enumeration methods for pathogenic bacteria in food}. Canadian journal of
  microbiology  \textbf{50}(11),  883--890 (2004)

\bibitem{Gray-2002-CIAS}
Gray, A., Young, D., Martin, N., Glasbey, C.: {Cell identification and sizing
  using digital image analysis for estimation of cell biomass in High Rate
  Algal Ponds}. Journal of applied phycology  \textbf{14}(3),  193--204 (2002)

\bibitem{Han-2020-ASVT}
Han, K., Wang, Y., Chen, H., Chen, X., Guo, J., Liu, Z., Tang, Y., Xiao, A.,
  Xu, C., Xu, Y., et~al.: {A Survey on Visual Transforme}r. arXiv preprint
  arXiv:2012.12556  (2020)

\bibitem{Haralick-1985-IST}
Haralick, R.M., Shapiro, L.G.: {Image segmentation techniques}. Computer
  vision, graphics, and image processing  \textbf{29}(1),  100--132 (1985)

\bibitem{Hay-1988-TDGE}
Hay, A.: {The derivation of global estimates from a confusion matrix}.
  International Journal of Remote Sensing  \textbf{9}(8),  1395--1398 (1988)

\bibitem{Heydorn-2000-ABST}
Heydorn, A., Nielsen, A.T., Hentzer, M., Sternberg, C., Givskov, M.,
  Ersb{\o}ll, B.K., Molin, S.: {Quantification of biofilm structures by the
  novel computer program COMSTAT}. Microbiology  \textbf{146}(10),  2395--2407
  (2000)

\bibitem{Hu-2018-SAEN}
Hu, J., Shen, L., Sun, G.: {Squeeze-and-excitation networks}. In: Proceedings
  of the IEEE conference on computer vision and pattern recognition. pp.
  7132--7141 (2018)

\bibitem{Huang-2016-SCMI}
Huang, X., Li, C., Shen, M., Shirahama, K., Nyffeler, J., Leist, M.,
  Grzegorzek, M., Deussen, O.: {Stem cell microscopic image segmentation using
  supervised normalized cuts}. In: 2016 IEEE International Conference on Image
  Processing (ICIP). pp. 4140--4144. IEEE (2016)

\bibitem{Huang-2020-SSCC}
Huang, X., Shirahama, K., Li, F., Grzegorzek, M.: {Sleep stage classification
  for child patients using DeConvolutional Neural Network}. Artificial
  Intelligence in Medicine  \textbf{110},  101981 (2020)

\bibitem{Hui-2020-TCET}
Hui, D.S., Azhar, E.I., Madani, T.A., Ntoumi, F., Kock, R., Dar, O., Ippolito,
  G., Mchugh, T.D., Memish, Z.A., Drosten, C., et~al.: {The continuing
  2019-nCoV epidemic threat of novel coronaviruses to global health—The
  latest 2019 novel coronavirus outbreak in Wuhan, China}. International
  Journal of Infectious Diseases  \textbf{91},  264--266 (2020)

\bibitem{Huttenlocher-1993-CIUT}
Huttenlocher, D.P., Klanderman, G.A., Rucklidge, W.J.: {Comparing images using
  the Hausdorff distance}. IEEE Transactions on pattern analysis and machine
  intelligence  \textbf{15}(9),  850--863 (1993)

\bibitem{Jaderberg-2015-STNW}
Jaderberg, M., Simonyan, K., Zisserman, A., Kavukcuoglu, K.: {Spatial
  transformer networks}. arXiv preprint arXiv:1506.02025  (2015)

\bibitem{Jiang-2019-CCAD}
Jiang, X., Xiao, Z., Zhang, B., Zhen, X., Cao, X., Doermann, D., Shao, L.:
  {Crowd counting and density estimation by trellis encoder-decoder networks}.
  In: Proceedings of the IEEE/CVF Conference on Computer Vision and Pattern
  Recognition. pp. 6133--6142 (2019)

\bibitem{Jung-2003-IALD}
Jung, S.K., Lee, S.B.: {Image analysis of light distribution in a
  photobioreactor}. Biotechnology and bioengineering  \textbf{84}(3),  394--397
  (2003)

\bibitem{Korber-1989-ELFV}
Korber, D.R., Lawrence, J.R., Sutton, B., Caldwell, D.E.: {Effect of laminar
  flow velocity on the kinetics of surface recolonization by Mot+ and Mot-
  Pseudomonas fluorescens}. Microbial ecology  \textbf{18}(1),  1--19 (1989)

\bibitem{Kosov-2018-EMCC}
Kosov, S., Shirahama, K., Li, C., Grzegorzek, M.: {Environmental microorganism
  classification using conditional random fields and deep convolutional neural
  networks}. Pattern recognition  \textbf{77},  248--261 (2018)

\bibitem{Krambeck-1981-MABD}
Krambeck, C., Krambeck, H.J., Overbeck, J.: {Microcomputer-assisted biomass
  determination of plankton bacteria on scanning electron micrographs}. Applied
  and Environmental Microbiology  \textbf{42}(1),  142--149 (1981)

\bibitem{Kuehn-1998-ACLS}
Kuehn, M., Hausner, M., Bungartz, H.J., Wagner, M., Wilderer, P.A., Wuertz, S.:
  {Automated confocal laser scanning microscopy and semiautomated image
  processing for analysis of biofilms}. Applied and environmental microbiology
  \textbf{64}(11),  4115--4127 (1998)

\bibitem{Kulwa-2021-ANPD}
Kulwa, F., Li, C., Zhang, J., Shirahama, K., Kosov, S., Zhao, X., Sun, H.,
  Jiang, T., Grzegorzek, M.: {A New Pairwise Deep Learning Feature For
  Environmental Microorganism Image Analysis}. arXiv preprint arXiv:2102.12147
  (2021)

\bibitem{Kulwa-2019-ASSM}
Kulwa, F., Li, C., Zhao, X., Cai, B., Xu, N., Qi, S., Chen, S., Teng, Y.: {A
  state-of-the-art survey for microorganism image segmentation methods and
  future potential}. IEEE Access  \textbf{7},  100243--100269 (2019)

\bibitem{Lawrence-1989-CEDM}
Lawrence, J., Korber, D., Caldwell, D.: {Computer-enhanced darkfield microscopy
  for the quantitative analysis of bacterial growth and behavior on surfaces}.
  Journal of microbiological methods  \textbf{10}(2),  123--138 (1989)

\bibitem{Leal-2016-MBEF}
Leal, C., Amaral, A.L., de~Lourdes~Costa, M.: {Microbial-based evaluation of
  foaming events in full-scale wastewater treatment plants by microscopy survey
  and quantitative image analysis}. Environmental Science and Pollution
  Research  \textbf{23}(15),  15638--15650 (2016)

\bibitem{Lecault-2009-AIAT}
Lecault, V., Patel, N., Thibault, J.: {An image analysis technique to estimate
  the cell density and biomass concentration of Trichoderma reesei}. Letters in
  applied microbiology  \textbf{48}(4),  402--407 (2009)

\bibitem{Lee-2020-MISA}
Lee, C.W., Ren, Y.J., Marella, M., Wang, M., Hartke, J., Couto, S.S.:
  {Multiplex immunofluorescence staining and image analysis assay for diffuse
  large B cell lymphoma}. Journal of immunological methods  \textbf{478},
  112714 (2020)

\bibitem{Chen-2020-ARCH}
Li, C., Chen, H., Li, X., Xu, N., Hu, Z., Xue, D., Qi, S., Ma, H., Zhang, L.,
  Sun, H.: {A review for cervical histopathology image analysis using machine
  vision approaches}. Artificial Intelligence Review pp. 1--42 (2020)

\bibitem{Li-2020-ARCM}
Li, C., Kulwa, F., Zhang, J., Li, Z., Xu, H., Zhao, X.: {A review of clustering
  methods in microorganism image analysis}. Information Technology in
  Biomedicine  \textbf{1186},  13--25 (2020)

\bibitem{Li-2021-ACRC}
Li, C., Li, X., Rahaman, M., Li, X., Sun, H., Zhang, H., Zhang, Y., Li, X., Wu,
  J., Yao, Y., et~al.: {A comprehensive review of computer-aided whole-slide
  image analysis: from datasets to feature extraction, segmentation,
  classification, and detection approaches}. arXiv preprint arXiv:2102.10553
  (2021)

\bibitem{Li-2021-ASSO}
Li, C., Ma, P., Rahaman, M.M., Yao, Y., Zhang, J., Zou, S., Zhao, X.,
  Grzegorzek, M.: {A State-of-the-art Survey of Object Detection Techniques in
  Microorganism Image Analysis: from Traditional Image Processing and Classical
  Machine Learning to Current Deep Convolutional Neural Networks and Potential
  Visual Transformers}. arXiv preprint arXiv:2105.03148  (2021)

\bibitem{Li-2015-ACIA}
Li, C., Shirahama, K., Grzegorzek, M.: {Application of content-based image
  analysis to environmental microorganism classification}. Biocybernetics and
  Biomedical Engineering  \textbf{35}(1),  10--21 (2015)

\bibitem{Li-2016-EMAC}
Li, C., Shirahama, K., Grzegorzek, M.: {Environmental microbiology aided by
  content-based image analysis}. Pattern Analysis and Applications
  \textbf{19}(2),  531--547 (2016)

\bibitem{Li-2019-ASTA}
Li, C., Wang, K., Xu, N.: {A survey for the applications of content-based
  microscopic image analysis in microorganism classification domains}.
  Artificial Intelligence Review  \textbf{51}(4),  577--646 (2019)

\bibitem{Li-2020-ASMI}
Li, C., Zhang, J., Kulwa, F., Qi, S., Qi, Z.: {A SARS-CoV-2 Microscopic Image
  Dataset with Ground Truth Images and Visual Features}. In: Chinese Conference
  on Pattern Recognition and Computer Vision (PRCV). pp. 244--255. Springer
  (2020)

\bibitem{Li-2018-CFLM}
Li, F., Shirahama, K., Nisar, M.A., K{\"o}ping, L., Grzegorzek, M.: {Comparison
  of feature learning methods for human activity recognition using wearable
  sensors}. Sensors  \textbf{18}(2), ~679 (2018)

\bibitem{Li-2021-ACRM}
Li, Y., Li, C., Li, X., Wang, K., Rahaman, M.M., Sun, C., Chen, H., Wu, X.,
  Zhang, H., Wang, Q.: {A Comprehensive Review of Markov Random Field and
  Conditional Random Field Approaches in Pathology Image Analysis}. Archives of
  Computational Methods in Engineering pp. 1--31 (2021)

\bibitem{Liu-2022-ITAT}
Liu, W., Li, C., Rahaman, M.M., Jiang, T., Sun, H., Wu, X., Hu, W., Chen, H.,
  Sun, C., Yao, Y., et~al.: {Is the aspect ratio of cells important in deep
  learning? A robust comparison of deep learning methods for multi-scale
  cytopathology cell image classification: From convolutional neural networks
  to visual transformers}. Computers in biology and medicine  \textbf{141},
  105026 (2022)

\bibitem{Liu-2004-HTIB}
Liu, X., Wang, S., Sendi, L., Caulfield, M.J.: {High-throughput imaging of
  bacterial colonies grown on filter plates with application to serum
  bactericidal assays}. Journal of immunological methods  \textbf{292}(1-2),
  187--193 (2004)

\bibitem{Lomander-2002-AMRA}
Lomander, A., Schreuders, P., Russek-Cohen, E., Ali, L.: {A method for rapid
  analysis of biofilm morphology and coverage on glass and polished and brushed
  stainless steel}. Transactions of the Asae  \textbf{45}(2), ~479 (2002)

\bibitem{Madigan-1997-BBOM}
Madigan, M.T., Martinko, J.M., Parker, J., et~al.: {Brock biology of
  microorganisms}, vol.~11. Prentice hall Upper Saddle River, NJ (1997)

\bibitem{Mariey-2001-DCIM}
Mariey, L., Signolle, J., Amiel, C., Travert, J.: {Discrimination,
  classification, identification of microorganisms using FTIR spectroscopy and
  chemometrics}. Vibrational spectroscopy  \textbf{26}(2),  151--159 (2001)

\bibitem{Mcnair-2021-PCAN}
Mcnair, H., Hammond, C.N., Menden-Deuer, S.: {Phytoplankton carbon and nitrogen
  biomass estimates are robust to volume measurement method and growth
  environment}. Journal of Plankton Research  \textbf{43}(2),  103--112 (2021)

\bibitem{Mesquita-2010-DAME}
Mesquita, D., Dias, O., Elias, R., Amaral, A.L., Ferreira, E.: {Dilution and
  magnification effects on image analysis applications in activated sludge
  characterization}. Microscopy and Microanalysis  \textbf{16}(05),  561--568
  (2010)

\bibitem{Miszkiewicz-2004-PPAE}
Miszkiewicz, H., Bizukojc, M., Rozwandowicz, A., Bielecki, S.: {Physiological
  properties and enzymatic activities of Rhizopus oligosporus in solid state
  fermentations}. Electronic Journal of Polish Agricultural Universities
  \textbf{7} (2004)

\bibitem{Mohan-2021-OTMP}
Mohan, J., Saros, J., Stone, J.R.: {On the matter of phytoplankton: A novel
  method using 3D computer models to calculate biovolume of microorganisms}.
  Limnology and Oceanography: Methods  \textbf{19}(5),  331--339 (2021)

\bibitem{Morgan-1991-AIAM}
Morgan, P., Cooper, C., Battersby, N., Lee, S., Lewis, S., Machin, T., Graham,
  S., Watkinson, R.: {Automated image analyss method to determine fungal
  biomass in soils and on solid matrices}. Soil Biology and Biochemistry
  \textbf{23}(7),  609--616 (1991)

\bibitem{Mueller-2006-AAMP}
Mueller, L.N., De~Brouwer, J.F., Almeida, J.S., Stal, L.J., Xavier, J.B.:
  {Analysis of a marine phototrophic biofilm by confocal laser scanning
  microscopy using the new image quantification software PHLIP}. BMC ecology
  \textbf{6}(1), ~1 (2006)

\bibitem{Mukhopadhyay-2003-MMSG}
Mukhopadhyay, S., Chanda, B.: {Multiscale morphological segmentation of
  gray-scale images}. IEEE transactions on Image Processing  \textbf{12}(5),
  533--549 (2003)

\bibitem{Otsu-1979-ATSM}
Otsu, N.: {A threshold selection method from gray-level histograms}. IEEE
  transactions on systems, man, and cybernetics  \textbf{9}(1),  62--66 (1979)

\bibitem{Pan-1999-TDMW}
Pan, Q., Zhang, L., Dai, G., Zhang, H.: {Two denoising methods by wavelet
  transform}. IEEE transactions on signal processing  \textbf{47}(12),
  3401--3406 (1999)

\bibitem{De-2009-IASB}
de~Paz, L.E.C.: {Image analysis software based on color segmentation for
  characterization of viability and physiological activity of biofilms}.
  Applied and environmental microbiology  \textbf{75}(6),  1734--1739 (2009)

\bibitem{Perez-1987-AITA}
Perez, A., Gonzalez, R.C.: {An iterative thresholding algorithm for image
  segmentation}. IEEE transactions on pattern analysis and machine intelligence
   \textbf{9}(6),  742--751 (1987)

\bibitem{Pernthaler-2003-AEGM}
Pernthaler, J., Pernthaler, A., Amann, R.: {Automated enumeration of groups of
  marine picoplankton after fluorescence in situ hybridization}. Applied and
  Environmental Microbiology  \textbf{69}(5),  2631--2637 (2003)

\bibitem{Pesatori-2013-IISF}
Pesatori, A., Magnani, A., Norgia, M.: {Infrared image sensor for fire
  location}. In: 2013 IEEE International Instrumentation and Measurement
  Technology Conference (I2MTC) (2013)

\bibitem{Petrisor-2004-RACM}
Petrisor, A.I., Cuc, A., Decho, A.W.: {Reconstruction and computation of
  microscale biovolumes using geographical information systems: potential
  difficulties}. Research in microbiology  \textbf{155}(6),  447--454 (2004)

\bibitem{Pettipher-1982-SACB}
Pettipher, G., Rodrigues, U.M.: {Semi-automated counting of bacteria and
  somatic cells in milk using epifluorescence microscopy and television image
  analysis}. Journal of Applied Bacteriology  \textbf{53}(3),  323--329 (1982)

\bibitem{Posch-2009-NIAT}
Posch, T., Franzoi, J., Prader, M., Salcher, M.M.: {New image analysis tool to
  study biomass and morphotypes of three major bacterioplankton groups in an
  alpine lake}. Aquatic Microbial Ecology  \textbf{54}(2),  113--126 (2009)

\bibitem{Puyen-2012-VABM}
Puyen, Z.M., Villagrasa, E., Maldonado, J., Esteve, I., Sol{\'e}, A.:
  {Viability and biomass of Micrococcus luteus DE2008 at different salinity
  concentrations determined by specific fluorochromes and CLSM-image analysis}.
  Current microbiology  \textbf{64}(1),  75--80 (2012)

\bibitem{Qi-2021-ACOI}
Qi, Y., Yang, Z., Sun, W., Lou, M., Lian, J., Zhao, W., Deng, X., Ma, Y.: {A
  Comprehensive Overview of Image Enhancement Techniques}. Archives of
  Computational Methods in Engineering pp. 1--25 (2021)

\bibitem{Qiu-2004-AMTA}
Qiu, D., Jiao, N., Qian, L.: {Advance in measured techniquesof aquatic
  bacterial counting and cell sizes}. Journal of Oceanography in Taiwan Strait
  \textbf{23}(3),  376--385 (2004)

\bibitem{Rahaman-2020-ASCC}
Rahaman, M.M., Li, C., Wu, X., Yao, Y., Hu, Z., Jiang, T., Li, X., Qi, S.: {A
  survey for cervical cytopathology image analysis using deep learning}. IEEE
  Access  \textbf{8},  61687--61710 (2020)

\bibitem{Rahaman-2020-ICSC}
Rahaman, M.M., Li, C., Yao, Y., Kulwa, F., Rahman, M.A., Wang, Q., Qi, S.,
  Kong, F., Zhu, X., Zhao, X.: {Identification of COVID-19 samples from chest
  X-Ray images using deep learning: A comparison of transfer learning
  approaches}. Journal of X-ray Science and Technology  \textbf{28}(5),
  821--839 (2020)

\bibitem{Rahaman-2021-DADL}
Rahaman, M.M., Li, C., Yao, Y., Kulwa, F., Wu, X., Li, X., Wang, Q.:
  {DeepCervix: A deep learning-based framework for the classification of
  cervical cells using hybrid deep feature fusion techniques}. Computers in
  Biology and Medicine  \textbf{136},  104649 (2021)

\bibitem{Rahman-2016-OIOU}
Rahman, M.A., Wang, Y.: {Optimizing intersection-over-union in deep neural
  networks for image segmentation}. In: International symposium on visual
  computing. pp. 234--244. Springer (2016)

\bibitem{Rajapaksha-2019-ARMT}
Rajapaksha, P., Elbourne, A., Gangadoo, S., Brown, R., Cozzolino, D., Chapman,
  J.: {A review of methods for the detection of pathogenic microorganisms}.
  Analyst  \textbf{144}(2),  396--411 (2019)

\bibitem{Rani-2021-MLAD}
Rani, P., Kotwal, S., Manhas, J., Sharma, V., Sharma, S.: {Machine learning and
  deep learning based computational approaches in automatic microorganisms
  image recognition: methodologies, challenges, and developments}. Archives of
  Computational Methods in Engineering pp. 1--37 (2021)

\bibitem{Rodriguez-2007-TDQS}
Rodr{\'\i}guez, S.J., Bishop, P.L.: {Three-dimensional quantification of soil
  biofilms using image analysis}. Environmental Engineering Science
  \textbf{24}(1),  96--103 (2007)

\bibitem{Ross-2012-EABS}
Ross, S.S., Reinhardt, J.M., Fiegel, J.: {Enhanced analysis of bacteria
  susceptibility in connected biofilms}. Journal of microbiological methods
  \textbf{90}(1),  9--14 (2012)

\bibitem{Rousso-2022-CSDA}
Rousso, B.Z., Bertone, E., Stewart, R.A., Hughes, S.P., Hobson, P., Hamilton,
  D.P.: {Cyanobacteria species dominance and diversity in three Australian
  drinking water reservoirs}. Hydrobiologia  \textbf{849}(6),  1453--1469
  (2022)

\bibitem{Schonholzer-1999-OAFF}
Sch{\"o}nholzer, F., Hahn, D., Zeyer, J.: {Origins and fate of fungi and
  bacteria in the gut of Lumbricus terrestris L. studied by image analysis}.
  FEMS Microbiology Ecology  \textbf{28}(3),  235--248 (1999)

\bibitem{Sekertekin-2021-ASGT}
Sekertekin, A.: {A survey on global thresholding methods for mapping open water
  body using Sentinel-2 satellite imagery and normalized difference water
  index}. Archives of Computational Methods in Engineering  \textbf{28}(3),
  1335--1347 (2021)

\bibitem{Shirahama-2016-TLMR}
Shirahama, K., Grzegorzek, M.: {Towards large-scale multimedia retrieval
  enriched by knowledge about human interpretation}. Multimedia Tools and
  Applications  \textbf{75}(1),  297--331 (2016)

\bibitem{Shuxin-2004-MAMS}
Shuxin, Z., Yun, C., Weihua, T., Shiru, J.: {Morphological analysis of
  monascuson surface fermentation}. Microbiology China  \textbf{31}(4),  65--70
  (2004)

\bibitem{Sole-2009-CLSM}
Sol{\'e}, A., Diestra, E., Esteve, I.: {Confocal laser scanning microscopy
  image analysis for cyanobacterial biomass determined at microscale level in
  different microbial mats}. Microbial ecology  \textbf{57}(4),  649--656
  (2009)

\bibitem{Sole-2007-ANMI}
Sol{\'e}, A., Mas, J., Esteve, I.: {A new method based on image analysis for
  determining cyanobacterial biomass by CLSM in stratified benthic sediments}.
  Ultramicroscopy  \textbf{107}(8),  669--673 (2007)

\bibitem{Song-2013-RSPA}
Song, S., Yongkun, B., Xiaoxia, S.: {Relationship between shape parameters and
  dry weight of the dominant zooplankton in Jiaozhou bay based on image
  method}. Oceanologia et Limnologia Sinica  \textbf{44}(1),  15--22 (2013)

\bibitem{Strahler-1957-QAWG}
Strahler, A.N.: {Quantitative analysis of watershed geomorphology}. Eos,
  Transactions American Geophysical Union  \textbf{38}(6),  913--920 (1957)

\bibitem{Su-2014-ANNA}
Su, M.C., Cheng, C.Y., Wang, P.C.: {A neural-network-based approach to white
  blood cell classification}. The scientific world journal  \textbf{2014}
  (2014)

\bibitem{Dodi-2012-APIP}
Sudiana, D., Rizkinia, M.: {ALOS/PALSAR Image Processing Using Dinsar and Log
  Ratio for Flood Early Detection in Jakarta Based on Land Subsidences}. Makara
  Journal of Technology  \textbf{15}(2),  193--200 (2012)

\bibitem{Tackx-1995-MSFE}
Tackx, M., Zhu, L., De~Coster, W., Billones, R., Daro, M.: {Measuring
  selectivity of feeding by estuarine copepods using image analysis combined
  with microscopic and Coulter counting}. ICES Journal of Marine Science
  \textbf{52}(3-4),  419--425 (1995)

\bibitem{Thiran-1994-ARCC}
Thiran, J.P., Becks, M.O., Macq, B.M., Mairesse, J.: {Automatic recognition of
  cancerous cells using mathematical morphology}. In: Visualization in
  Biomedical Computing 1994. vol.~2359, pp. 392--401. International Society for
  Optics and Photonics (1994)

\bibitem{Thyagharajan-2020-ARND}
Thyagharajan, K., Kalaiarasi, G.: {A review on near-duplicate detection of
  images using computer vision techniques}. Archives of Computational Methods
  in Engineering pp. 1--20 (2020)

\bibitem{Tucker-1992-FAMM}
Tucker, K.G., Kelly, T., Delgrazia, P., Thomas, C.R.: {Fully-automatic
  measurement of mycelial morphology by image analysis}. Biotechnology progress
   \textbf{8}(4),  353--359 (1992)

\bibitem{JHU-2020-CCGC}
University, J.H.: Coronavirus covid-19 global cases by the center for systems
  science and engineering (csse) at johns hopkins university (jhu). {Available
  at: \url{https://coronavirus.jhu.edu/map.html}} (2020)

\bibitem{Webster-2007-ITF}
Webster, J., Weber, R.: {Introduction to fungi}. Cambridge university press
  (2007)

\bibitem{Xu-2020-AEFG}
Xu, H., Li, C., Rahaman, M.M., Yao, Y., Li, Z., Zhang, J., Kulwa, F., Zhao, X.,
  Qi, S., Teng, Y.: {An enhanced framework of generative adversarial networks
  (EF-GANs) for environmental microorganism image augmentation with limited
  rotation-invariant training data}. IEEE Access  \textbf{8},  187455--187469
  (2020)

\bibitem{Xu-2011-CAOT}
Xu, X., Xu, S., Jin, L., Song, E.: {Characteristic analysis of Otsu threshold
  and its applications}. Pattern recognition letters  \textbf{32}(7),  956--961
  (2011)

\bibitem{Yang-2020-RPNP}
Yang, Y., Li, G., Wu, Z., Su, L., Huang, Q., Sebe, N.: {Reverse perspective
  network for perspective-aware object counting}. In: Proceedings of the
  IEEE/CVF Conference on Computer Vision and Pattern Recognition. pp.
  4374--4383 (2020)

\bibitem{Yu-2004-MFAI}
Yu, H., MacGregor, J.F.: {Monitoring flames in an industrial boiler using
  multivariate image analysis}. AIChE Journal  \textbf{50}(7),  1474--1483
  (2004)

\bibitem{Zhang-2008-ISEA}
Zhang, H., Fritts, J.E., Goldman, S.A.: {Image segmentation evaluation: A
  survey of unsupervised methods}. computer vision and image understanding
  \textbf{110}(2),  260--280 (2008)

\bibitem{Zhang-2021-ACRI}
Zhang, J., Li, C., Rahaman, M.M., Yao, Y., Ma, P., Zhang, J., Zhao, X., Jiang,
  T., Grzegorzek, M.: {A comprehensive review of image analysis methods for
  microorganism counting: from classical image processing to deep learning
  approaches}. Artificial Intelligence Review pp. 1--70 (2021)

\bibitem{Zhang-2022-AAPI}
Zhang, J., Xu, N., Li, C., Rahaman, M.M., Yao, Y.D., Lin, Y.H., Zhang, J.,
  Jiang, T., Qin, W., Grzegorzek, M.: {An application of Pixel Interval
  Down-sampling (PID) for dense tiny microorganism counting on environmental
  microorganism images}. arXiv preprint arXiv:2204.01341  (2022)

\bibitem{Zhang-2021-AANN}
Zhang, J., Li, C., Grzegorzek, M.: {Applications of Artificial Neural Networks
  in Microorganism Image Analysis: A Comprehensive Review from Conventional
  Multilayer Perceptron to Popular Convolutional Neural Network and Potential
  Visual Transformer}. arXiv preprint arXiv:2108.00358  (2021)

\bibitem{Zhang-2021-LANL}
Zhang, J., Li, C., Kosov, S., Grzegorzek, M., Shirahama, K., Jiang, T., Sun,
  C., Li, Z., Li, H.: {LCU-Net: A Novel Low-cost U-Net for Environmental
  Microorganism Image Segmentation}. Pattern Recognition p. 107885 (2021)

\bibitem{Zhao-2022-ACAD}
Zhao, P., Li, C., Rahaman, M., Xu, H., Yang, H., Sun, H., Jiang, T.,
  Grzegorzek, M.: {A Comparative Study of Deep Learning Classification Methods
  on a Small Environmental Microorganism Image Dataset (EMDS-6): From
  Convolutional Neural Networks to Visual Transformers}. Frontiers in
  Microbiology  \textbf{13} (2022)

\bibitem{Zhao-2022-EEMI}
Zhao, P., Li, C., Rahaman, M.M., Xu, H., Ma, P., Yang, H., Sun, H., Jiang, T.,
  Xu, N., Grzegorzek, M.: {EMDS-6: Environmental Microorganism Image Dataset
  Sixth Version for Image Denoising, Segmentation, Feature Extraction,
  Classification, and Detection Method Evaluation}. Frontiers in Microbiology
  p.~1334 (2022)

\bibitem{Zhou-2020-ACRB}
Zhou, X., Li, C., Rahaman, M.M., Yao, Y., Ai, S., Sun, C., Wang, Q., Zhang, Y.,
  Li, M., Li, X., et~al.: {A comprehensive review for breast histopathology
  image analysis using classical and deep neural networks}. IEEE Access
  \textbf{8},  90931--90956 (2020)

\bibitem{Zhu-2007-AISA}
Zhu, S., Xia, X., Zhang, Q., Belloulata, K.: {An image segmentation algorithm
  in image processing based on threshold segmentation}. In: 2007 third
  international IEEE conference on signal-image technologies and internet-based
  system. pp. 673--678. IEEE (2007)

\bibitem{Ziegler-2004-MFCD}
Ziegler, U., Groscurth, P.: {Morphological features of cell death}. Physiology
  \textbf{19}(3),  124--128 (2004)

\bibitem{Zou-2022-TAEC}
Zou, S., Li, C., Sun, H., Xu, P., Zhang, J., Ma, P., Yao, Y., Huang, X.,
  Grzegorzek, M.: {TOD-CNN: An effective convolutional neural network for tiny
  object detection in sperm videos}. Computers in Biology and Medicine p.
  105543 (2022)

\end{thebibliography}
}

\end{document}